\documentclass[runningheads]{llncs}

\usepackage{eccv}

\usepackage{eccvabbrv}

\usepackage{graphicx}
\usepackage{booktabs}
\usepackage{amsmath}
\usepackage{amssymb}
\usepackage{multirow}
\usepackage{wrapfig}
\usepackage{algorithm}

\usepackage{algpseudocode}
\usepackage{amsfonts}
\usepackage{makecell}
\usepackage{float}
\usepackage{listings}

\usepackage{tikz}
\usepackage{pgfplots}
\usetikzlibrary{patterns}
\pgfplotsset{compat=1.17}

\usepackage[table]{xcolor}
\definecolor{softblue}{RGB}{180,210,240}

\usepackage[accsupp]{axessibility}

\usepackage{hyperref}

\usepackage{orcidlink}

\newif\ifwithappendix
\withappendixtrue    

\begin{document}



\title{Reweighting Framewise Attention in Video Transformers for Facial Expression Understanding}

\titlerunning{MiRA for Facial Expression Understanding}


\author{
Seongro Yoon\inst{1}
\and
Donghyeon Cho\inst{2}
\and
Jinsun Park\inst{3}
\and
François Brémond\inst{1}
}

\authorrunning{Yoon et al.}

\institute{
Inria, Université Côte d'Azur, France\\
\email{\{seong-ro.yoon, francois.bremond\}@inria.fr}
\and
Hanyang University, South Korea\\
\email{doncho@hanyang.ac.kr}
\and
Pusan National University, South Korea\\
\email{jspark@pusan.ac.kr}
}


\maketitle

\begingroup
\renewcommand\thefootnote{}
\footnotetext{
Code and pretrained models are available at
\url{https://github.com/ysrinria/MiRA}.
}
\endgroup


\begin{abstract}
Understanding facial expressions in videos requires modeling subtle and localized facial dynamics under unconstrained conditions.
Although recent Vision Transformer~(ViT)-based video models have shown strong performance through large-scale self-supervised pretraining, their attention mechanisms often emphasize dominant global motions and coarse temporal dynamics, limiting sensitivity to fine-grained facial variations.
To address this limitation, we propose MiRA (Marginal-induced Attention Redistribution), a plug-in frame-marginal attention redistribution framework for ViT backbones that enhances spatio-temporal selectivity toward subtle facial dynamics without introducing additional trainable parameters.
MiRA derives frame-level confidence and intra-frame concentration statistics from self-attention maps to estimate frame-wise marginal importance and redistribute attention toward spatiotemporally localized facial cues.
We first introduce a principled \textit{exact mode} based on post-softmax attention redistribution.
To further improve efficiency, we propose \textit{flashLite mode}, a lightweight pre-softmax approximation that integrates frame-marginal redistribution into FlashAttention kernels while preserving the effectiveness of the exact formulation.
Experimental results on challenging Facial Expression Recognition~(FER) benchmarks demonstrate consistent improvements over strong ViT baselines.


\end{abstract}


\section{Introduction}
Learning effective representations of facial dynamics in videos requires capturing subtle spatio-temporal expression cues, such as transient wrinkles, frowns, and fine-grained muscle movements.
These subtle facial cues are often entangled with dominant global motions (\eg, head movements) and external variations (\eg, camera motion), making it challenging to accurately model facial dynamics.
Under such conditions, video-based facial expression understanding requires selectively allocating attention to frames containing informative expression dynamics, while suppressing redundant or motion-dominated frames.

Recent advances in general video understanding have demonstrated impressive performance on video benchmarks through large-scale self-supervised pretraining and downstream finetuning. 
Various paradigms have been explored for video representation learning, including contrastive approaches~\cite{rai2021cocon, han2020cotraining, qian2021spatiotemporal, dave2021tclr}, self-distillation frameworks~\cite{alayrac2020xdc, bardes2024vjepa}, and, more recently, masked video modeling based on Masked Autoencoders (MAEs)~\cite{wei2022maskfeat, wang2022bevt,feichtenhofer2022maest,tong2022videomae,wang2023videomae,bardes2024vjepa,bandara2023adamae,huang2023mgmae,hwang2022everest,pei2024videomac,gupta2023siammae,gundavarapu2024extending}. 
However, these models are primarily optimized to capture dominant scene-level dynamics, often focusing on coarse object motions.
When applied to facial videos, subtle intra-facial variations are often treated as redundant when co-occurring with more pronounced global movements.
To address this limitation, we investigate how generic ViT-based video models can be adapted to better capture facial dynamics.

Despite differences in formulation, most modern video models are built upon ViTs~\cite{vaswani2017attention, dosovitskiy2020image, arnab2021vivit}, which have become the canonical backbone for video representation learning.
Motivated by this, we redesign the self-attention mechanism in ViTs to shift attention from dominant global motions toward fine-grained facial contexts, enabling more effective modeling of subtle facial dynamics while retaining the strengths of the original ViT backbone.

\begin{figure}[t]
    \centering
    \includegraphics[width=0.8\linewidth]{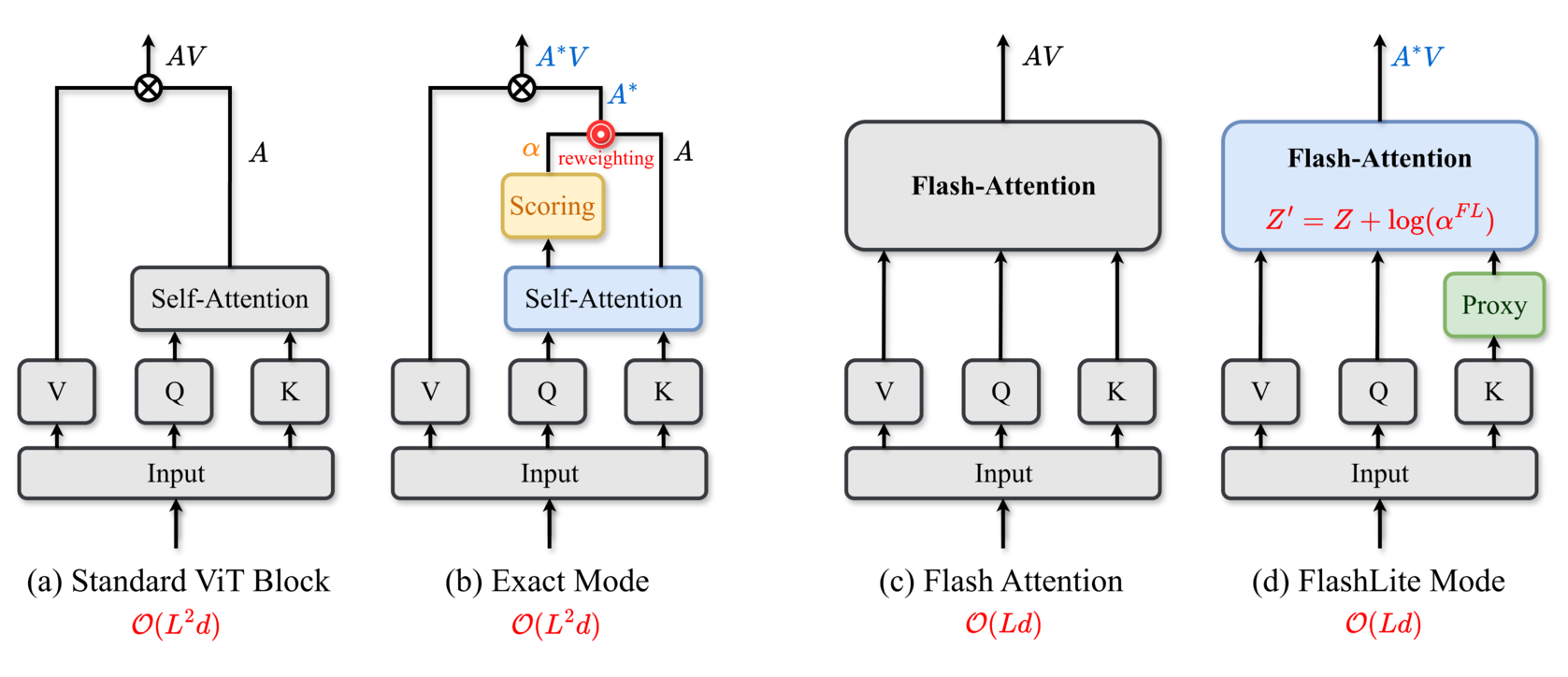}
    \caption{\textbf{Frame-marginal attention reweighting modules.}
    (b) \textit{Exact mode} applies principled framewise scoring and reweighting on the post-softmax attention map (a) by evaluating all query--key interactions. However, this design requires repeated memory access and is incompatible with fused kernels such as FlashAttention (c). Although FLOPs are similar, the additional memory I/O significantly slows execution.
    (d) \textit{FlashLite mode} instead computes key-wise proxies before attention computation and integrates the resulting weights ($\alpha^{\mathrm{FL}}$) directly into the FlashAttention kernel, reducing memory I/O while preserving accuracy.
    (\textit{Notation:} $L$: tokens, $d$: feature dimension, $Z = QK^\top / \sqrt{d_K}$: attention logits, $\mathcal{O}(\cdot)$: memory I/O complexity.)}
    \label{fig:modes}
\end{figure} 

In this work, we propose MiRA, a plug-in frame-marginal attention redistribution module for ViT backbones, and begin with its principled \textit{exact mode}.
In unconstrained facial videos without explicit face cropping, the proposed method redistributes attention using frame-level confidence and intra-frame concentration statistics, producing sharper and more localized attention patterns than those induced by dominant global motions.
Given the self-attention maps of a ViT block, the proposed method derives two complementary frame-level statistics: a \textit{confidence score}, measuring the total attention mass assigned to each frame across the sequence, and an \textit{intra-frame concentration score}, computed via inverse entropy to quantify the spatial concentration of attention within each frame.
These statistics are combined to estimate frame-wise marginal importance, which is then used to redistribute attention toward salient and localized facial dynamics.

The \textit{exact mode}, while effective, incurs additional memory I/O overhead since it requires access to the full post-softmax attention map, leading to repeated memory access beyond standard attention computation (see~\cref{fig:modes}).
To improve efficiency, we introduce \textit{flashLite mode}, which integrates frame-marginal redistribution into FlashAttention kernels~\cite{dao2022flashattention, dao2023flashattention2}.
While fused kernels achieve high efficiency by computing attention on the fly without materializing the full attention map, the exact mode relies on post-softmax attention values and therefore cannot be directly incorporated.
We address this limitation through a pre-softmax approximation that injects redistribution factors directly into the fused attention computation.
As a result, flashLite preserves the effectiveness of the exact mode while inheriting FlashAttention's efficiency benefits.

Our contributions are summarized as follows:
\vspace{-2pt}
\begin{itemize}
    \item[$\bullet$] We propose a plug-in frame-marginal attention redistribution module for ViT backbones that enhances sensitivity to subtle facial dynamics without introducing additional trainable parameters.
    
    \item[$\bullet$] We further introduce flashLite mode, a lightweight pre-softmax approximation that integrates the proposed frame-marginal redistribution into FlashAttention kernels, reducing memory I/O while preserving the effectiveness of the exact mode.
    
    \item[$\bullet$] We demonstrate consistent improvements over strong ViT baselines on large-scale video FER benchmarks without relying on task-specific heuristics (\eg, tight facial alignment), which are often impractical in scalable pretraining and real-world video settings.
\end{itemize}
\vspace{-4pt}
\section{Related Work}

\paragraph{Video Representation Learning.}
The success of ViTs in image recognition has inspired extensive efforts to extend them to video understanding tasks, where the goal is to model rich spatio-temporal dependencies across frames~\cite{arnab2021vivit}. 
Early approaches adapt the vanilla ViT by incorporating temporal attention to jointly learn spatial and temporal correlations within video clips. 
To improve efficiency and scalability, subsequent works introduced architectural refinements, including factorized spatio-temporal attentions~\cite{bertasius21a}, locality bias through windowed attention~\cite{liu2022video_swin}, and multiscale or pooled attention mechanisms that capture motion at different temporal resolutions~\cite{arnab2021vivit, li2021mvitv2, zhang2021vidtr, li2022uniformer}.
In parallel, masked prediction has emerged as a dominant paradigm for self-supervised video representation learning. 
Inspired by the success of masked language modeling~\cite{devlin2019bert} and masked autoencoders in the image domain~\cite{he2022mae, zhou2022ibot, oquab2023dinov2}, these approaches train models to reconstruct missing spatio-temporal information from sparsely observed inputs~\cite{wei2022maskfeat, wang2022bevt, feichtenhofer2022maest, tong2022videomae}. 
More recent developments enhance temporal reasoning by enforcing consistency across paired frames~\cite{pei2024videomac, gupta2023siammae}, introducing adaptive and content-aware masking strategies~\cite{wu2023dropmae, huang2023mgmae, bandara2023adamae, hwang2022everest}, and scaling pretraining to longer input sequences for improved temporal coverage~\cite{gundavarapu2024extending}.

\paragraph{Facial Dynamics Modeling in Video.}
The main challenge of facial expression modeling in videos lies in capturing fine-grained intra-facial dynamics under diverse and unconstrained conditions.
A growing body of work has proposed transformer-based models that directly model spatio-temporal facial dynamics, incorporating mechanisms such as temporal attention, context encoding, or landmark guidance~\cite{ma2022spatio, liu2022clip, wang2023rethinking, li2023intensity, liu2025robust}. 
Building on this trend, masked autoencoding has also been adapted to facial video, showing strong effectiveness for dynamic expression recognition. 
Recent advances include self-supervised MAE adaptations~\cite{sun2023mae, chen2024unilearn}, hierarchical or contrastive extensions~\cite{sun2024svfap, zhang2024mart}, and multimodal variants that integrate audio-visual information~\cite{sun2024hicmae, chumachenko2024mma, wu2025avf, cheng2024emotion}.
These efforts underscore a growing shift toward self-supervised and transformer-based frameworks as the dominant direction for facial expression understanding.

Despite significant progress, existing methods still struggle to capture subtle facial dynamics, particularly under unconstrained conditions.
Many approaches primarily emphasize dominant motion patterns or global temporal correlations, making it difficult to selectively focus on fine-grained intra-facial variations.
Consequently, spatio-temporally localized expression cues often remain insufficiently modeled.
Our work revisits the ViT attention mechanism through plug-in frame-marginal attention redistribution, enhancing spatio-temporal selectivity toward subtle facial dynamics.

\section{Method}
\label{sec:Method}
\begin{figure}[t]
    \centering
        \includegraphics[width=0.8\linewidth]{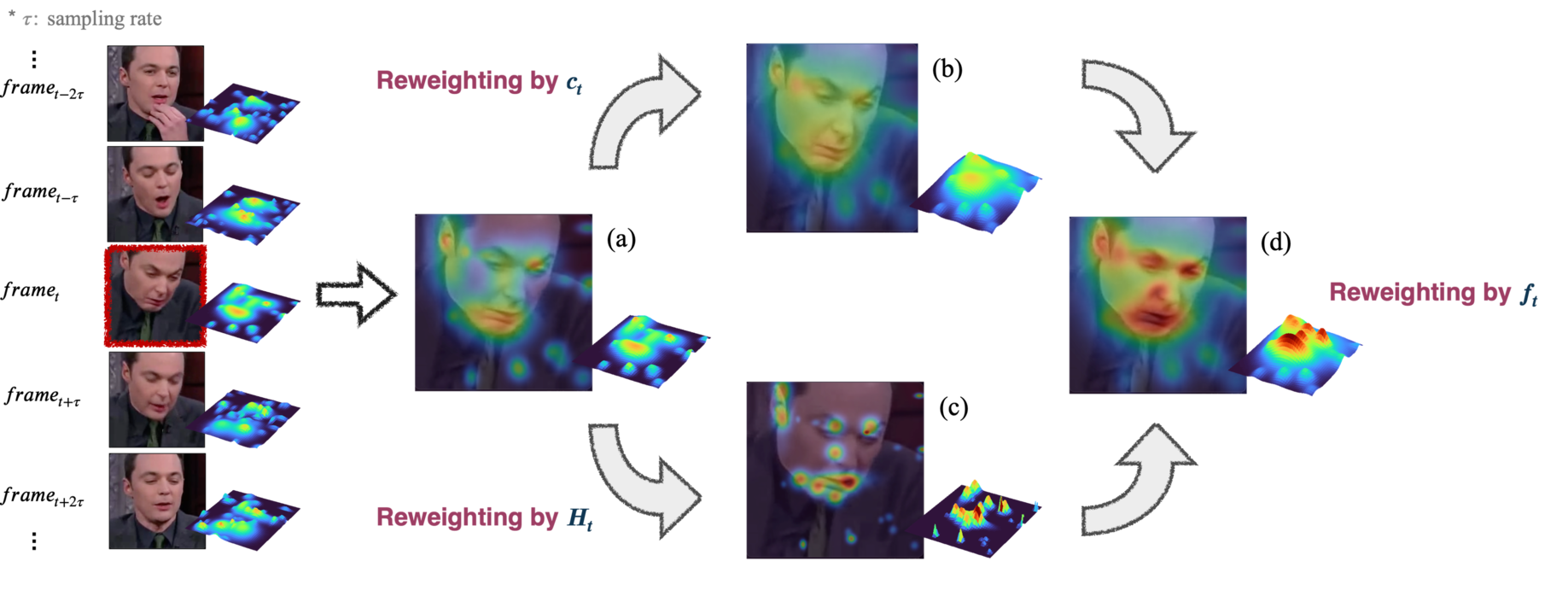}
    \caption{\textbf{Illustration of the effects of frame-level confidence and intra-frame concentration.}
    In-the-wild facial videos exhibit temporal redundancy and dominant global motions (e.g., head movements), which can overshadow subtle facial dynamics.
    (a) Baseline attention focuses on dominant scene-level variations and coarse global motions.
    (b) The frame-level confidence $c_t$ emphasizes informative frames while suppressing redundancy.
    (c) The intra-frame concentration $H_t$ promotes localized attention by emphasizing concentrated responses within each frame.
    (d) Their combination improves spatio-temporal selectivity toward informative facial moments.}
    \label{fig:scores}
\end{figure}

Facial expression representations rely on subtle spatio-temporal variations that are typically weak, transient, and spatially localized.
In facial videos, consistent face appearance induces temporal redundancy, and global motions can dominate self-attention, jointly overshadowing subtle facial dynamics.
We introduce MiRA, a frame-marginal reweighting mechanism that identifies informative facial moments through (i) temporal salience at the frame level and (ii) attention concentration within each frame, jointly enhancing spatio-temporal selectivity.
\cref{fig:scores} conceptually illustrates the intended effect of the proposed attention redistribution.
We first present the principled formulation of frame-marginal attention reweighting, referred to as \textit{exact mode}, followed by \textit{FlashLite mode}, a lightweight approximation that enables the effect of the exact formulation to be integrated into FlashAttention kernels~\cite{dao2022flashattention, dao2023flashattention2}.
\ifwithappendix
Standard background on video transformers is provided in Appendix~\ref{Appx:Preliminary}.
\fi

\subsection{Frame-Marginal Attention Reweighting: \textit{Exact Mode}}
\label{subsec:frame-marginal_attention_reweighting}

\paragraph{Frame-Level Scoring.}
Let $\mathcal{I}_{t} = \{ i_{t,n} = tN + n \}_{n=1}^{N}$ denote the set of patch-token indices corresponding to frame $t$ in a transformer block.
For simplicity, we define each temporal unit as a \emph{frame} $t$, where each frame corresponds to a temporal snippet at index $t$ determined by the temporal granularity of the tokenizer.
Given the head-averaged attention matrix $\bar{A} \in \mathbb{R}^{L \times L}$, we define the \emph{frame-level confidence score} as
\begin{equation}
    c_t = \frac{1}{L} \sum_{q=1}^{L} \sum_{i_{t,n} \in \mathcal{I}_t} \bar{A}_{q,i_{t,n}},
    \label{eq:confidence}
\end{equation}
which measures the marginal attention mass assigned to frame $t$ when treated as keys, indicating how strongly it is attended to across all queries.
We further normalize $c_t$ across frames, i.e., $c_t \leftarrow c_t / \sum_{t'} c_{t'}$.
To quantify the concentration of attention over tokens within each frame, we first define a normalized intra-frame attention distribution:
\begin{equation}    
    p_{t,n} =
    \frac{\sum_{q=1}^{L} \bar{A}_{q,i_{t,n}}}
         {\sum_{q=1}^{L} \sum_{j \in \mathcal{I}_{t}} \bar{A}_{q,j}}.
    \label{eq:entropy_proportion}
\end{equation}
Here, $p_{t,n}$ defines the relative attention distribution over tokens within frame $t$.
Based on this distribution, we define the \emph{intra-frame concentration score} via inverse entropy:
\begin{equation}
    H_t =
    \left(
        -\sum_{n=1}^{N}
        p_{t,n} \log(p_{t,n} + \epsilon)
    \right)^{-1},
    \label{eq:inv_entropy}
\end{equation}
which assigns larger values to more spatially concentrated attention patterns, encouraging localized attention over diffuse global patterns.
While $c_t$ captures global salience, $H_t$ reflects the spatial selectivity of attention within a frame, jointly guiding the model to identify facial expression cues ranging from global changes to subtle, region-specific dynamics.

To enable attention redistribution using these complementary signals, we define a frame-marginal importance score:
\begin{equation}
    f_t = w_{\text{con}}\, c_t + w_{\text{ent}}\, H_t,
    \label{eq:composite}
\end{equation}
where $w_{\text{con}}$ and $w_{\text{ent}}$ are fixed weights that balance the confidence and concentration terms.
To use $f_t$ as a prior for attention reweighting, we convert it into a normalized distribution over frames.
We first apply per-frame min-max normalization to mitigate scale variations, and then transform the normalized scores into a probability distribution using a power-softmax:
\begin{equation}
f'_t =
\frac{f_t - \min_{t'} f_{t'}}{\max_{t'} f_{t'} - \min_{t'} f_{t'} + \epsilon},
\quad
\pi_t =
(1 - \lambda_\pi)
\frac{(f'_t)^\beta}{\sum_{t'} (f'_{t'})^\beta}
+ \frac{\lambda_\pi}{T}.
\label{eq:prior}
\end{equation}
Here, $\beta$ controls the sharpness of the distribution, while $\lambda_\pi$ mixes a uniform component to prevent degenerate solutions.
The resulting prior $\pi_t$ defines a \emph{frame-marginal redistribution} of attention, reallocating the frame-level aggregated attention mass across frames based on the confidence and concentration scores, thereby enhancing spatio-temporal selectivity.
This aligns with the challenges in unconstrained facial videos, where attention is often biased toward dominant motions and temporal redundancy, obscuring informative facial moments.

\paragraph{Attention Reweighting.}
The resulting scaling factors are applied to the original multi-head attention $A$, while the scoring functions are computed from the head-averaged attention $\bar{A}$.
Directly replacing $A$ with $\pi_t$ may distort learned attention patterns, as $\pi_t$ encodes only frame-level marginals and does not preserve the original query-conditioned distribution over keys.
To address this, we perform a ratio-based alignment that adjusts the frame-level attention distribution toward $\pi_t$ while preserving the original query--key attention structure.
Conceptually, this can be interpreted as a marginal-alignment rescaling, analogous to a single step of Sinkhorn normalization~\cite{cuturi2013sinkhorn}, a procedure that alternates row- and column-wise normalizations to match prescribed marginals.

We define the scaling factors as
\begin{equation}
    \alpha_{t} = \mathtt{clip}\left(\frac{\pi_{t}}{c_{t} + \epsilon}, \alpha_{\text{min}}, \alpha_{\text{max}}\right),
    \label{eq:alpha}
\end{equation}
where $\mathtt{clip}(\cdot)$ bounds the scaling to $[\alpha_{\min}, \alpha_{\max}]$.
This scaling adjusts the frame-level attention mass by amplifying underrepresented frames ($c_t < \pi_t$) and attenuating overrepresented ones ($c_t > \pi_t$), 
thereby aligning the attention with the target prior $\pi_t$ while reinforcing frame-level selectivity and spatial localization.
Importantly, the ratio-based formulation preserves the relative query-wise attention structure while correcting only the frame-level marginals.
Applying $\alpha_t$ rescales all key tokens in frame $t$, followed by a small uniform mixing and row-wise normalization:
\begin{equation}
\tilde{A}_{q,i_{t,n}} = \alpha_t A_{q,i_{t,n}}, \quad
\hat{A}_{q,i} = (1 - \epsilon_{\mathrm{floor}})\tilde{A}_{q,i} + \frac{\epsilon_{\mathrm{floor}}}{L}, \quad
A'_{q,i} = \frac{\hat{A}_{q,i}}{\sum_{j=1}^{L} \hat{A}_{q,j}}.
\label{eq:reweight}
\end{equation}
Here, $\tilde{A}_{q,i_{t,n}}$ denotes the frame-level rescaled attention, $\hat{A}_{q,i}$ introduces a small uniform component to prevent vanishing or overly peaky values, and $A'_{q,i}$ enforces a valid probability distribution via row-wise normalization.
A final interpolation with the original attention preserves the learned query--key structure:
\begin{equation}
A^{\star} = (1 - \eta)A + \eta A',
\label{eq:residual}
\end{equation}
while preventing excessive deviations from the original attention structure.

\subsection{Approximation for FlashAttention: \textit{FlashLite Mode}}
\label{subsec:flashLite_approximation}
While the exact mode of MiRA does not introduce additional trainable parameters, it remains computationally demanding because reweighting is applied after softmax in self-attention.
This requires materializing the full attention map $A \in \mathbb{R}^{L \times L}$, performing row-wise normalization as in~\eqref{eq:reweight}, and retaining post-softmax activations during backpropagation, all of which require additional passes over the $\mathcal{O}(L^2)$ attention map, leading to increased memory I/O and compute compared to standard self-attention.
In contrast, FlashAttention~\cite{dao2022flashattention, dao2023flashattention2} achieves $\mathcal{O}(L)$ memory I/O by computing attention in a streaming manner: it tiles $Q$ and $K$, computes blocks of $QK^{\top}$, applies blockwise softmax with numerically stable accumulators, and immediately contracts with $V$ to produce $Y = AV$.
This streaming design avoids materializing the full attention map, significantly reducing memory I/O.
However, our exact formulation is not directly compatible with this design, as it requires access to post-softmax attention values that are not retained in the streaming computation.

To overcome this limitation, we integrate the frame-level scaling factors into the attention computation as a pre-softmax bias.
Instead of rescaling the attention after normalization as in~\eqref{eq:reweight}, the scaling factors are incorporated directly into the logits as biases:
\begin{equation}
    Z'_{q,i_{t,n}} = Z_{q,i_{t,n}} + \log \alpha_t, \quad
    A' = \mathtt{softmax}(Z'),
\label{eq:fma_flashlite}
\end{equation}
where $Z = QK^\top / \sqrt{d_K}$ denotes the attention logits.
Since post-softmax scaling is followed by renormalization, multiplying an attention entry by $\alpha_t$ is equivalent to scaling its softmax numerator, i.e., $\alpha_t \exp(z_i)=\exp(z_i+\log\alpha_t)$, and thus to adding $\log\alpha_t$ to the logits before the softmax.
Unlike the exact mode, which performs redistribution directly in the attention probability space after softmax normalization, flashLite applies framewise modulation in logit space before softmax.
Consequently, the two modes require different parameterizations to achieve stable framewise redistribution.

\paragraph{Key-Based Proxy for $\alpha_t$.}
The scaling factors $\alpha_t$ in the exact mode depend on frame-level attention marginals derived from the attention map, which is not materialized in FlashAttention due to its streaming computation.
To enable framewise modulation without explicit attention reconstruction, flashLite instead approximates the required frame-level statistics directly from the multi-head key representations $K$.
Since attention logits are computed from query--key similarities, tokens with larger key energy can contribute to larger query--key logit magnitudes, providing a lightweight surrogate for estimating frame-level salience.
Let $M$ denote the number of attention heads, and $d_m = d_K / M$ the dimensionality of each head.
Given the multi-head key tensor $K \in \mathbb{R}^{M \times L \times d_{m}}$, the per-token key energy $E \in \mathbb{R}^{T \times N}$ is defined as
\begin{equation}
    E_{t, n} = \frac{1}{M}\sum_{m=1}^{M}
    \frac{1}{d_{m}}
    \left\| K_{m, i_{t,n}} \right\|_2^2,
    \label{eq:kenergy}
\end{equation}
where $E_{t,n}$ represents the average squared key energy across attention heads for token $n$ in frame $t$.
Using $E_{t,n}$, we compute the corresponding scores as
\begin{equation}
\tilde{c}_{t} = \sum_{n=1}^{N} E_{t, n}, \quad
\tilde{p}_{t, n} = \frac{E_{t, n}}{\sum_{n^{\prime}=1}^{N} E_{t, n^{\prime}}}, \quad
\tilde{H}_{t} =
\left(
    -\sum_{n=1}^{N}
    \tilde{p}_{t, n}\,\log\!\big(\tilde{p}_{t, n} + \epsilon \big)
\right)^{-1}
\label{eq:kstats_all}
\end{equation}
where $\tilde{c}_t$ and $\tilde{H}_t$ approximate the attention-based confidence~\eqref{eq:confidence} and inverse entropy scores~\eqref{eq:inv_entropy}, respectively.

The two proxies are combined into the frame-marginal importance score $\tilde{f}_t$ following~\eqref{eq:composite}.
Since the key-energy surrogate is computed prior to softmax normalization, its framewise statistics exhibit weaker inter-frame contrast than post-softmax attention marginals.
We therefore standardize $\tilde{f}_t$ across frames and convert it into a frame prior using a temperature-controlled softmax:
\begin{equation}
\tilde{f}_t^{\prime} =
\frac{\tilde{f}_t - \mu_{\tilde{f}}}
     {\sigma_{\tilde{f}} + \epsilon},
\quad
\pi_t^{\mathrm{FL}} =
(1-\lambda_\pi)
\frac{\exp(\tau \tilde{f}_t^{\prime})}
     {\sum_{t'} \exp(\tau \tilde{f}_{t'}^{\prime})}
+ \frac{\lambda_\pi}{T}.
\label{eq:flashlite_prior}
\end{equation}
where $\mu_{\tilde{f}}$ and $\sigma_{\tilde{f}}$ denote the mean and standard deviation of $\tilde{f}_t$ over frames, while $\tau$ controls the sharpness of the resulting frame prior.
Unlike the exact mode, which redistributes normalized attention marginals in probability space after softmax, flashLite performs modulation directly in logit space through additive biases.
Accordingly, flashLite employs a temperature-controlled softmax instead of the power transformation in~\eqref{eq:prior}, since the modulation is applied before softmax normalization and therefore interacts differently with the attention distribution.
The resulting prior is converted into a unit-centered multiplicative factor, i.e., $\alpha_t^{\mathrm{FL}} = \operatorname{clip}(T \cdot \pi_t^{\mathrm{FL}}, \alpha_{\min}, \alpha_{\max})$.
In contrast to the exact formulation, we do not use the ratio $\pi_t^{\mathrm{FL}}/\tilde{c}_t$, since both terms are derived from the same key-energy surrogate.
Such a ratio would substantially suppress framewise contrast, weakening the intended corrective effect of the modulation and causing the scaling factors to collapse toward near-uniform values.
Finally, these scaling factors are injected into the logits following~\eqref{eq:fma_flashlite}, with $\alpha_t$ replaced by $\alpha_t^{\mathrm{FL}}$, enabling framewise redistribution without disrupting the streaming computation pattern of FlashAttention.
\ifwithappendix
Further analysis and justification of this approximation are provided in Appendix~\ref{Appx:FlashLiteRationale}.
\fi
\section{Experiments}
We adopt VideoMAE~\cite{tong2022videomae} as the backbone under the standard pretrain--finetune paradigm.
During self-supervised pretraining, the model learns spatio-temporal representations via masked reconstruction, and the same ViT encoder is subsequently finetuned for downstream facial expression recognition tasks.
The proposed MiRA module is integrated into each self-attention block in a plug-and-play manner without introducing additional learnable parameters.
\ifwithappendix
Further details on implementation and model configurations are provided in Appendix~\ref{Appx:ImplementationDetails} and \ref{Appx:ModelConfigurations}.
\fi

\subsection{Datasets}
\label{subsec:datasets}
For pretraining, we use VoxCeleb2~\cite{voxceleb2}, a large-scale face video corpus with over 1.2M real-world utterances spanning diverse speakers and natural affective variations.
For finetuning, we evaluate on large-scale in-the-wild benchmarks: DFEW~\cite{dfew}, MAFW~\cite{mafw}, and FERV39k~\cite{wang2022ferv39k}.
DFEW and MAFW include around 12K and 10K clips with 7 and 11 emotion categories, respectively, sourced from movies and TV shows. We follow their standard 5-fold cross-validation protocols.
FERV39k consists of about 39K video clips with 7 expression categories, drawn from online videos across 22 real-world scenes.

\subsection{Ablation Study}
\label{subsec:ablation_study}
We primarily evaluate the exact mode as the canonical form to isolate the intrinsic effect of the proposed formulation, with ViT-B as the vanilla architecture.
All ablations are conducted on DFEW and MAFW with a shared VideoMAE-based ViT-B backbone, pretrained on VoxCeleb2 and finetuned on the target datasets.
Unless otherwise specified, results are reported on the fold-1 split, as data characteristics across folds are largely similar.
Hyperparameters are initialized with default values ($w_{\text{con}}{=}0.5$, $w_{\text{ent}}{=}0.5$, $\beta{=}1.5$, and $\tau{=}1.7$ for flashLite), tuned via validation, and then fixed for all experiments.

\paragraph{Effect of Reweighting Depth.}
We study the impact of applying the reweighting module at different transformer depths, as its role may vary between upper-layer refinement and full-backbone adaptation.
To isolate the intrinsic effect of the proposed module, we perform ablations by finetuning a fixed VideoMAE backbone pretrained on VoxCeleb2.
For a 12-layer ViT-B, we vary the number of topmost reweighting blocks ($\kappa$).
As shown in \cref{tab:ablation_K}, full-depth reweighting ($\kappa{=}12$) achieves the best UAR/WAR on both DFEW and MAFW, while restricting the module to the top layers ($\kappa{=}2$) remains competitive with only a minor drop.
In contrast, intermediate settings ($\kappa{=}6$ or $10$) result in lower performance.
We hypothesize that shallow reweighting primarily refines high-level semantics, whereas full-depth application enables coherent adaptation across the entire feature hierarchy, while intermediate depths may introduce inconsistencies across the attention hierarchy.
Based on these observations, $\kappa{=}12$ yields the best performance, whereas $\kappa{=}2$ may offer an efficient trade-off.
\begin{table}[t]
\centering

\begin{minipage}[t]{0.49\linewidth}
\centering

\scriptsize
\setlength{\tabcolsep}{5pt}
\renewcommand{\arraystretch}{0.95}

\caption{Effect of reweighting blocks ($\kappa$).}
\vspace{-4pt}
\label{tab:ablation_K}

\resizebox{\linewidth}{!}{
\begin{tabular}{c c cccc}
\toprule

Dataset
& Metric
& $\kappa=2$
& $\kappa=6$
& $\kappa=10$
& $\kappa=12$ \\

\midrule

\multirow{2}{*}{DFEW-f.1}
& {\tiny UAR}
& \underline{62.81}
& 62.75
& 61.65
& \textbf{63.28} \\

& {\tiny WAR}
& \underline{75.14}
& 74.46
& 74.63
& \textbf{75.57} \\

\midrule

\multirow{2}{*}{MAFW-f.1}
& {\tiny UAR}
& \underline{36.74}
& 36.55
& 36.62
& \textbf{38.81} \\

& {\tiny WAR}
& \underline{48.59}
& 48.53
& 47.77
& \textbf{49.40} \\

\bottomrule
\end{tabular}
}

\end{minipage}
\hfill
\begin{minipage}[t]{0.49\linewidth}
\centering

\scriptsize
\setlength{\tabcolsep}{3pt}
\renewcommand{\arraystretch}{1.03}

\caption{Effect of $w_{\text{con}}$ and $w_{\text{ent}}$.}
\vspace{-4pt}
\label{tab:ablation_weights}

\resizebox{\linewidth}{!}{
\begin{tabular}{cccccccccc}
\toprule

& & \multicolumn{4}{c}{$\kappa=2$}
& \multicolumn{4}{c}{$\kappa=12$} \\

\cmidrule(lr){3-6}
\cmidrule(lr){7-10}

& & \multicolumn{2}{c}{DFEW-f.1}
& \multicolumn{2}{c}{MAFW-f.1}
& \multicolumn{2}{c}{DFEW-f.1}
& \multicolumn{2}{c}{MAFW-f.1} \\[1.pt]

$w_\text{con}$
& $w_\text{ent}$
& {\tiny UAR}
& {\tiny WAR}
& {\tiny UAR}
& {\tiny WAR}
& {\tiny UAR}
& {\tiny WAR}
& {\tiny UAR}
& {\tiny WAR} \\[1.pt]

\midrule

0.3
& 0.7
& \underline{62.01}
& \textbf{75.18}
& 36.05
& 48.42
& 62.06
& 75.05
& 36.50
& \textbf{49.51} \\

0.5
& 0.5
& \textbf{62.81}
& \underline{75.14}
& \underline{36.74}
& \underline{48.59}
& \textbf{63.28}
& \underline{75.57}
& \textbf{38.81}
& \underline{49.40} \\

0.7
& 0.3
& 61.10
& 74.97
& \textbf{36.87}
& \textbf{48.69}
& \underline{62.31}
& \textbf{75.78}
& \underline{37.80}
& 48.20 \\[1.pt]

\bottomrule
\end{tabular}
}
\end{minipage}
\vspace{-4pt}
\end{table}

\paragraph{Effect of $w_{\text{con}}$ and $w_{\text{ent}}$.}
Under the same experimental setting, we examine the effect of weighting factors in the composite score.
The frame-marginal importance score $f_t$ in~\eqref{eq:composite} combines frame-level confidence $c_t$ and intra-frame concentration $H_t$, weighted by $w_{\text{con}}$ and $w_{\text{ent}}$, respectively.
We vary $[w_{\text{con}}, w_{\text{ent}}]$ to assess sensitivity.
As shown in~\cref{tab:ablation_weights}, performance differences across settings are minor, with balanced weighting ($[0.5, 0.5]$) achieving near-optimal results.

\paragraph{Effect of Power Factor $\beta$.}
In~\eqref{eq:prior}, $\beta$ controls the sharpness of the transformed importance score distribution before normalization, thereby determining how strongly the prior emphasizes salient frames.
A larger $\beta$ increases the contrast among frame-level importance scores $f_t$, whereas a smaller $\beta$ compresses these differences, producing a more uniform prior $\pi_t$.
To assess its influence, we first evaluate three representative values ($0.7$, $1.0$, $1.5$), as shown in~\cref{tab:beta-results}.
$\beta{=}1.5$ generally provides the strongest performance across datasets and configurations ($\kappa{=}2$ and $\kappa{=}12$).
Guided by this observation, we further examine values in the neighborhood of $\beta{=}1.5$, where performance remains consistently strong with only marginal fluctuations across settings.
\begin{table}[t]
\centering
\scriptsize
\caption{Effect of the power factor $\beta$ under different $\kappa$ settings. {\scriptsize(\textbf{Best}, \underline{Second})}}
\vspace{-8pt}
\setlength{\tabcolsep}{3pt}
\resizebox{0.5\linewidth}{!}{
\begin{tabular}{c @{\hspace{14pt}} cc cc @{\hspace{14pt}} cc cc}
\toprule
  & \multicolumn{4}{c}{$\kappa=2$} & \multicolumn{4}{c}{$\kappa=12$} \\
  \midrule
  & \multicolumn{2}{c}{\textbf{DFEW-f.1}} & \multicolumn{2}{c}{\textbf{MAFW-f.1}} & \multicolumn{2}{c}{\textbf{DFEW-f.1}} & \multicolumn{2}{c}{\textbf{MAFW-f.1}} \\
  & {\tiny UAR} & {\tiny WAR} & {\tiny UAR} & {\tiny WAR}
  & {\tiny UAR} & {\tiny WAR} & {\tiny UAR} & {\tiny WAR} \\
\midrule
$\beta=0.7$ & 55.87 & 68.56 & 36.38 & 48.37 & 54.03 & 66.98 & 37.97 & 49.29 \\
$\beta=1.0$ & 56.62 & 68.94 & 36.45 & 48.26 & 55.01 & 67.88 & \underline{38.53} & \textbf{49.62} \\
$\beta=1.4$ & 61.66 & \underline{75.18} & 35.92 & 48.42 & 62.51 & 75.14 & 38.42 & 49.29 \\ 
$\beta=1.5$ & \textbf{62.81} & 75.14 & \textbf{36.74} & \textbf{48.59} & \textbf{63.28} & \textbf{75.57} & \textbf{38.81} & 49.40 \\
$\beta=1.6$ & \underline{61.93} & \textbf{75.48} & \underline{36.68} & \underline{48.53} & \underline{62.85} & 75.01 & 36.78 & \underline{49.51} \\
$\beta=1.7$ & 61.81 & \underline{75.18} & 36.31 & 48.26 & 62.66 & \underline{75.18} & 38.29 & 48.75 \\
\bottomrule
\end{tabular}%
}
\label{tab:beta-results}
\end{table}

\paragraph{Beyond Naïve Framewise Modeling.}
\begin{table}[t]
\centering
\begin{minipage}[t]{0.4\linewidth}
\centering
\scriptsize
\setlength{\tabcolsep}{2.5pt}
\renewcommand{\arraystretch}{0.36}
\caption{Comparison with baseline variants on DFEW fold-1.}
\vspace{-4pt}
\label{tab:module-results}
\begin{tabular}{lcc}
\toprule
Method & \tiny UAR & \tiny WAR \\
\midrule
Vanilla                     & 59.94 & 72.90 \\
(a) Reused Intra Att.       & 59.31 & 72.36 \\
(b) Framewise Aug. Block    & 62.68 & 74.93 \\
(c) Param-Matched (b)       & 60.83 & 72.62 \\
Ours finetuned ($\kappa=2$)        & 62.81 & 75.14 \\
Ours finetuned ($\kappa=12$)       & \underline{63.28} & \underline{75.57} \\
Ours pretrained                     & \textbf{63.88} & \textbf{76.21} \\
\bottomrule
\end{tabular}
\end{minipage}
\hfill
\begin{minipage}[t]{0.56\linewidth}
\vspace{0pt}
\centering
\captionof{figure}{Illustration of baseline attention variants.}
\vspace{-4pt}
\label{fig:baselines}
\includegraphics[width=\linewidth]{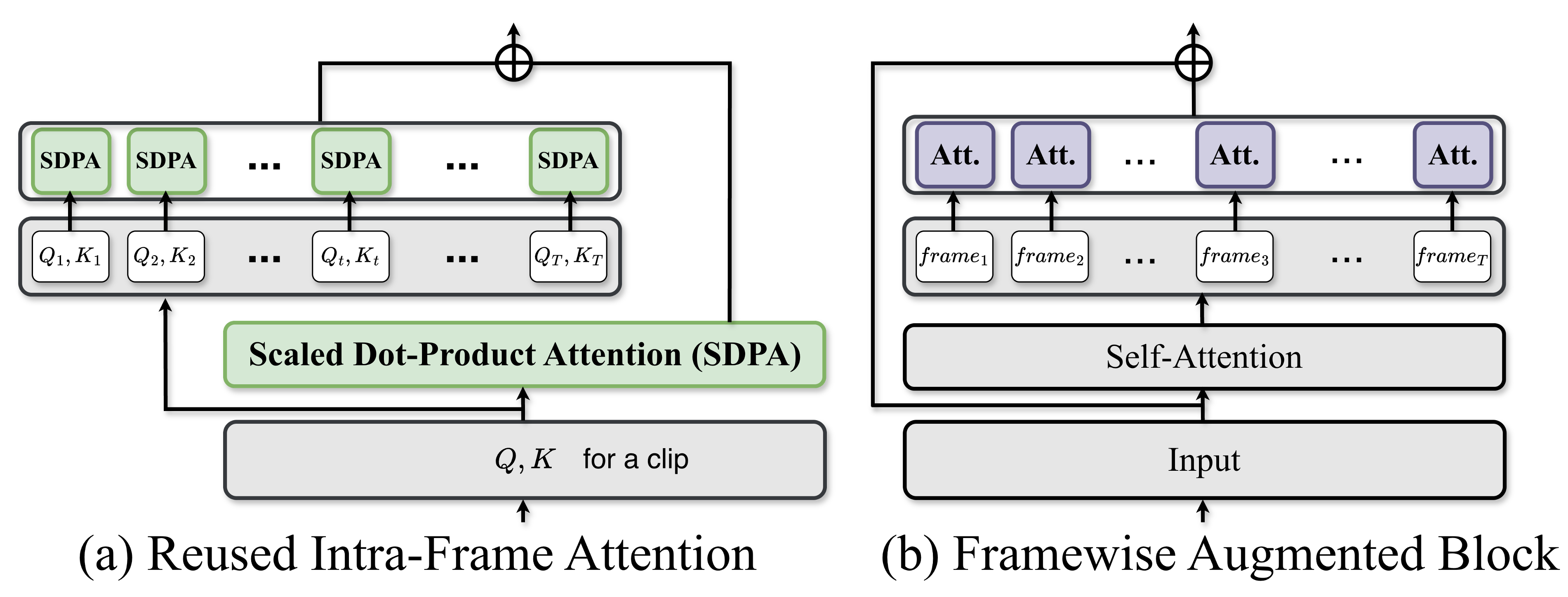}
\end{minipage}
\end{table}
We compare against three baseline variants that enhance framewise or temporal modeling:
(a) \textit{reused intra-frame attention}, adding a branch that confines attention within each frame and combines it with global attention;
(b) \textit{framewise augmented block} (+20\% params), which augments the transformer with a framewise block operating independently within each frame; and
(c) \textit{parameter-matched version of (b)}, where global head dimensions are reduced to match the parameter budget of the vanilla model, as illustrated in~\cref{fig:baselines}.
To ensure a fair comparison, all baselines are trained consistently from pretraining on VoxCeleb2 through finetuning on the target dataset.
As shown in~\cref{tab:module-results}, our models outperform all baselines, indicating that naive framewise enhancements alone are insufficient to capture informative facial dynamics.
Facial expressions are often temporally sparse and spatially localized, requiring the model to focus on informative moments.
By jointly leveraging frame-level confidence and intra-frame concentration, the proposed approach demonstrates improved capacity to model these dynamics.

\paragraph{Exact Mode vs.\ FlashLite Mode.}
\begin{figure*}[h]
    \centering
    \includegraphics[width=\textwidth]{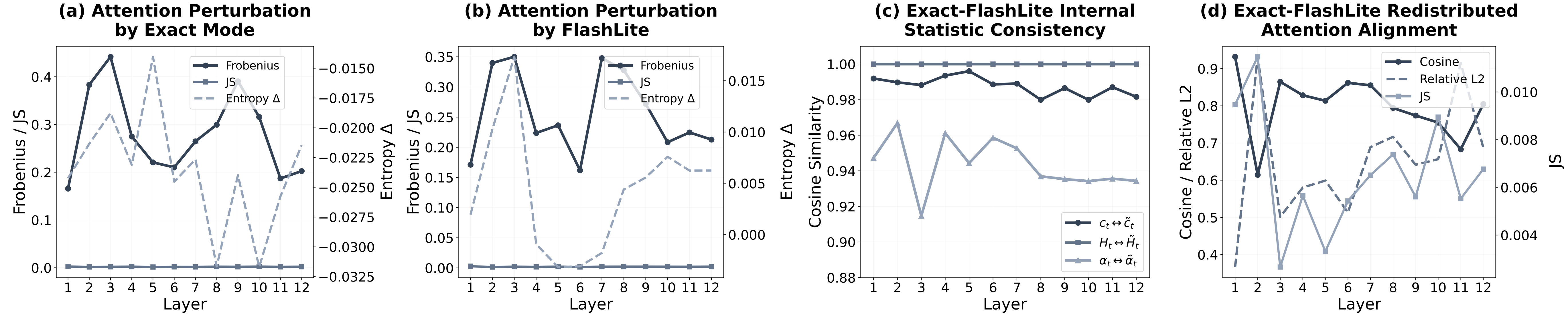}
    \caption{
\textbf{Behavioral analysis of Exact and FlashLite modes.}
(a,b) Attention perturbation relative to vanilla attention, measured by Frobenius distance on attention maps, together with JS divergence and entropy difference on framewise temporal attention distributions.
(c) Consistency of frame-level confidence $c_t$, concentration $H_t$, and modulation factor $\alpha_t$ between the two modes.
(d) Alignment between redistributed attentions, measured using cosine similarity (directional agreement) and relative $\ell_2$ distance (relative deviation magnitude).
}

\label{fig:behavioral_analysis}
\end{figure*}
We conduct layer-wise analyses of attention perturbation, internal statistics, and redistributed attention alignment for both modes.
In~\cref{fig:behavioral_analysis}(a,b), each mode is compared against vanilla attention to examine how modulation alters transformer attention across layers.
Frobenius distance $\|A_{\mathrm{reweighted}}-A_{\mathrm{vanilla}}\|_F$ measures attention perturbation magnitude, indicating that both exact and flashLite consistently modify transformer attention relative to vanilla attention.
JS divergence is computed between temporal attention distributions $p_{\mathrm{vanilla}}$ and $p_{\mathrm{reweighted}}$, obtained by marginalizing attention over spatial tokens within each frame.
Its consistently low values indicate that the global temporal allocation pattern remains largely preserved after modulation.
Likewise, entropy differences $\mathcal{H}(p_{\mathrm{reweighted}})-\mathcal{H}(p_{\mathrm{vanilla}})$ are computed on the same temporal attention distributions.
Their consistently small magnitudes suggest only modest variations in concentration behavior.
Overall, these analyses suggest that both exact and flashLite introduce structured redistribution while preserving the original temporal attention organization.
\cref{fig:behavioral_analysis}(c) compares the internal statistics used for modulation, showing that flashLite closely preserves the internal modulation dynamics of the exact formulation.
\cref{fig:behavioral_analysis}(d) directly compares redistributed attentions between the two modes, indicating that redistribution derived from key-space statistics closely approximates that of the exact formulation.
Consistent with these analyses, exact and flashLite exhibit nearly identical recognition performance across all folds on DFEW and MAFW (\cref{tab:Main-exact-flash}).
Together with this close approximation, flashLite provides a latency reduction of 20\% (400.6\,ms $\rightarrow$ 318.7\,ms), corresponding to a throughput gain of about 26\% on a single H100 GPU in practice.
\ifwithappendix
Detailed computational analysis is provided in Appendix~\ref{Appx:ComputationalEfficiencyAnalysis}.
\fi
\begin{table}[t]
\centering
\scriptsize

\begin{minipage}[h]{0.56\columnwidth}
\centering

\caption{Performance comparison of Exact and FlashLite on ViT-B over 5 folds.}
\vspace{-4pt}
\label{tab:Main-exact-flash}

\renewcommand{\arraystretch}{0.95}

\resizebox{\linewidth}{!}{
\begin{tabular}{@{}lcccccccccc@{}}
\toprule

\textbf{DFEW}
& \multicolumn{2}{c}{\textbf{Fold1}}
& \multicolumn{2}{c}{\textbf{Fold2}}
& \multicolumn{2}{c}{\textbf{Fold3}}
& \multicolumn{2}{c}{\textbf{Fold4}}
& \multicolumn{2}{c}{\textbf{Fold5}} \\

& {\tiny UAR} & {\tiny WAR}
& {\tiny UAR} & {\tiny WAR}
& {\tiny UAR} & {\tiny WAR}
& {\tiny UAR} & {\tiny WAR}
& {\tiny UAR} & {\tiny WAR} \\

\midrule

Exact
& 63.88 & 76.21
& 64.32 & 75.10
& 65.67 & 75.64
& 64.28 & 74.35
& 68.28 & 78.00 \\

FlashLite
& 63.77 & 76.51
& 64.16 & 74.93
& 65.06 & 76.15
& 64.87 & 75.63
& 67.69 & 78.04 \\

\bottomrule
\end{tabular}
}

\vspace{2pt}

\resizebox{\linewidth}{!}{
\begin{tabular}{lcccccccccc}
\toprule

\textbf{MAFW}
& \multicolumn{2}{c}{\textbf{Fold1}}
& \multicolumn{2}{c}{\textbf{Fold2}}
& \multicolumn{2}{c}{\textbf{Fold3}}
& \multicolumn{2}{c}{\textbf{Fold4}}
& \multicolumn{2}{c}{\textbf{Fold5}} \\

& {\tiny UAR} & {\tiny WAR}
& {\tiny UAR} & {\tiny WAR}
& {\tiny UAR} & {\tiny WAR}
& {\tiny UAR} & {\tiny WAR}
& {\tiny UAR} & {\tiny WAR} \\

\midrule

Exact
& 39.67 & 50.87
& 42.44 & 55.63
& 49.69 & 62.19
& 48.36 & 63.37
& 46.02 & 61.39 \\

FlashLite
& 39.20 & 51.20
& 43.65 & 55.53
& 48.81 & 61.54
& 49.83 & 63.48
& 45.91 & 60.13 \\

\bottomrule
\end{tabular}
}

\end{minipage}
\hfill
\begin{minipage}[h]{0.42\columnwidth}
\centering

\caption{Comparison with general video models. {\scriptsize(\textbf{Best}, \underline{Second})}}
\vspace{-4pt}
\label{tab:Main-general-video-only}

\renewcommand{\arraystretch}{1.13}

\resizebox{0.79\linewidth}{!}{
\begin{tabular}{lcccc}
\toprule

\textbf{Method}
& \multicolumn{2}{c}{\textbf{DFEW-f.1}}
& \multicolumn{2}{c}{\textbf{MAFW-f.1}} \\

& {\tiny UAR} & {\tiny WAR}
& {\tiny UAR} & {\tiny WAR} \\

\midrule

VideoMAE\,\cite{tong2022videomae}
& 59.94 & 72.90
& 38.80 & 48.80 \\

VideoMAEv2\,\cite{wang2023videomae}
& 59.16 & 73.56
& 37.40 & 49.51 \\

AdaMAE\,\cite{bandara2023adamae}
& 61.73 & 75.31
& 37.59 & 50.27 \\

MGMAE\,\cite{huang2023mgmae}
& 60.95 & 75.01
& 37.68 & 50.27 \\

EVEREST\,\cite{hwang2022everest}
& 54.83 & 66.85
& 36.26 & 47.77 \\

\midrule

\rowcolor[rgb]{0.90,0.95,1.0}
ViT-B Exact
& \textbf{63.88} & \underline{76.21}
& \textbf{39.67} & \underline{50.87} \\

\rowcolor[rgb]{0.90,0.95,1.0}
ViT-B FlashLite
& \underline{63.77} & \textbf{76.51}
& \underline{39.20} & \textbf{51.20} \\

\bottomrule
\end{tabular}
}

\end{minipage}
\end{table}

\subsection{Main Results}
\paragraph{Comparison with General Video Models.}
All models are pretrained on VoxCeleb2 under identical settings and evaluated on the fold-1 splits of DFEW and MAFW.
Building upon VideoMAE, VideoMAEv2 focuses on scaling to larger models by improving efficiency through sparse masking.
AdaMAE employs a classification-based sampler to prioritize informative tokens, MGMAE enforces temporal consistency via motion-guided masking, and EVEREST reduces redundancy by focusing on information-dense frames, all aiming to enhance spatio-temporal token correlation.
As shown in~\cref{tab:Main-general-video-only}, our approach outperforms these baselines, demonstrating the effectiveness of frame-marginal modulation for characterizing fine-grained facial dynamics.

\paragraph{Performance on Facial Expression Recognition.}
\begin{table*}[t]
\caption{
Comparison with extended state-of-the-art methods on DFEW, MAFW, and FERV39k.
Results on DFEW and MAFW are averaged over 5 folds.
\ifwithappendix
See Appendix~\ref{Appx:DetailedComparisonsonFacialEmotionRecognition} for detailed comparisons.
\fi
{\scriptsize(\textbf{Best} / \underline{Second} in video-only methods.
Pretrain: \textbf{U}nsupervised and \textbf{S}upervised.
Modality: \textbf{V}ideo, \textbf{I}mage, \textbf{A}udio, and \textbf{L}anguage.)}
}
\label{tab:Main-facial-video-only}

\centering
\scriptsize
\setlength{\tabcolsep}{4pt}
\renewcommand{\arraystretch}{1.05}

\resizebox{\linewidth}{!}{
\begin{tabular}{lcccccccccc}
\toprule

\multirow{2}{*}{\textbf{Method}}
& \multirow{2}{*}{\textbf{Pretrain}}
& \multirow{2}{*}{\textbf{Modality}}
& \multirow{2}{*}{\makecell{\textbf{Face}\\\textbf{Align.}}}
& \multirow{2}{*}{\makecell{\textbf{Params}\\\textbf{(M)}}}
& \multicolumn{2}{c}{\textbf{DFEW}}
& \multicolumn{2}{c}{\textbf{MAFW}}
& \multicolumn{2}{c}{\textbf{FERV39k}} \\

\cmidrule(lr){6-7}
\cmidrule(lr){8-9}
\cmidrule(lr){10-11}

& & & &
& {\scriptsize UAR} & {\scriptsize WAR}
& {\scriptsize UAR} & {\scriptsize WAR}
& {\scriptsize UAR} & {\scriptsize WAR} \\

\midrule

R(2+1)D-18\,\cite{avcaffe}
&  & V &  & 31
& 42.79 & 53.22
& -- & --
& -- & -- \\

C3D\,\cite{tran2015learning}
&  & V &  & 79
& 42.74 & 53.54
& 31.17 & 42.25
& -- & -- \\

3D ResNet-18\,\cite{hara2018can}
&  & V &  & 33
& 46.52 & 58.27
& -- & --
& -- & -- \\

IAL\,\cite{li2023intensity}
& S & V &  & 19
& 55.71 & 69.24
& -- & --
& 35.82 & 48.54 \\

M3DFEL\,\cite{wang2023rethinking}
& S & V &  &
& 56.10 & 69.25
& -- & --
& 35.94 & 47.67 \\

Former-DFER\,\cite{zhao2021former}
& S & V &  & 18
& 53.69 & 65.70
& 31.16 & 43.27
& 37.20 & 46.85 \\

NR-DFERNet\,\cite{li2022nr}
& S & V &  &
& 54.21 & 68.19
& -- & --
& 33.99 & 45.97 \\

STT\,\cite{ma2022spatio}
& S & V &  &
& 54.58 & 66.65
& -- & --
& 37.76 & 48.11 \\

LOGO-Former\,\cite{ma2023logo}
& S & V &  &
& 54.21 & 66.98
& -- & --
& 38.22 & 48.13 \\

MoCo\,\cite{he2020momentum}
& U & V &  & 32
& 53.47 & 67.45
& -- & --
& -- & -- \\

VideoMAE$^{*}$\,\cite{tong2022videomae}
& U & V & \checkmark & 86
& 63.60 & 74.60
& 40.87 & 53.51
& 43.33 & 52.39 \\

Video Swin-B\,\cite{liu2022video_swin}
& S & V &  & 88
& 59.38 & 71.90
& -- & --
& -- & -- \\

MAE-DFER\,\cite{sun2023mae}
& U & V & \checkmark & 85
& 63.41 & 74.43
& 41.62 & 54.31
& 43.12 & 52.07 \\

SVFAP\,\cite{sun2024svfap}
& U & V & \checkmark & 78
& 62.83 & 74.27
& 41.19 & 54.28
& 42.14 & 52.29 \\

\midrule

\rowcolor[rgb]{0.90,0.95,1.0}
ViT-B Exact
& U & V & \textcolor{red}{\textbf{$\times$}} & 86
& 65.29 & 75.86
& 45.24 & 58.69
& -- & -- \\

\rowcolor[rgb]{0.90,0.95,1.0}
ViT-B FlashLite
& U & V & \textcolor{red}{\textbf{$\times$}} & 86
& 65.11 & 76.25
& 45.48 & 58.38
& 43.81 & 53.80 \\

\rowcolor[rgb]{0.90,0.95,1.0}
ViT-L FlashLite
& U & V & \textcolor{red}{\textbf{$\times$}} & 305
& \underline{67.75} & \underline{77.32}
& \underline{46.80} & \underline{60.04}
& \underline{44.41} & \underline{54.62} \\

\rowcolor[rgb]{0.90,0.95,1.0}
ViT-H FlashLite
& U & V & \textcolor{red}{\textbf{$\times$}} & 521
& \textbf{68.25} & \textbf{78.24}
& \textbf{48.22} & \textbf{61.45}
& \textbf{45.12} & \textbf{55.64} \\

\midrule

\textcolor{gray!90}{MMA-DFER\,\cite{chumachenko2024mma}}
& \textcolor{gray!90}{U}
& \textcolor{gray!90}{V{+}A}
& \textcolor{gray!90}{\checkmark}
& \textcolor{gray!90}{}
& \textcolor{gray!90}{66.85}
& \textcolor{gray!90}{77.43}
& \textcolor{gray!90}{44.25}
& \textcolor{gray!90}{58.45}
& \textcolor{gray!90}{--}
& \textcolor{gray!90}{--} \\

\textcolor{gray!90}{HiCMAE\,\cite{sun2024hicmae}}
& \textcolor{gray!90}{U}
& \textcolor{gray!90}{V{+}A}
& \textcolor{gray!90}{\checkmark}
& \textcolor{gray!90}{81}
& \textcolor{gray!90}{63.76}
& \textcolor{gray!90}{75.01}
& \textcolor{gray!90}{42.65}
& \textcolor{gray!90}{56.17}
& \textcolor{gray!90}{--}
& \textcolor{gray!90}{--} \\

\textcolor{gray!90}{AVF-MAE++-B\,\cite{wu2025avf}}
& \textcolor{gray!90}{U+S}
& \textcolor{gray!90}{V{+}A}
& \textcolor{gray!90}{\checkmark}
& \textcolor{gray!90}{169}
& \textcolor{gray!90}{63.74}
& \textcolor{gray!90}{75.42}
& \textcolor{gray!90}{43.10}
& \textcolor{gray!90}{57.50}
& \textcolor{gray!90}{--}
& \textcolor{gray!90}{--} \\

\textcolor{gray!90}{AVF-MAE++-L\,\cite{wu2025avf}}
& \textcolor{gray!90}{U+S}
& \textcolor{gray!90}{V{+}A}
& \textcolor{gray!90}{\checkmark}
& \textcolor{gray!90}{303}
& \textcolor{gray!90}{65.14}
& \textcolor{gray!90}{76.24}
& \textcolor{gray!90}{45.36}
& \textcolor{gray!90}{59.13}
& \textcolor{gray!90}{--}
& \textcolor{gray!90}{--} \\

\textcolor{gray!90}{AVF-MAE++-H\,\cite{wu2025avf}}
& \textcolor{gray!90}{U+S}
& \textcolor{gray!90}{V{+}A}
& \textcolor{gray!90}{\checkmark}
& \textcolor{gray!90}{521}
& \textcolor{gray!90}{66.88}
& \textcolor{gray!90}{77.45}
& \textcolor{gray!90}{46.05}
& \textcolor{gray!90}{60.24}
& \textcolor{gray!90}{--}
& \textcolor{gray!90}{--} \\

\midrule

\textcolor{gray!90}{CLIPER\,\cite{li2024cliper}}
& \textcolor{gray!90}{U}
& \textcolor{gray!90}{V{+}L}
& \textcolor{gray!90}{\checkmark}
& \textcolor{gray!90}{}
& \textcolor{gray!90}{57.56}
& \textcolor{gray!90}{70.84}
& \textcolor{gray!90}{--}
& \textcolor{gray!90}{--}
& \textcolor{gray!90}{41.23}
& \textcolor{gray!90}{51.34} \\

\textcolor{gray!90}{DFER-CLIP\,\cite{zhao2023prompting}}
& \textcolor{gray!90}{U}
& \textcolor{gray!90}{V{+}L}
& \textcolor{gray!90}{\checkmark}
& \textcolor{gray!90}{153}
& \textcolor{gray!90}{59.61}
& \textcolor{gray!90}{71.25}
& \textcolor{gray!90}{39.89}
& \textcolor{gray!90}{52.55}
& \textcolor{gray!90}{41.27}
& \textcolor{gray!90}{51.65} \\

\textcolor{gray!90}{EmoCLIP\,\cite{foteinopoulou2024emoclip}}
& \textcolor{gray!90}{U}
& \textcolor{gray!90}{V{+}L}
& \textcolor{gray!90}{}
& \textcolor{gray!90}{}
& \textcolor{gray!90}{58.04}
& \textcolor{gray!90}{62.12}
& \textcolor{gray!90}{31.41}
& \textcolor{gray!90}{36.18}
& \textcolor{gray!90}{34.24}
& \textcolor{gray!90}{41.46} \\

\midrule

\textcolor{gray!90}{GPT-4o\,\cite{hurst2024gpt}}
& \textcolor{gray!90}{}
& \textcolor{gray!90}{V{+}A{+}L}
& \textcolor{gray!90}{}
& \textcolor{gray!90}{}
& \textcolor{gray!90}{50.57}
& \textcolor{gray!90}{57.19}
& \textcolor{gray!90}{38.29}
& \textcolor{gray!90}{48.82}
& \textcolor{gray!90}{--}
& \textcolor{gray!90}{--} \\

\textcolor{gray!90}{Emotion-LLaMA\,\cite{cheng2024emotion}}
& \textcolor{gray!90}{U}
& \textcolor{gray!90}{V{+}A{+}L}
& \textcolor{gray!90}{}
& \textcolor{gray!90}{}
& \textcolor{gray!90}{64.21}
& \textcolor{gray!90}{77.06}
& \textcolor{gray!90}{--}
& \textcolor{gray!90}{--}
& \textcolor{gray!90}{--}
& \textcolor{gray!90}{--} \\

\bottomrule
\end{tabular}
}
\end{table*}
As shown in~\cref{tab:Main-facial-video-only}, MiRA achieves state-of-the-art performance within video-only methods and remains competitive with multimodal and hybrid pretraining approaches.
Audio–visual models rely on more extensive audio--visual pretraining, often with supervised objectives.
Other multimodal approaches leverage textual supervision or pretrained language knowledge to enrich facial semantics.
Despite relying solely on video input, our results suggest that effective modeling of facial dynamics remains critical for representing affective context.
This aligns with our modeling strategy, which suppresses nuisance global motion, reduces redundancy, and enhances sensitivity to localized intra-facial cues.
This, in turn, allows our approach to operate without task-specific preprocessing, unlike prior works that rely on heuristics (e.g., tight facial cropping or alignment) to constrain the model to intra-facial regions.
It further supports consistent generalization to large-scale and diverse data where alignment is impractical.
Finally, by improving computational efficiency, flashLite enables scaling to larger models, allowing us to extend from ViT-B to ViT-L and ViT-H and consistently obtain gains as model capacity increases.

\paragraph{Probing Pretrained Representations.}
To analyze the quality of pretrained representations, we perform $k$-NN probing under a leave-one-subject-out (LOSO) protocol, where each subject is used for testing and the others for training.
Unlike conventional FER, where large intra-class variation necessitates task-specific classification, SAMM~\cite{davison2016samm} and MMEW~\cite{ben2021video} are tightly face-aligned micro-expression datasets%
\ifwithappendix\ (see Appendix~\ref{ExamplesofMicro-ExpressionDatasets})\fi,
making recognition primarily dependent on highly fine-grained temporal dynamics.
This setting enables a direct evaluation of the intrinsic quality of pretrained representations.
Features from the frozen backbone are evaluated using cosine similarity-based $k$-NN, and \cref{tab:knn_topk_category} reports Top-1 accuracy averaged over all subjects for different $k$.
Our method consistently improves performance across datasets and $k$ values, indicating that the learned representations better capture subtle temporal dynamics.
\begin{table}[t]
\caption{Evaluation of representation quality via $k$-NN probing with frozen video features on micro-facial expression datasets, reported in Top-$k$ accuracy. {\scriptsize(\textbf{Best} / \underline{Second})}}
\centering
\resizebox{0.8\linewidth}{!}{
\scriptsize
\setlength{\tabcolsep}{7pt}
\renewcommand{\arraystretch}{1.05}
\begin{tabular}{c l ccc ccc}
\toprule
\multirow{2}{*}{\textbf{Model Family}}
& \multirow{2}{*}{\textbf{Method}}
& \multicolumn{3}{c}{\textbf{SAMM}} 
& \multicolumn{3}{c}{\textbf{MMEW}} \\
\cmidrule(lr{0.5em}){3-5} \cmidrule(lr{0.5em}){6-8}
& 
& \textbf{k = 1} & \textbf{k = 3} & \textbf{k = 5}
& \textbf{k = 1} & \textbf{k = 3} & \textbf{k = 5} \\

\midrule

\multirow{3}{*}{\makecell{\textbf{General}\\\textbf{Video}\\\textbf{Models}}}
 & VideoMAE   & 0.6876 & 0.7213 & 0.7270 & \underline{0.3124} & 0.2883 & 0.3074 \\
 & MGMAE      & \underline{0.7011} & \underline{0.7350} & \underline{0.7524} & 0.2881 & \underline{0.2903} & 0.2975 \\
 & AdaMAE     & 0.6276 & 0.6730 & 0.7126 & 0.1674 & 0.1895 & 0.2443 \\

\specialrule{0.6pt}{0.6pt}{0.6pt}

\multirow{3}{*}{\makecell{\textbf{FER}\\\textbf{Models}}}
 & SVFAP      & 0.6898 & 0.6963 & 0.7373 & 0.2684 & 0.2596 & 0.2710 \\
 & MAE-DFER   & 0.5897 & 0.6970 & 0.7568 & 0.3106 & 0.2845 & \textbf{0.3122} \\

 & \cellcolor{softblue!30}ViT-B FlashLite
 & \cellcolor{softblue!30}\textbf{0.7329}
 & \cellcolor{softblue!30}\textbf{0.7369}
 & \cellcolor{softblue!30}\textbf{0.7730}
 & \cellcolor{softblue!30}\textbf{0.3148}
 & \cellcolor{softblue!30}\textbf{0.3037}
 & \cellcolor{softblue!30}\underline{0.3118} \\

\bottomrule
\end{tabular}
}
\label{tab:knn_topk_category}
\end{table}

\paragraph{Qualitative Analysis of Learned Attention Allocation.}
\begin{figure*}[t]
\centering
\includegraphics[width=\textwidth]{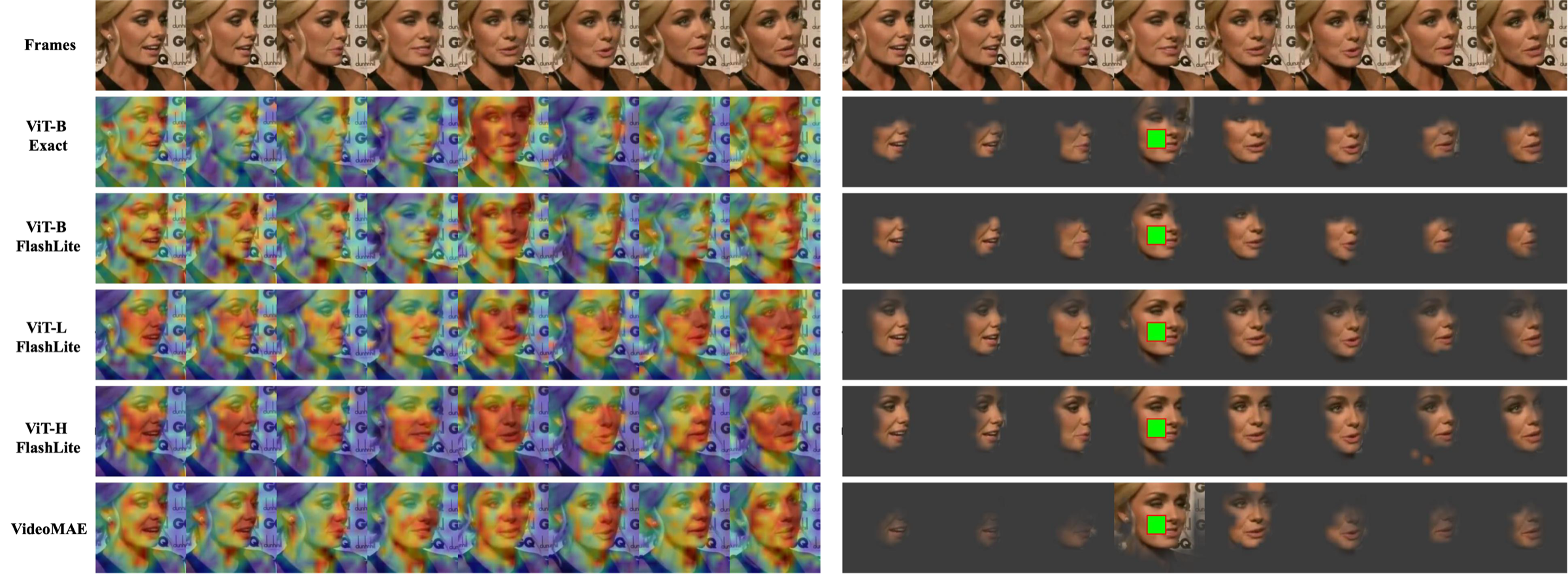}
\\[2mm]
\includegraphics[width=\textwidth]{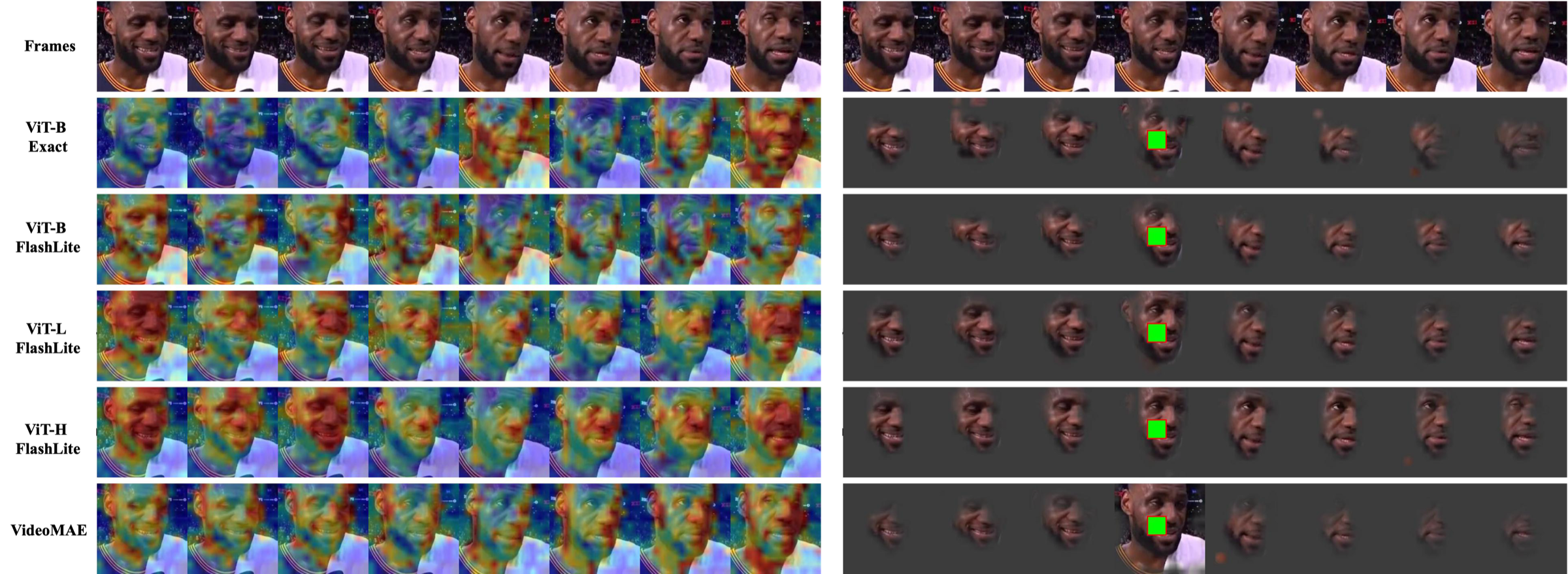}
\caption{
\textbf{Global and local attention visualizations during masked video reconstruction pretraining.}
Left: global attention visualizations obtained by modulating image visibility according to attention accumulated over the full spatio-temporal sequence.
Right: region-conditioned attention heatmaps associated with the selected facial region (green box), illustrating the interaction strength between the highlighted region and the remaining spatio-temporal locations across the video sequence.
}
\vspace{-6pt}
\label{fig:attention_visualization}
\end{figure*}
Because masked video reconstruction favors cues that most effectively reduce reconstruction error, pretrained attention tends to emphasize dominant appearance variations rather than subtle facial dynamics.
In the left panel of Fig.~\ref{fig:attention_visualization}, the VideoMAE baseline consistently emphasizes visually salient boundary regions induced by head-pose variation.
In the upper example, attention concentrates on the left cheek during head turning, whereas in the lower example it accumulates around outer facial contours under lateral head motion.
In contrast, MiRA focuses on a smaller subset of frames while distributing attention across multiple facial regions, yielding more face-centered attention patterns.
The region-conditioned attention maps further show how a selected facial region interacts with the remaining spatio-temporal locations.
VideoMAE attention remains largely confined to the selected frame, with substantial attention extending to surrounding background regions.
By contrast, MiRA establishes stronger interactions with facial regions in other frames, enabling complementary facial information to be aggregated across time.
\ifwithappendix
Additional attention visualizations are provided in Appendix~\ref{AttentionVisualizations}.
\fi

\section{Conclusion}
This work investigates learning facial dynamics from videos in scalable settings, where subtle expression cues are often obscured by temporally redundant content and dominant global motion.
We revisit self-attention and introduce a frame-marginal attention redistribution mechanism that promotes selective and localized spatio-temporal attention, together with an efficient approximation compatible with large-scale training.
Across both pretraining and downstream evaluations, the proposed approach consistently improves performance while preserving computational efficiency.
Future work will explore integrating frame-marginal attention redistribution into emerging video foundation models and multimodal pretraining frameworks.

\section*{Acknowledgements}
This work was supported by the French government through the France 2030 plan, managed by the National Research Agency (ANR) with the reference number “ANR-23-IACL-0001”;
by the research fund of Hanyang University (HY-202600000000788), Korea;
and by the Institute of Information \& Communications Technology Planning \& Evaluation (IITP)-Information Technology Research Center (ITRC) grant funded by the Korea government (Ministry of Science and ICT) (IITP-2026-RS-2023-00260098).
This work was also granted access to the HPC resources at IDRIS under the allocation 2026-[A010617132,
A0191016918] made by GENCI, France.
The authors would like to thank Rémi Lacroix for valuable technical support with the HPC system.


\bibliographystyle{splncs04}
\bibliography{main}

@String(PAMI  = {IEEE Trans. Pattern Anal. Mach. Intell.})

@String(CVPR  = {IEEE Conf. Comput. Vis. Pattern Recog.})

@String(ICCV  = {Int. Conf. Comput. Vis.})

@String(NeurIPS = {Adv. Neural Inform. Process. Syst.})

@String(ICML  = {Int. Conf. Mach. Learn.})

@String(ICLR  = {Int. Conf. Learn. Represent.})

@String(AAAI  = {AAAI})

@String(ICASSP=	{ICASSP})

@String(ICME  = {Int. Conf. Multimedia and Expo})

@String(ACMMM = {ACM Int. Conf. Multimedia})

@inproceedings{dao2022flashattention,
  author    = {Dao, Tri and Fu, Daniel Y. and Ermon, Stefano and Rudra, Atri and R{\'e}, Christopher},
  title     = {FlashAttention: Fast and Memory-Efficient Exact Attention with IO-Awareness},
  booktitle = NeurIPS,
  year      = {2022}
}

@inproceedings{dao2023flashattention2,
  title={Flashattention-2: Faster attention with better parallelism and work partitioning},
  author={Dao, Tri},
  booktitle=ICLR,
  volume={2024},
  pages={35549--35562},
  year={2024}
}

@article{polyak1992acceleration,
  author    = {Polyak, Boris T. and Juditsky, Anatoli B.},
  title     = {Acceleration of Stochastic Approximation by Averaging},
  journal = {SIAM Journal on Control and Optimization},
  volume    = {30},
  number    = {4},
  pages     = {838--855},
  year      = {1992}
}

@inproceedings{tarvainen2017mean,
  author    = {Tarvainen, Antti and Valpola, Harri},
  title     = {Mean Teachers are Better Role Models: Weight-Averaged Consistency Targets Improve Semi-Supervised Deep Learning Results},
  booktitle = NeurIPS,
  volume    = {30},
  year      = {2017}
}

@inproceedings{voxceleb2,
  author    = {Chung, Joon Son and Nagrani, Arsha and Zisserman, Andrew},
  title     = {VoxCeleb2: Deep Speaker Recognition},
  booktitle = {Proceedings of Interspeech},
  year      = {2018}
}

@inproceedings{dfew,
  title={Dfew: A large-scale database for recognizing dynamic facial expressions in the wild},
  author={Jiang, Xingxun and Zong, Yuan and Zheng, Wenming and Tang, Chuangao and Xia, Wanchuang and Lu, Cheng and Liu, Jiateng},
  booktitle=ACMMM,
  pages={2881--2889},
  year={2020}
}

@inproceedings{mafw,
  title={Mafw: A large-scale, multi-modal, compound affective database for dynamic facial expression recognition in the wild},
  author={Liu, Yuanyuan and Dai, Wei and Feng, Chuanxu and Wang, Wenbin and Yin, Guanghao and Zeng, Jiabei and Shan, Shiguang},
  booktitle=ACMMM,
  pages={24--32},
  year={2022}
}

@inproceedings{avcaffe,
  author    = {Sarkar, Priyanka and Posen, Aaron and Etemad, Ali},
  title     = {{AVCAffe}: A Large Scale Audio-Visual Dataset of Cognitive Load and Affect for Remote Work},
  booktitle = AAAI,
  pages     = {76--85},
  year      = {2023}
}

@inproceedings{vaswani2017attention,
  author    = {Vaswani, Ashish and Shazeer, Noam and Parmar, Niki and Uszkoreit, Jakob and Jones, Llion and Gomez, Aidan N. and Kaiser, {\L}ukasz and Polosukhin, Illia},
  title     = {Attention Is All You Need},
  booktitle = NeurIPS,
  volume    = {30},
  year      = {2017}
}

@article{devlin2019bert,
  author    = {Devlin, Jacob and Chang, Ming-Wei and Lee, Kenton and Toutanova, Kristina},
  title     = {BERT: Pre-training of deep bidirectional transformers for language understanding},
  journal = {NAACL-HLT},
  year      = {2019}
}

@inproceedings{he2022mae,
  author    = {He, Kaiming and Chen, Xinlei and Xie, Saining and Li, Yanghao and Doll{\'a}r, Piotr and Girshick, Ross},
  title     = {Masked autoencoders are scalable vision learners},
  booktitle = CVPR,
  year      = {2022}
}

@article{zhou2022ibot,
  title={iBOT: Image BERT Pre-Training with Online Tokenizer},
  author={Zhou, Jinghao and Wei, Chen and Wang, Huiyu and Shen, Wei and Xie, Cihang and Yuille, Alan and Kong, Tao},
  journal={International Conference on Learning Representations (ICLR)},
  year={2022}
}

@inproceedings{li2022uniformer,
  author={Li, Kunchang and Wang, Yali and Gao, Peng and Song, Guanglu and Liu, Yu and Li, Hongsheng and Qiao, Yu},
  title     = {Uniformer: Unified transformer for efficient spatial-temporal representation learning},
  booktitle = ICLR,
  year      = {2022}
}

@inproceedings{han2020cotraining,
  author    = {Han, Tengda and Xie, Weidi and Zisserman, Andrew},
  title     = {Self-Supervised Co-Training for Video Representation Learning},
  booktitle = NeurIPS,
  year      = {2020}
}

@inproceedings{qian2021spatiotemporal,
  author    = {Qian, Rui and Meng, Tianjian and Gong, Boqing and Yang, Ming-Hsuan and Wang, Huisheng and Belongie, Serge and Cui, Yin},
  title     = {Spatiotemporal Contrastive Video Representation Learning},
  booktitle = CVPR,
  pages     = {6964--6974},
  year      = {2021}
}

@inproceedings{alayrac2020xdc,
  author    = {Alwassel, Humam and Mahajan, Dhruv and Korbar, Bruno and Torresani, Lorenzo and Ghanem, Bernard and Tran, Du},
  title     = {Self-Supervised Learning by Cross-Modal Audio-Video Clustering},
  booktitle = NeurIPS,
  year      = {2020}
}

@article{dave2021tclr,
  title={Tclr: Temporal contrastive learning for video representation},
  author={Dave, Ishan and Gupta, Rohit and Rizve, Mamshad Nayeem and Shah, Mubarak},
  journal={Computer Vision and Image Understanding},
  volume={219},
  pages={103406},
  year={2022},
  publisher={Elsevier}
}

@article{bardes2024vjepa,
  author    = {Bardes, Adrien and Garrido, Quentin and Ponce, Jean and Chen, Xinlei and Rabbat, Michael and LeCun, Yann and Assran, Mahmoud and Ballas, Nicolas},
  title     = {Revisiting Feature Prediction for Learning Visual Representations from Video},
  journal = {arXiv preprint arXiv:2404.08471},
  year      = {2024}
}

@inproceedings{wei2022maskfeat,
  author    = {Wei, Chen and Fan, Haoqi and Xie, Saining and Wu, Chao-Yuan and Yuille, Alan and Feichtenhofer, Christoph},
  title     = {Masked Feature Prediction for Self-Supervised Visual Pre-Training},
  booktitle = CVPR,
  pages     = {14668--14678},
  year      = {2022}
}

@inproceedings{wang2022bevt,
  title={Bevt: Bert pretraining of video transformers},
  author={Wang, Rui and Chen, Dongdong and Wu, Zuxuan and Chen, Yinpeng and Dai, Xiyang and Liu, Mengchen and Jiang, Yu-Gang and Zhou, Luowei and Yuan, Lu},
  booktitle={2022 IEEE/CVF Conference on Computer Vision and Pattern Recognition (CVPR)},
  pages={14713--14723},
  year={2022},
  organization={IEEE}
}

@inproceedings{feichtenhofer2022maest,
  author    = {Feichtenhofer, Christoph and Fan, Haoqi and Li, Yanghao and He, Kaiming},
  title     = {Masked Autoencoders As Spatiotemporal Learners},
  booktitle = NeurIPS,
  volume    = {35},
  pages     = {35946--35958},
  year      = {2022}
}

@inproceedings{tong2022videomae,
  author    = {Tong, Zhan and Song, Yibing and Wang, Jue and Wang, Limin},
  title     = {VideoMAE: Masked Autoencoders are Data-Efficient Learners for Self-Supervised Video Pre-Training},
  booktitle = NeurIPS,
  volume    = {35},
  pages     = {10078--10093},
  year      = {2022}
}

@inproceedings{rai2021cocon,
  author    = {Rai, Nishant and Adeli, Ehsan and Lee, Kuan-Hui and Gaidon, Adrien and Niebles, Juan Carlos},
  title     = {Cocon: Cooperative-contrastive learning},
  booktitle = CVPR,
  pages     = {3384--3393},
  year      = {2021}
}

@article{sun2024hicmae,
  author    = {Sun, Licai and Lian, Zheng and Liu, Bin and Tao, Jianhua},
  title     = {Hicmae: Hierarchical contrastive masked autoencoder for self-supervised audio-visual emotion recognition},
  journal = {Information Fusion},
  volume    = {108},
  pages     = {102382},
  year      = {2024},
  publisher = {Elsevier}
}

@inproceedings{wu2025avf,
  author    = {Wu, Xuecheng and Sun, Heli and Wang, Yifan and Nie, Jiayu and Zhang, Jie and Wang, Yabing and Xue, Junxiao and He, Liang},
  title     = {AVF-MAE++: Scaling Affective Video Facial Masked Autoencoders via Efficient Audio-Visual Self-Supervised Learning},
  booktitle = CVPR,
  pages     = {9142--9153},
  year      = {2025}
}

@inproceedings{wang2023videomae,
  author    = {Wang, Limin and Huang, Bingkun and Zhao, Zhiyu and Tong, Zhan and He, Yinan and Wang, Yi and Wang, Yali and Qiao, Yu},
  title     = {Videomae v2: Scaling video masked autoencoders with dual masking},
  booktitle = CVPR,
  pages     = {14549--14560},
  year      = {2023}
}

@inproceedings{bandara2023adamae,
  author    = {Bandara, Wele Gedara Chaminda and Patel, Naman and Gholami, Ali and Nikkhah, Mehdi and Agrawal, Motilal and Patel, Vishal M},
  title     = {Adamae: Adaptive masking for efficient spatiotemporal learning with masked autoencoders},
  booktitle = CVPR,
  pages     = {14507--14517},
  year      = {2023}
}

@inproceedings{huang2023mgmae,
  author    = {Huang, Bingkun and Zhao, Zhiyu and Zhang, Guozhen and Qiao, Yu and Wang, Limin},
  title     = {Mgmae: Motion guided masking for video masked autoencoding},
  booktitle = ICCV,
  pages     = {13493--13504},
  year      = {2023}
}

@inproceedings{hwang2022everest,
  author    = {Hwang, Sunil and Yoon, Jaehong and Lee, Youngwan and Hwang, Sung Ju},
  title     = {EVEREST: Efficient Masked Video Autoencoder by Removing Redundant Spatiotemporal Tokens},
  booktitle = ICML,
  year      = {2024}
}

@article{chen2024unilearn,
  author    = {Chen, Yin and Li, Jia and Zhang, Yu and Hu, Zhenzhen and Shan, Shiguang and Wang, Meng and Hong, Richang},
  title     = {UniLearn: Enhancing Dynamic Facial Expression Recognition through Unified Pre-Training and Fine-Tuning on Images and Videos},
  journal = {arXiv e-prints},
  pages     = {arXiv:2409.06154v1 },
  year      = {2024}
}

@article{sun2024svfap,
  author    = {Sun, Licai and Lian, Zheng and Wang, Kexin and He, Yu and Xu, Mingyu and Sun, Haiyang and Liu, Bin and Tao, Jianhua},
  title     = {Svfap: Self-supervised video facial affect perceiver},
  journal = {IEEE Transactions on Affective Computing},
  year      = {2024},
  publisher = {IEEE}
}

@inproceedings{sun2023mae,
  author    = {Sun, Licai and Lian, Zheng and Liu, Bin and Tao, Jianhua},
  title     = {Mae-dfer: Efficient masked autoencoder for self-supervised dynamic facial expression recognition},
  booktitle = ACMMM,
  pages     = {6110--6121},
  year      = {2023}
}

@inproceedings{chumachenko2024mma,
  author    = {Chumachenko, Kateryna and Iosifidis, Alexandros and Gabbouj, Moncef},
  title     = {MMA-DFER: MultiModal Adaptation of unimodal models for Dynamic Facial Expression Recognition in-the-wild},
  booktitle = CVPR,
  pages     = {4673--4682},
  year      = {2024}
}

@inproceedings{bertasius21a,
  author    = {Bertasius, Gedas and Wang, Heng and Torresani, Lorenzo},
  title     = {Is Space-Time Attention All You Need for Video Understanding?},
  booktitle = ICML,
  year      = {2021}
}

@inproceedings{li2021mvitv2,
  author    = {Li, Yanghao and Wu, Chao-Yuan and Fan, Haoqi and Mangalam, Karttikeya and Xiong, Bo and Malik, Jitendra and Feichtenhofer, Christoph},
  title     = {MViTv2: Improved Multiscale Vision Transformers for Classification and Detection},
  booktitle = CVPR,
  pages     = {4804--4814},
  year      = {2022}
}

@article{oquab2023dinov2,
  author    = {Oquab, Maxime and Darcet, Timoth{\'e}e and Moutakanni, Theo and Vo, Huy V and Szafraniec, Marc and Khalidov, Vasil and Fernandez, Pierre and Haziza, Daniel and Massa, Francisco and El-Nouby, Alaaeldin and Assran, Mahmoud and Ballas, Nicolas and Galuba, Wojciech and Howes, Russell and Huang, Po-Yao and Li, Shang-Wen and Misra, Ishan and Rabbat, Michael and Sharma, Vasu and Synnaeve, Gabriel and Xu, Hu and J{\'e}gou, Herv{\'e} and Mairal, Julien and Schmid, Cordelia and Bojanowski, Piotr},
  title     = {DINOv2: Learning Robust Visual Features without Supervision},
  journal = {arXiv preprint arXiv:2304.07193},
  year      = {2023}
}

@inproceedings{zhang2021vidtr,
  title={Vidtr: Video transformer without convolutions},
  author={Zhang, Yanyi and Li, Xinyu and Liu, Chunhui and Shuai, Bing and Zhu, Yi and Brattoli, Biagio and Chen, Hao and Marsic, Ivan and Tighe, Joseph},
  booktitle= ICCV,
  pages={13557--13567},
  year={2021},
}

@inproceedings{gupta2023siammae,
  title={Siamese masked autoencoders},
  author={Gupta, Agrim and Wu, Jiajun and Deng, Jia and Li, Fei-Fei},
  booktitle=NeurIPS,
  volume={36},
  pages={40676--40693},
  year={2023}
}

@inproceedings{wu2023dropmae,
  author    = {Wu, Qiangqiang and Yang, Tianyu and Liu, Ziquan and Wu, Baoyuan and Shan, Ying and Chan, Antoni B.},
  title     = {DropMAE: Masked Autoencoders with Spatial-Attention Dropout for Tracking Tasks},
  booktitle = CVPR,
  pages     = {14561--14570},
  year      = {2023}
}

@inproceedings{zhang2024mart,
  author    = {Zhang, Zhicheng and Zhao, Pancheng and Park, Eunil and Yang, Jufeng},
  title     = {Mart: Masked affective representation learning via masked temporal distribution distillation},
  booktitle = CVPR,
  pages     = {12830--12840},
  year      = {2024}
}

@article{ma2022spatio,
  author    = {Ma, Fuyan and Sun, Bin and Li, Shutao},
  title     = {Spatio-temporal transformer for dynamic facial expression recognition in the wild},
  journal = {arXiv preprint arXiv:2205.04749},
  year      = {2022}
}

@article{liu2025robust,
  author    = {Liu, Feng and Wang, Hanyang and Shen, Siyuan},
  title     = {Robust Dynamic Facial Expression Recognition},
  journal = {IEEE Transactions on Biometrics, Behavior, and Identity Science},
  year      = {2025},
  publisher = {IEEE}
}

@inproceedings{li2023intensity,
  author    = {Li, Hanting and Niu, Hongjing and Zhu, Zhaoqing and Zhao, Feng},
  title     = {Intensity-aware loss for dynamic facial expression recognition in the wild},
  booktitle = AAAI,
  volume    = {37},
  pages     = {67--75},
  year      = {2023}
}

@article{liu2022clip,
  author    = {Liu, Yuanyuan and Feng, Chuanxu and Yuan, Xiaohui and Zhou, Lin and Wang, Wenbin and Qin, Jie and Luo, Zhongwen},
  title     = {Clip-aware expressive feature learning for video-based facial expression recognition},
  journal = {Information Sciences},
  volume    = {598},
  pages     = {182--195},
  year      = {2022},
  publisher = {Elsevier}
}

@inproceedings{arnab2021vivit,
  author    = {Arnab, Anurag and Dehghani, Mostafa and Heigold, Georg and Sun, Chen and Lu{\v{c}}i{\'c}, Mario and Schmid, Cordelia},
  title     = {ViViT: A Video Vision Transformer},
  booktitle = ICCV,
  pages     = {6836--6846},
  year      = {2021}
}

@inproceedings{fan2021multiscale,
  author={Fan, Haoqi and Xiong, Bo and Mangalam, Karttikeya and Li, Yanghao and Yan, Zhicheng and Malik, Jitendra and Feichtenhofer, Christoph},
  title={Multiscale vision transformers},
  booktitle=ICCV,
  pages={6824--6835},
  year={2021}
}

@article{dosovitskiy2020image,
  title={An image is worth 16x16 words: Transformers for image recognition at scale},
  author={Dosovitskiy, Alexey and Beyer, Lucas and Kolesnikov, Alexander and Weissenborn, Dirk and Zhai, Xiaohua and Unterthiner, Thomas and Dehghani, Mostafa and Minderer, Matthias and Heigold, Georg and Gelly, Sylvain and others},
  journal={arXiv preprint arXiv:2010.11929},
  year={2020}
}

@inproceedings{cuturi2013sinkhorn,
  author={Cuturi, Marco},
  title={Sinkhorn distances: Lightspeed computation of optimal transport},
  booktitle=NeurIPS,
  volume={26},
  year={2013}
}

@inproceedings{cheng2024emotion,
  title={Emotion-llama: Multimodal emotion recognition and reasoning with instruction tuning},
  author={Cheng, Zebang and Cheng, Zhi-Qi and He, Jun-Yan and Sun, Jingdong and Wang, Kai and Lin, Yuxiang and Lian, Zheng and Peng, Xiaojiang and Hauptmann, Alexander G},
  booktitle=NeurIPS,
  volume={37},
  pages={110805--110853},
  year={2024}
}

@article{hurst2024gpt,
  title={Gpt-4o system card},
  author={Hurst, Aaron and Lerer, Adam and Goucher, Adam P and Perelman, Adam and Ramesh, Aditya and Clark, Aidan and Ostrow, AJ and Welihinda, Akila and Hayes, Alan and Radford, Alec and others},
  journal={arXiv preprint arXiv:2410.21276},
  year={2024}
}

@inproceedings{tran2015learning,
  title={Learning spatiotemporal features with 3d convolutional networks},
  author={Tran, Du and Bourdev, Lubomir and Fergus, Rob and Torresani, Lorenzo and Paluri, Manohar},
  booktitle=ICCV,
  pages={4489--4497},
  year={2015}
}

@article{zhao2023prompting,
  title={Prompting visual-language models for dynamic facial expression recognition},
  author={Zhao, Zengqun and Patras, Ioannis},
  journal={arXiv preprint arXiv:2308.13382},
  year={2023}
}

@inproceedings{hara2018can,
  title={Can spatiotemporal 3d cnns retrace the history of 2d cnns and imagenet?},
  author={Hara, Kensho and Kataoka, Hirokatsu and Satoh, Yutaka},
  booktitle=CVPR,
  pages={6546--6555},
  year={2018}
}

@inproceedings{he2020momentum,
  title={Momentum contrast for unsupervised visual representation learning},
  author={He, Kaiming and Fan, Haoqi and Wu, Yuxin and Xie, Saining and Girshick, Ross},
  booktitle=CVPR,
  pages={9729--9738},
  year={2020}
}

@inproceedings{wang2023rethinking,
  title={Rethinking the learning paradigm for dynamic facial expression recognition},
  author={Wang, Hanyang and Li, Bo and Wu, Shuang and Shen, Siyuan and Liu, Feng and Ding, Shouhong and Zhou, Aimin},
  booktitle=CVPR,
  pages={17958--17968},
  year={2023}
}

@inproceedings{liu2022video_swin,
  title={Video swin transformer},
  author={Liu, Ze and Ning, Jia and Cao, Yue and Wei, Yixuan and Zhang, Zheng and Lin, Stephen and Hu, Han},
  booktitle=CVPR,
  pages={3202--3211},
  year={2022}
}

@inproceedings{pei2024videomac,
  title={Videomac: Video masked autoencoders meet convnets},
  author={Pei, Gensheng and Chen, Tao and Jiang, Xiruo and Liu, Huafeng and Sun, Zeren and Yao, Yazhou},
  booktitle=CVPR,
  pages={22733--22743},
  year={2024}
}

@article{gundavarapu2024extending,
  title={Extending video masked autoencoders to 128 frames},
  author={Gundavarapu, Nitesh B and Friedman, Luke and Goyal, Raghav and Hegde, Chaitra and Agustsson, Eirikur and Waghmare, Sagar and Sirotenko, Mikhail and Yang, Ming-Hsuan and Weyand, Tobias and Gong, Boqing and others},
  journal=NeurIPS,
  volume={37},
  pages={121376--121400},
  year={2024}
}

@article{davison2016samm,
  title={Samm: A spontaneous micro-facial movement dataset},
  author={Davison, Adrian K and Lansley, Cliff and Costen, Nicholas and Tan, Kevin and Yap, Moi Hoon},
  journal={IEEE transactions on affective computing},
  volume={9},
  number={1},
  pages={116--129},
  year={2016},
  publisher={IEEE}
}

@article{ben2021video,
  title={Video-based facial micro-expression analysis: A survey of datasets, features and algorithms},
  author={Ben, Xianye and Ren, Yi and Zhang, Junping and Wang, Su-Jing and Kpalma, Kidiyo and Meng, Weixiao and Liu, Yong-Jin},
  journal=PAMI,
  volume={44},
  number={9},
  pages={5826--5846},
  year={2021},
  publisher={IEEE}
}

@inproceedings{wang2022ferv39k,
  title={Ferv39k: A large-scale multi-scene dataset for facial expression recognition in videos},
  author={Wang, Yan and Sun, Yixuan and Huang, Yiwen and Liu, Zhongying and Gao, Shuyong and Zhang, Wei and Ge, Weifeng and Zhang, Wenqiang},
  booktitle=CVPR,
  pages={20922--20931},
  year={2022}
}

@inproceedings{foteinopoulou2024emoclip,
  title={Emoclip: A vision-language method for zero-shot video facial expression recognition},
  author={Foteinopoulou, Niki Maria and Patras, Ioannis},
  booktitle={2024 IEEE 18th international conference on automatic face and gesture recognition (FG)},
  pages={1--10},
  year={2024},
  organization={IEEE}
}

@inproceedings{zhao2021former,
  title={Former-dfer: Dynamic facial expression recognition transformer},
  author={Zhao, Zengqun and Liu, Qingshan},
  booktitle=ACMMM,
  pages={1553--1561},
  year={2021}
}

@article{li2022nr,
  title={Nr-dfernet: Noise-robust network for dynamic facial expression recognition},
  author={Li, Hanting and Sui, Mingzhe and Zhu, Zhaoqing and others},
  journal={arXiv preprint arXiv:2206.04975},
  year={2022}
}

@inproceedings{ma2023logo,
  title={Logo-former: Local-global spatio-temporal transformer for dynamic facial expression recognition},
  author={Ma, Fuyan and Sun, Bin and Li, Shutao},
  booktitle=ICASSP,
  pages={1--5},
  year={2023},
  organization={IEEE}
}

@inproceedings{li2024cliper,
  title={Cliper: A unified vision-language framework for in-the-wild facial expression recognition},
  author={Li, Hanting and Niu, Hongjing and Zhu, Zhaoqing and Zhao, Feng},
  booktitle={2024 IEEE international conference on multimedia and expo (ICME)},
  pages={1--6},
  year={2024},
  organization={IEEE}
}

\ifwithappendix

\clearpage
\appendix
\clearpage

\section{Appendix}

\subsection{Preliminary on Video Transformers}
\label{Appx:Preliminary}
\begin{wrapfigure}{r}{0.22\linewidth}
    \vspace{-20pt}
    \hspace{-14pt} 
    \centering
    \includegraphics[width=\linewidth]{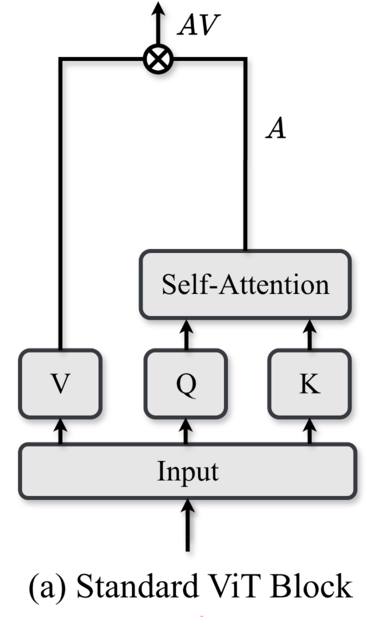}
    \caption{\textbf{Transformer block.}}
    \label{fig:transformer_block}
    \vspace{-16pt}
\end{wrapfigure}
In recent ViT-based video representation learning methods, an input video is divided into non-overlapping tubelet patches~\cite{arnab2021vivit}.
Let $T$, $N_{h}$, and $N_{w}$ denote the numbers of splits along the temporal, height, and width dimensions, respectively.
The total number of tubelets is then $L = T \times N$, where $N = N_{h} \times N_{w}$.
Each tubelet is embedded into a vector $x \in \mathbb{R}^{d}$ via a patch embedding module, and stacking all embeddings forms $X \in \mathbb{R}^{L \times d}$.
The embedding sequence $X$ is then processed by a stack of transformer blocks, as illustrated in \cref{fig:transformer_block}, to encode spatio-temporal representations.
Within each block, the standard self-attention module produces queries, keys, and values defined as
$Q = XW_{Q} \in \mathbb{R}^{L \times d_{K}}$,
$K = XW_{K} \in \mathbb{R}^{L \times d_{K}}$, and
$V = XW_{V} \in \mathbb{R}^{L \times d_{V}}$,
where $W_{Q}, W_{K} \in \mathbb{R}^{d \times d_{K}}$ and $W_{V} \in \mathbb{R}^{d \times d_{V}}$, and $d_{K}$ and $d_{V}$ denote the query/key and value dimensions, respectively.
The scaled dot-product attention is computed as
$A = \mathtt{softmax}\!\left(QK^{\top} / \sqrt{d_{K}}\right) \in \mathbb{R}^{L \times L}$,
where $A_{q,k}$ denotes the $(q,k)$-th element (query--key order).
The output embeddings are obtained as $Y = AV$, where all tokens (corresponding to tubelets) are treated uniformly except for differences in positional encodings.
Note that $T$ does not necessarily correspond to the original number of video frames, as it depends on the temporal stride of the tokenizer that constructs each tubelet (\eg, a single tubelet may span multiple consecutive frames)~\cite{arnab2021vivit,fan2021multiscale,liu2022video_swin}.

\subsection{Approximation Rationale for FlashLite Mode}
\label{Appx:FlashLiteRationale}
In self-attention, the total attention mass that a frame receives as keys is correlated with the magnitude of its key representations through the query--key similarity.
The frame-sum energy defined in~\eqref{eq:kenergy}--\eqref{eq:kstats_all} thus serves as a practical surrogate for the true confidence $c_t$.
Similarly, the inverse entropy of the normalized key energy distribution, $\tilde{H}_t$, reflects whether attention is likely to be spatially diffuse or concentrated within a frame, mirroring the role of $H_t$.
Consequently, the flashLite scaling factors $\alpha_{t}^{\text{FL}}$ empirically exhibit redistribution behavior similar to their exact counterparts without requiring access to the attention map $A$.

In practice, however, flashLite is not strictly equivalent to the exact formulation, as the query-wise renormalization in~\eqref{eq:reweight} is replaced by framewise scaling applied as log-biases prior to the softmax.
Nevertheless, because the scaling factors are derived from frame-level statistics aggregated across attention heads, the approximation remains stable during training while preserving the intended framewise modulation.
Moreover, shifting the reweighting to the pre-softmax logits preserves the fused streaming pipeline of FlashAttention and avoids materializing the $L \times L$ attention map, thereby thereby preserving the linear-memory I/O efficiency of standard FlashAttention during both training and inference.

\subsection{Implementation Details}
\label{Appx:ImplementationDetails}
\paragraph{Backbone Context.}
We adopt VideoMAE~\cite{tong2022videomae} as the baseline under the pretrain--finetune paradigm.
It learns video representations via self-supervised pretraining on unlabeled videos and adapts the same ViT backbone to downstream tasks by supervised finetuning.
In pretraining, the model masks $90\%$ of spatio-temporal tubelet patches and reconstructs the masked tokens from the visible subset, using a frame-shared mask to preserve spatial alignment and maintain consistent frame-wise token indexing.
For classification, we finetune the encoder with a lightweight task head.
We integrate FAR into each self-attention block as a drop-in module that performs attention redistribution and framewise modulation; it adds no learnable parameters and preserves the original transformer pipeline while adaptively reweighting attention.

\paragraph{Stabilizing Confidence Score with EMA.}
Compared to the inverse entropy $H_t$, the confidence score $c_t$ tends to exhibit higher variance, as it aggregates raw attention magnitudes across all queries.
As a result, its value may fluctuate significantly depending on batch composition.
To mitigate this variability during pretraining, we maintain a dataset-level prior by applying an exponential moving average (EMA)~\cite{polyak1992acceleration, tarvainen2017mean} to the batch-averaged confidence values:
\begin{equation}
    \bar{c}_t \leftarrow m\,\bar{c}_{t} + (1 - m)\,c^{\mathrm{batch}}_{t},
    \quad m \in [0, 1),
    \label{eq:ema_conf}
\end{equation}
where $m$ denotes the momentum coefficient controlling the trade-off between the historical estimate $\bar{c}_t$ and the current batch statistic $c^{\mathrm{batch}}_t$.
This EMA-based smoothing provides a more stable estimate of frame-level importance across training iterations, which is particularly beneficial during large-scale pretraining.
In contrast, during finetuning, we use instance-level statistics without EMA, allowing the model to adapt more flexibly to sample-specific dynamics and task-dependent variations.

\paragraph{FlashLite Temperature.}
For flashLite, the temperature parameter $\tau$ in~\eqref{eq:flashlite_prior} is fixed to $1.7$ across all datasets and model scales.
The parameter controls the sharpness of the frame prior distribution derived from the standardized frame importance scores.
Larger values produce more selective framewise modulation by emphasizing high-response frames, whereas smaller values yield smoother redistribution across frames.
We empirically found $\tau=1.7$ to provide a stable balance between selective emphasis and optimization stability.

\paragraph{Numerical Safeguards.}
To ensure numerical stability, small constants ($\epsilon=10^{-6}$) are added to denominators and logarithms in \eqref{eq:inv_entropy} and \eqref{eq:kstats_all}.
A smoothing factor $\lambda_{\pi}=0.1$ in \eqref{eq:prior} mixes a fraction of the uniform distribution into the prior to avoid frame collapse.
Additionally, a uniform floor mass ($\epsilon_{\mathrm{floor}}=10^{-3}$) is added before row-wise renormalization to prevent query distributions from collapsing to zero.

\begin{algorithm*}[h]
\caption{Pseudocode of \textbf{Exact} and \textbf{FlashLite} modes in a transformer block.}
\label{alg:our_modes}
\centering

\lstset{
  basicstyle=\ttfamily\bfseries\fontsize{6.3}{7.2}\selectfont,
  breaklines=true,
  columns=fullflexible,
  keepspaces=true,
  showstringspaces=false,
  commentstyle=\color{green!50!black}\normalfont,
  morecomment=[l]{\#}
}

\begin{minipage}[t]{0.48\textwidth}
\textbf{Exact mode}

\begin{lstlisting}
X: video tokens [B,T,N,d], L=T*N
    # B: batch size, T: frames
    # N: tokens/frame, d: embed dim
    # Hh: heads, dh: head dim

for X in loader:

    q,k,v = QKV(X)                  # [B,Hh,L,dh]

    A_orig = softmax(q @ k.T / sqrt(dh))
    A_bar = MeanOverHeads(A_orig)   # [B,L,L]

    c = FrameConfidence(A_bar)      # [B,T]
    p = FrameTokenDistribution(A_bar)   # [B,T,N]
    H = IntraFrameConcentration(p)      # [B,T]

    if pretrain:
        c = EMA(c)

    f = w_con*c + w_ent*H               # [B,T]
    f = MinMaxNorm(f)
    pi = PowerSoftmax(f, beta)          # [B,T]
    pi = (1-lambda_pi)*pi + lambda_pi/T

    alpha = clip(pi/(c+eps), alpha_min, alpha_max)                                    # [B,T]

    A_update = ScaleKeysByFrame(A_orig, alpha)             
    A_update = (1-eps_floor)*A_update + eps_floor/L
    A_update = RowNormalize(A_update)
    A_final = (1-eta)*A_orig + eta*A_update  
                                    # [B,Hh,L,L]
    Y = OutputProjection(A_final @ v)        
                                    # [B,T,N,d]
\end{lstlisting}
\end{minipage}
\hfill
\vrule width 0.5pt
\hfill
\begin{minipage}[t]{0.48\textwidth}
\textbf{FlashLite mode}

\begin{lstlisting}
X: video tokens [B,T,N,d], L=T*N
    # B: batch size, T: frames
    # N: tokens/frame, d: embed dim
    # Hh: heads, dh: head dim

for X in loader:

    q,k,v = QKV(X)                   # [B,Hh,L,dh]

    E = HeadAveragedKeyEnergy(k)        # [B,T,N]


    c = FrameConfidence(E)              # [B,T]
    p = FrameTokenDistribution(E)       # [B,T,N]
    H = IntraFrameConcentration(p)      # [B,T]

    if pretrain:
        c = EMA(c)

    f = w_con*c + w_ent*H               # [B,T]
    f = Standardize(f)
    pi = Softmax(tau*f)                 # [B,T]
    pi = (1-lambda_pi)*pi + lambda_pi/T

    alpha = clip(T*pi, alpha_min, alpha_max)                                          # [B,T]

    Y = FlashAttention(q,k,v, frame_bias=log(alpha))                              # [B,Hh,L,dh]

    Y = OutputProjection(Y)         # [B,T,N,d]
\end{lstlisting}
\end{minipage}

\end{algorithm*}
%
\paragraph{Exact vs. FlashLite Modes.}
We summarize the practical differences between the exact and flashLite modes,
as implemented in \cref{alg:our_modes}.
Exact mode operates on the explicitly materialized attention map $A \in \mathbb{R}^{B \times N_h \times L \times L}$, from which the head-averaged attention $\bar{A}$ is used to compute the frame-level confidence $c_t$ and intra-frame concentration $H_t$.
The composite prior $\pi_t$ is then constructed, and a ratio-based alignment is applied to adjust the frame-level marginal distribution.
This reweighting is performed after softmax, followed by uniform smoothing, row-wise renormalization, and residual interpolation with the original attention.

FlashLite mode, in contrast, avoids materializing $A$ by replacing attention-derived statistics with key-based proxy statistics.
Specifically, frame-level scores $(c_t, H_t)$ are approximated from the key energy tensor $E \in \mathbb{R}^{B \times T \times N}$, following the same confidence--concentration formulation used in exact mode.
The resulting prior $\pi_t$ is converted into frame-level scaling factors, which are injected as additive log-biases into the attention logits.
This enables framewise modulation to be incorporated into the pre-softmax computation, thereby preserving compatibility with fused FlashAttention kernels.

The two modes primarily differ in where the modulation is applied within the attention pipeline.
Exact mode performs \emph{post-softmax} attention redistribution with explicit row-wise renormalization, whereas FlashLite applies a corresponding
\emph{pre-softmax} modulation via logit bias.
Despite this difference, both modes share a common scoring and prior
construction pipeline, differing primarily in the source of statistics
(attention vs.\ key energy) and the point at which the modulation is applied.

\subsection{Model Configurations}
\label{Appx:ModelConfigurations}
We follow a two-stage training pipeline consisting of large-scale self-supervised pretraining on VoxCeleb2 and supervised finetuning for downstream facial emotion recognition.
Most hyperparameters are kept consistent with the VideoMAE framework~\cite{tong2022videomae} to ensure fair comparison with established baselines.
A detailed summary of training configurations is provided in \cref{tab:pretrain_finetune_setup}.

\paragraph{Backbone Architecture.}
We adopt ViT-B/16, ViT-L/16, and ViT-H/16 as the backbone, following the standard VideoMAE~\cite{tong2022videomae} configuration.
Videos are partitioned into non-overlapping $2\times16\times16$ spatio-temporal tubelets and linearly projected into a sequence of tokens.
The tokens are processed by a ViT encoder with multi-head self-attention, while a lightweight decoder is used only during pretraining for masked reconstruction.
%
\begin{table*}[t]
\caption{Training configurations for self-supervised pretraining and supervised finetuning across ViT-B/16, ViT-L/16, and ViT-H/16 backbones.}
\label{tab:pretrain_finetune_setup}

\centering
\small
\setlength{\tabcolsep}{5pt}
\renewcommand{\arraystretch}{1.12}

\resizebox{\linewidth}{!}{
\begin{tabular}{p{0.20\linewidth}p{0.37\linewidth}p{0.37\linewidth}}
\toprule

\textbf{Configurations}
& \textbf{Pretraining}
& \textbf{Finetuning} \\

\midrule

backbone
& ViT-B/16, ViT-L/16, ViT-H/16
& ViT-B/16, ViT-L/16, ViT-H/16 \\

dataset
& VoxCeleb2
& target datasets \\

input resolution
& $224\times224$
& $224\times224$ \\

frames per clip
& 16
& 16 \\

sampling rate
& 4
& 4 \\

tubelet size
& $2\times16\times16$
& $2\times16\times16$ \\

masking ratio
& 0.9
& -- \\

ViT decoder
& 4-layer / 8-layer / 8-layer
& -- \\

pooling
& --
& mean pooling over tokens \\

global batch size
& 128$\times$32 / 64$\times$40 / 64$\times$40
& 16$\times$1 \\

optimizer
& AdamW
& AdamW \\

base learning rate
& $1.5\mathrm{e}{-4}$ / $7\mathrm{e}{-5}$ / $8\mathrm{e}{-5}$
& $1\mathrm{e}{-3}$ \\

min learning rate
& $1\mathrm{e}{-5}$
& $1\mathrm{e}{-6}$ \\

optimizer momentum
& $\beta_1=0.9,\ \beta_2=0.95$
& $\beta_1=0.9,\ \beta_2=0.999$ \\

weight decay
& 0.05
& 0.05 \\

layer-wise decay
& --
& 0.75 \\

lr schedule
& cosine decay
& cosine decay \\

warmup epochs
& 20
& 5 \\

training epochs
& 200
& 100 (best epoch selected) \\

drop path
& --
& 0.1 \\

label smoothing
& --
& 0.1 \\

augmentation
& MultiScaleCrop, normalization
& RandAug, color jitter 0.4, random erasing 0.25,
mixup 0.8, cutmix 1.0 \\

\bottomrule
\end{tabular}
}

\end{table*}

\paragraph{Pretraining.}
We pretrain all models on VoxCeleb2 using a masked video modeling objective with a tube masking ratio of 0.9.
Each input clip consists of 16 RGB frames at $224\times224$ resolution, sampled every 4 frames.
Training is conducted on NVIDIA H100 80GB GPUs, using 32 GPUs for ViT-B and 40 GPUs for both ViT-L and ViT-H.
We adopt relatively lightweight data augmentation, consisting of MultiScaleCrop with scale factors $\{1.0, 0.875, 0.75, 0.66\}$ and standard normalization, following the original VideoMAE recipe.
This design emphasizes representation learning from masked reconstruction rather than aggressive input perturbation.

\paragraph{Finetuning.}
For downstream tasks, the encoder is initialized from the 200-epoch self-supervised checkpoint.
The pretraining decoder is removed, and a mean-pooling classification head is applied over the token embeddings.
Finetuning is basically performed on a single GPU with a batch size of 16 for 100 epochs using AdamW with layer-wise learning rate decay, while minor adjustments to the batch size or number of GPUs are made for larger datasets to accommodate practical runtime constraints.
In contrast to pretraining, we employ stronger data augmentation, including RandAugment, color jitter, random erasing, and mixup/cutmix regularization, to improve generalization.
For evaluation, we follow the standard multi-view testing protocol used in video classification.
Each video is uniformly divided into multiple temporal segments, and multiple spatial crops are extracted per segment.
The model processes all segment–crop pairs independently, and the prediction logits are averaged to produce the final video-level output.

\subsection{Computational Efficiency Analysis}
\label{Appx:ComputationalEfficiencyAnalysis}
\begin{table*}[t]
\centering
\scriptsize
\setlength{\tabcolsep}{4pt}

\caption{
Runtime and memory comparison of our ViT-B/16 and ViT-L/16 training variants on a single H100 GPU.
FLOPs are reported per 16-frame 224$\times$224 clip under the full-token convention.
Encoder+decoder FLOPs apply only to the self-supervised pretraining stage (using a 4-layer / 8-layer transformer decoder for ViT-B / ViT-L, respectively), while finetuning uses the encoder only.
Peak memory denotes the maximum allocated GPU memory, and iteration time is measured in milliseconds per training step.
Speedup is reported relative to the Exact ViT-B/16 baseline within each stage.
}

\label{tab:flops_pretrain_finetune_vitb_vitl}

\resizebox{\linewidth}{!}{
\begin{tabular}{lccccccc}
\toprule

\multicolumn{8}{c}{\textbf{Self-supervised pretraining (per 16-frame 224$\times$224 clip, 1$\times$H100)}} \\
\midrule

Method
& Backbone
& \makecell{Enc \\ FLOPs(G)}
& \makecell{Enc+Dec \\ FLOPs(G)}
& \makecell{Peak \\ Mem(GB)}
& \makecell{Iter \\ Time(ms)}
& Speedup ($\times$)
& Batch \\

\midrule

Exact
& ViT-B/16
& 357
& 394
& 24.6
& 400.6
& 1.00
& 96 \\

FlashLite
& ViT-B/16
& 357
& 394
& 21.6
& 318.7
& 1.26
& 96 \\

FlashLite
& ViT-L/16
& 1189
& 1226
& 58.5
& 558.9
& 0.72
& 96 \\

\midrule

\multicolumn{8}{c}{\textbf{Supervised finetuning (per 16-frame 224$\times$224 clip, 1$\times$H100)}} \\
\midrule

Method
& Backbone
& \makecell{Enc \\ FLOPs(G)}
& \makecell{Enc+Dec \\ FLOPs(G)}
& \makecell{Peak \\ Mem(GB)}
& \makecell{Iter \\ Time(ms)}
& Speedup ($\times$)
& Batch \\

\midrule

Exact
& ViT-B/16
& 357
& --
& 71.4
& 700.8
& 1.00
& 16 \\

FlashLite
& ViT-B/16
& 357
& --
& 10.2
& 169.9
& 4.12
& 16 \\

FlashLite
& ViT-L/16
& 1189
& --
& 27.1
& 414.2
& 1.69
& 16 \\

\bottomrule
\end{tabular}
}

\end{table*}
\cref{tab:flops_pretrain_finetune_vitb_vitl} summarizes the computational characteristics of our variants on a single H100 GPU.
We report FLOPs, peak memory, throughput, and the batch sizes used in pretraining and finetuning.
FlashLite approximates the functional effect of exact-mode reweighting and integrates it into the attention computation, enabling compatibility FlashAttention kernels~\cite{dao2022flashattention, dao2023flashattention2}.
As a result, flashLite inherits the memory and throughput efficiency of FlashAttention, with larger gains under full-token finetuning.

Both memory and runtime are primarily governed by attention computation, whose cost scales with the number of active tokens and associated memory accesses.
During pretraining, masked autoencoding removes 90\% of patches, substantially reducing the number of visible tokens and thus limiting both memory usage and the throughput gap between exact and flashLite.
In contrast, during finetuning, all tokens are processed, making attention the dominant factor in both memory and runtime.
In this regime, exact mode incurs additional overhead due to explicit reweighting passes over attention maps, whereas flashLite avoids these by integrating reweighting into the attention computation.
This leads to substantial improvements in both memory efficiency and training throughput, achieving up to an order-of-magnitude reduction in memory and \textbf{4.1$\times$} speedup for ViT-B/16, while remaining consistently more efficient for ViT-L/16.

\subsection{Detailed Comparisons on Facial Expression Recognition}
\label{Appx:DetailedComparisonsonFacialEmotionRecognition}
For clarity of evaluation, we first provide a brief description of the metrics (WAR and UAR), which are the standard evaluation metrics adopted by prior state-of-the-art methods in facial emotion recognition.
Then, we present confusion matrix analyses to further examine class-wise performance and error patterns.

\paragraph{Main Evaluation Metrics.}
We employ weighted average recall (WAR) and unweighted average recall (UAR), following standard evaluation protocols in prior state-of-the-art facial emotion recognition studies.
Let $C$ denote the number of classes, $N_i$ the number of samples in class $i$, and $R_i$ the recall of class $i$, defined as
\[
R_i = \frac{TP_i}{TP_i + FN_i},
\]
where $TP_i$ and $FN_i$ denote the number of true positives and false negatives for class $i$, respectively.
WAR is defined as the class-frequency-weighted average of recalls:
\[
\mathrm{WAR} = \frac{\sum_{i=1}^{C} N_i R_i}{\sum_{i=1}^{C} N_i}.
\]
In standard single-label classification, each sample belongs to exactly one class, implying $TP_i + FN_i = N_i$ and thus $R_i = TP_i/N_i$.
Substituting this into WAR gives
\[
\mathrm{WAR} = \frac{\sum_{i=1}^{C} N_i \cdot \frac{TP_i}{N_i}}{\sum_{i=1}^{C} N_i}
= \frac{\sum_{i=1}^{C} TP_i}{\sum_{i=1}^{C} N_i}.
\]
Since $\sum_i TP_i$ is the total number of correctly classified samples and $\sum_i N_i$ is the total number of samples, WAR is equivalent to the overall accuracy:
\[
\mathrm{ACC} = \frac{\sum_i TP_i}{\sum_i N_i} = \mathrm{WAR}.
\]
In contrast, UAR computes the average recall across classes:
\[
\mathrm{UAR} = \frac{1}{C}\sum_{i=1}^{C} R_i.
\]
When the class distribution is balanced (i.e., $N_i$ is identical for all $i$), the weighting coefficients in WAR become uniform, making WAR equivalent to UAR.
However, in imbalanced settings, WAR reflects performance under the empirical data distribution and therefore places greater emphasis on frequent classes.
In contrast, UAR corresponds to the average recall under equal weighting across classes, assigning equal importance to all classes regardless of frequency.
From this perspective, WAR measures overall classification accuracy, whereas UAR provides a class-balanced evaluation metric. 
Since facial emotion recognition datasets are typically imbalanced, UAR is often preferred for evaluating balanced recognition performance across emotion categories.

\paragraph{Detailed Results with Confusion Matrices.}
To provide a finer-grained analysis of the main results, we present confusion matrices for each dataset.
These analyses complement the extended comparisons above by revealing class-wise performance and error patterns.
As shown in Figs.~\ref{fig:dfew_confusion_all}, \ref{fig:mafw_confusion_all}, and~\ref{fig:ferv39k_confusion_all}, the confusion matrices on DFEW, MAFW, and FERV39k provide insights into class-wise recognition performance and the dominant misclassification patterns across emotions.

\subsection{Examples of Micro-Expression Datasets}
\label{ExamplesofMicro-ExpressionDatasets}
To provide qualitative insight into facial dynamics representations, we leverage micro-expression datasets, namely SAMM~\cite{davison2016samm} and MMEW~\cite{ben2021video}, and illustrate with representative examples.
As shown in Fig.~\ref{fig:micro_examples}, facial regions are tightly aligned across frames, and the observable changes are extremely subtle, highlighting the importance of modeling fine-grained temporal dynamics.
Accordingly, we use a 3-class setting (negative, positive, surprise) in SAMM to mitigate label ambiguity and focus on these subtle temporal cues.
For MMEW, we extend this formulation to a 4-class scheme (negative, positive, surprise, neutral) while preserving consistency across datasets.
\begin{figure}[h]
    \centering
    \begin{minipage}{0.49\linewidth}
        \centering
        \includegraphics[width=\linewidth]{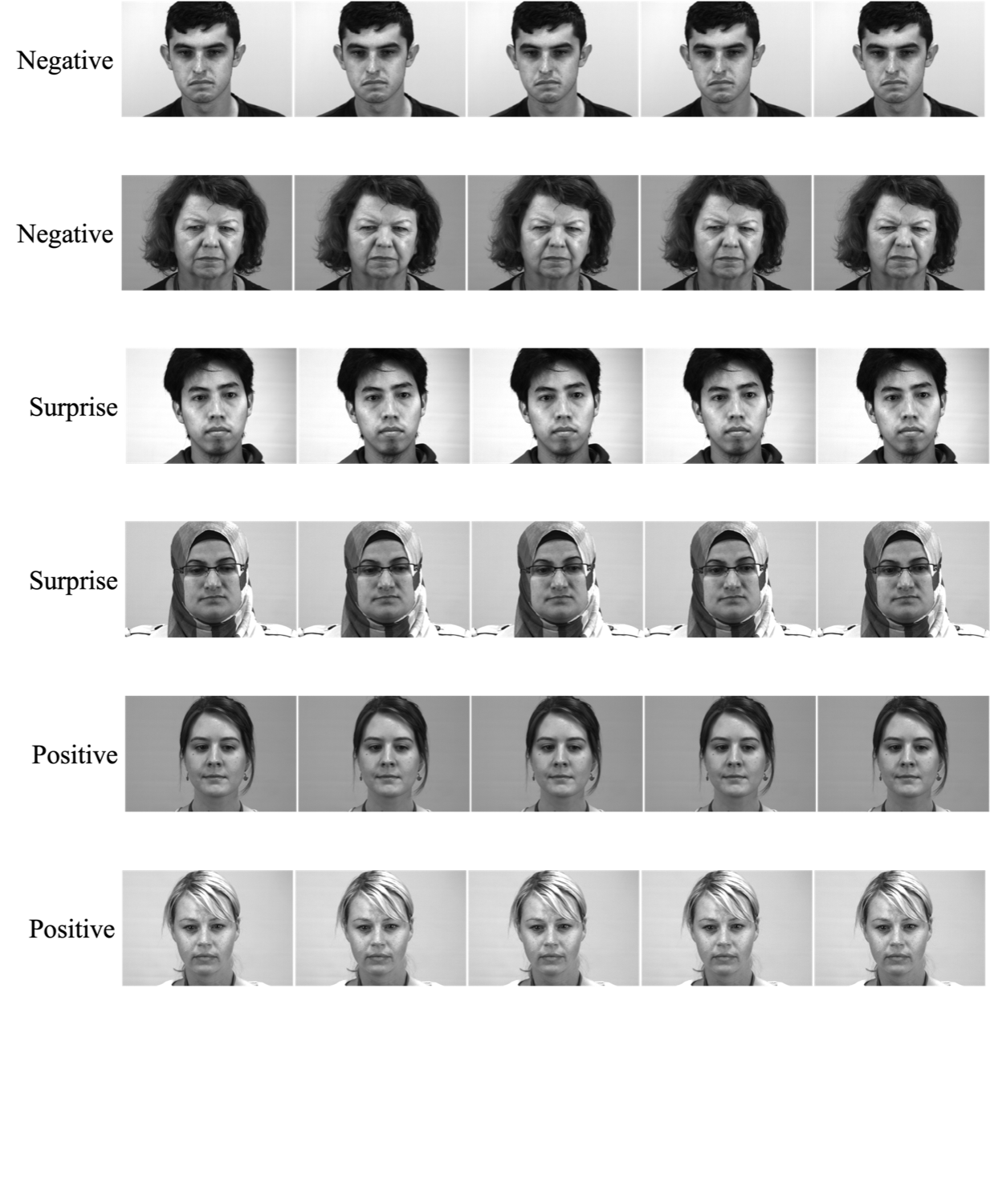}
    \end{minipage}
    \hfill
    \begin{minipage}{0.49\linewidth}
        \centering
        \includegraphics[width=\linewidth]{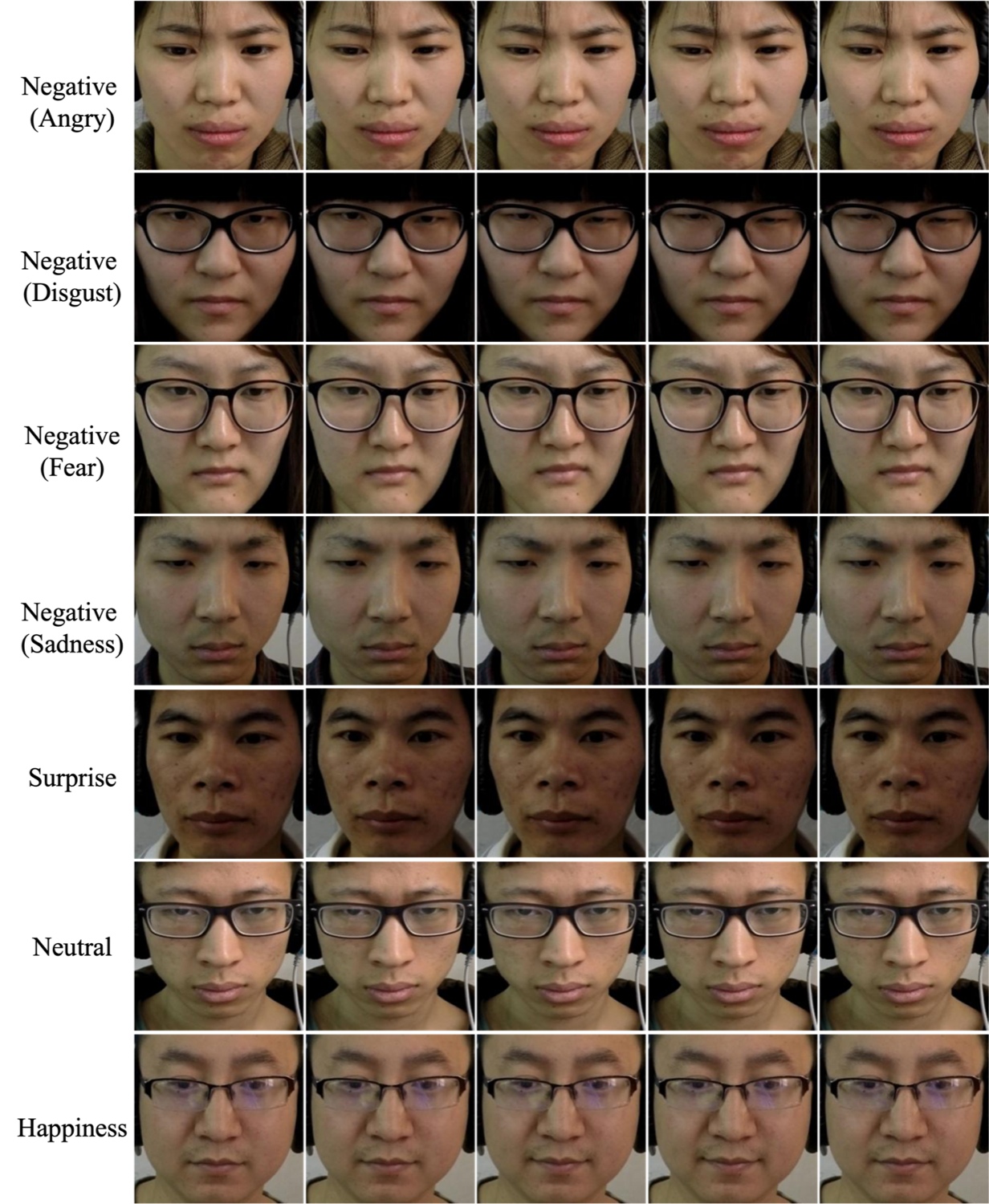}
    \end{minipage}
    \caption{\textbf{Representative examples from SAMM (left) and MMEW (right).} Facial regions are tightly aligned across frames, with only subtle temporal variations observable.}
    \label{fig:micro_examples}
\vspace{-25pt}
\end{figure}

\subsection{Additional Attention Visualizations}
\label{AttentionVisualizations}
\cref{fig:local_global_attention_appx} presents additional qualitative examples of global and region-conditioned attention visualizations.
Across different samples and model scales, MiRA consistently exhibits selective attention allocation and stronger cross-frame interactions, supporting the observations discussed in the main text.
\begin{figure*}[h]
\centering
\includegraphics[width=\textwidth]{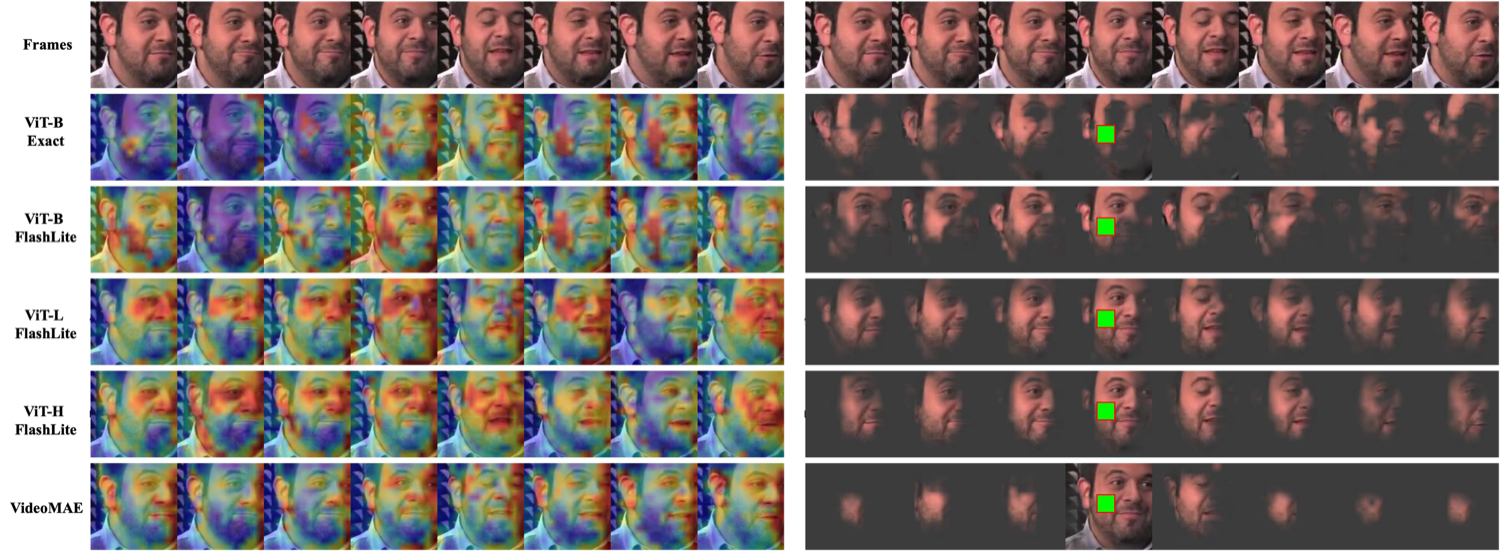}
\end{figure*}
\begin{figure*}[h]
\centering
\includegraphics[width=\textwidth]{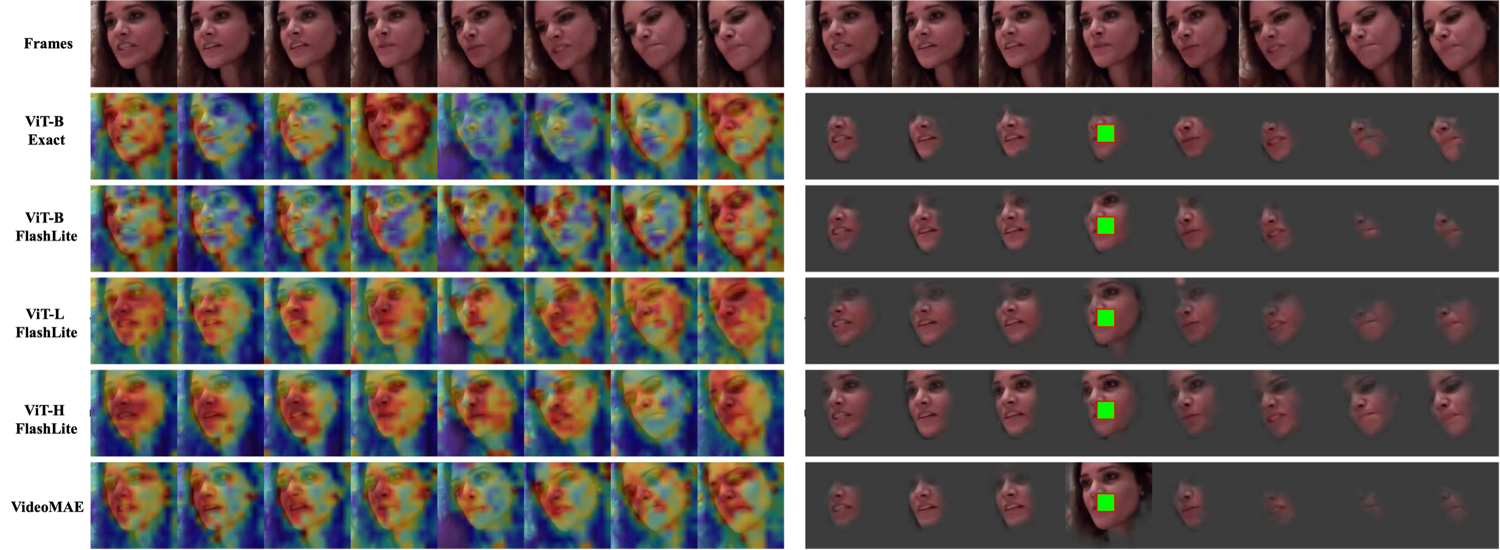}\\[5mm]
\includegraphics[width=\textwidth]{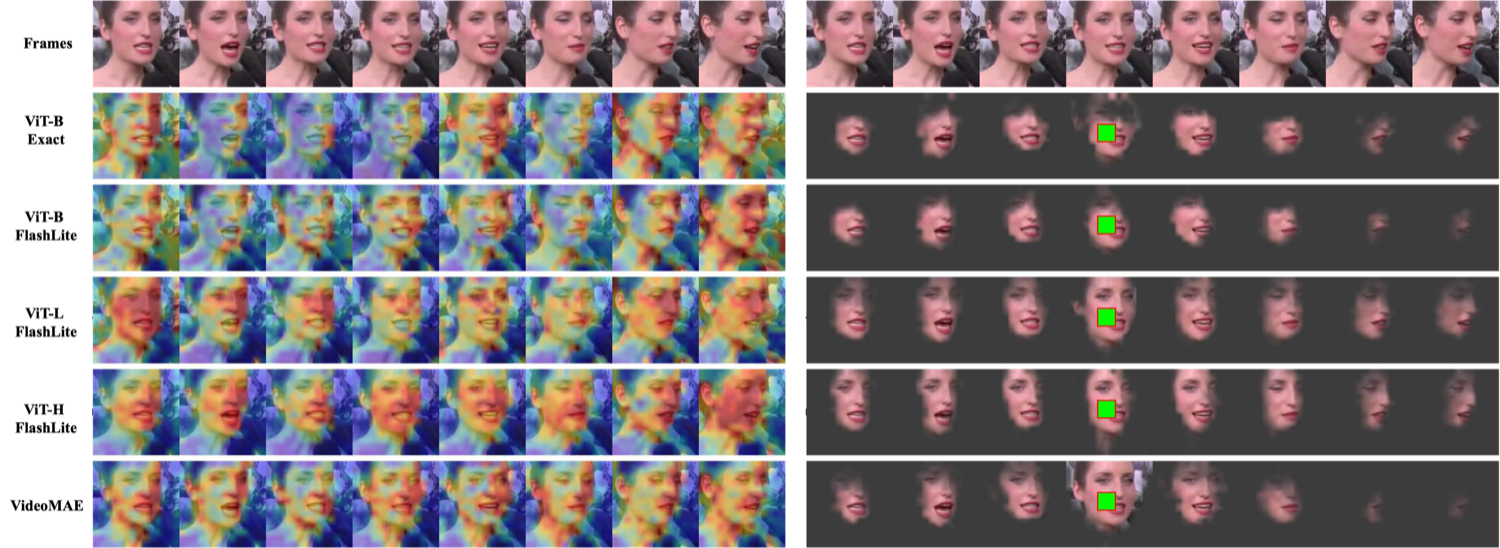}\\[5mm]
\includegraphics[width=\textwidth]{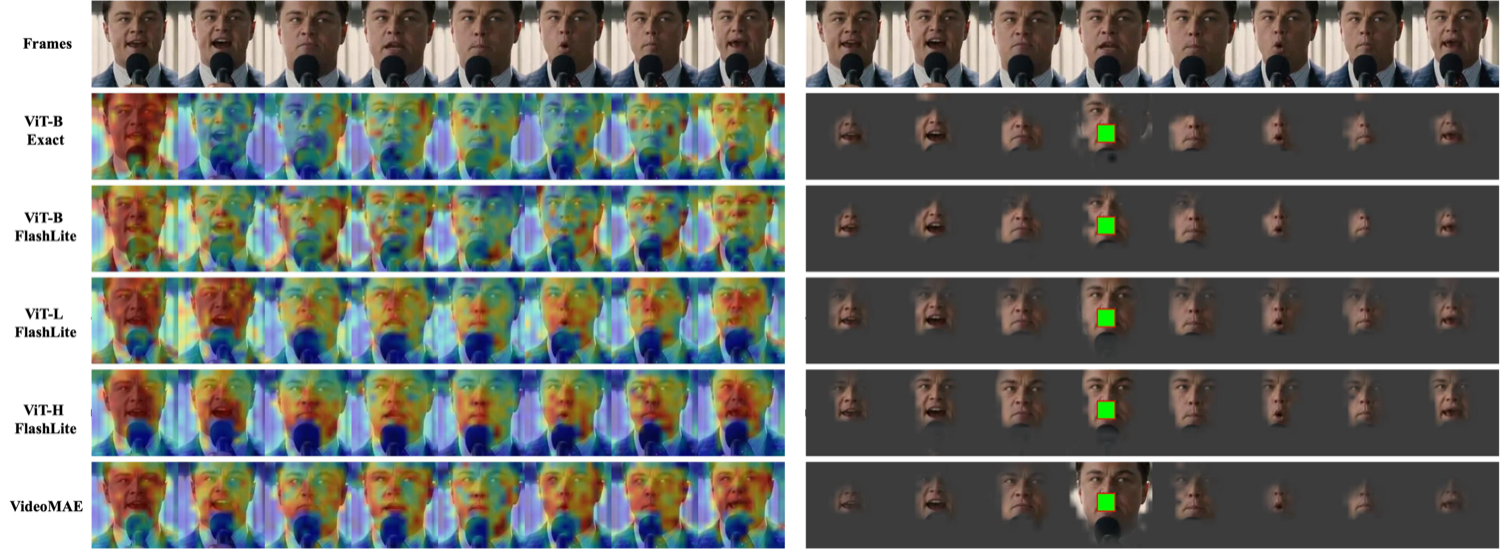}
\caption{
\textbf{Local and global attention visualizations.}
Left: global attention-preserved frames, where the visibility of image regions is modulated according to the attention accumulated over the full spatio-temporal sequence.
Right: region-conditioned attention heatmaps associated with the selected facial region (green box), illustrating the strength of interactions between the highlighted region and spatial-temporal locations across the entire video sequence.
}
\label{fig:local_global_attention_appx}
\end{figure*}

\begin{figure}[t]
\centering
\setlength{\tabcolsep}{2pt}
\begin{tabular}{>{\centering\arraybackslash}m{1.4cm}ccc}

&
\textbf{ViT-B}
&
\textbf{ViT-L}
&
\textbf{ViT-H}
\\[2mm]

\textbf{Fold 1}
&
\includegraphics[width=0.29\textwidth]{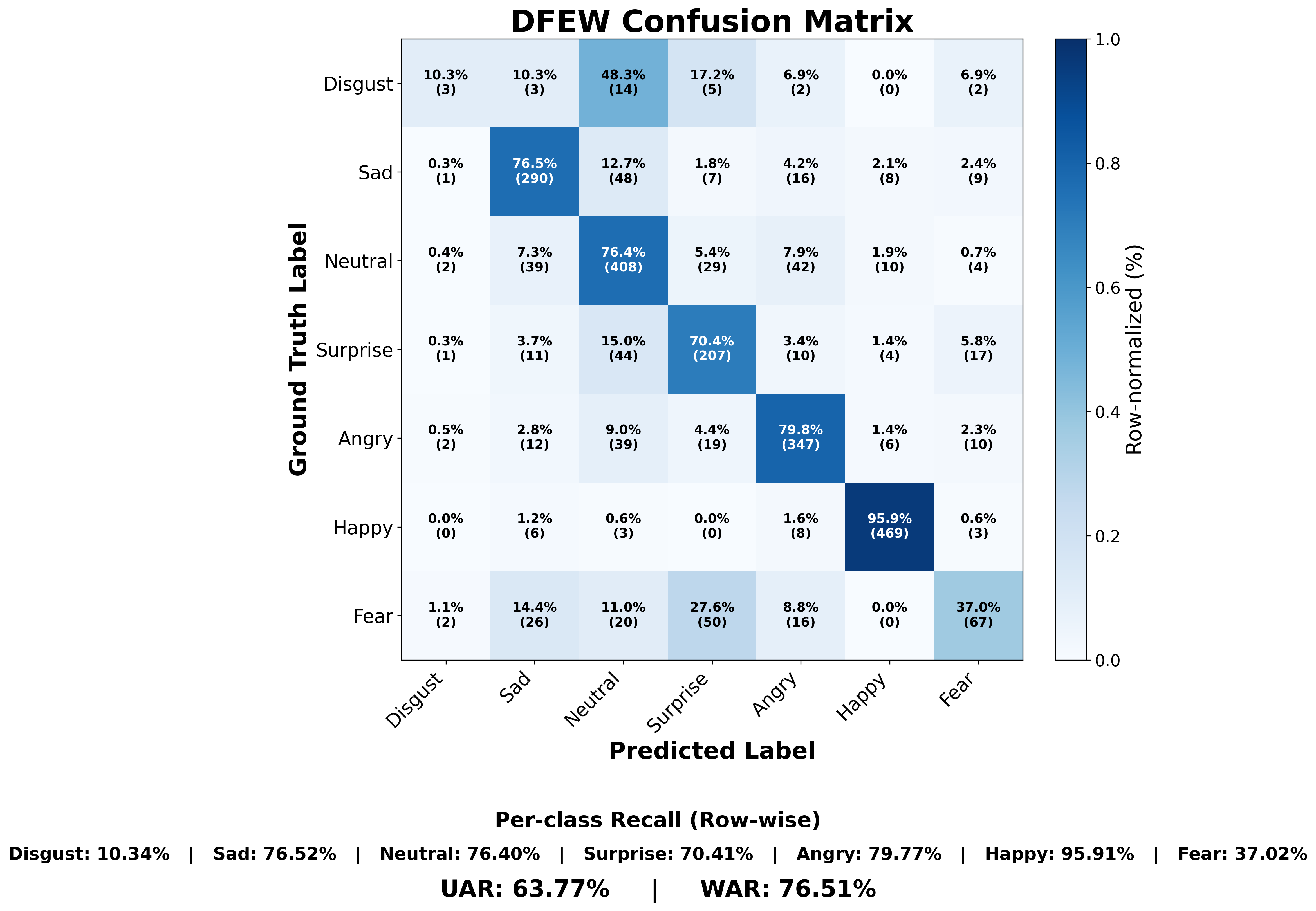}
&
\includegraphics[width=0.29\textwidth]{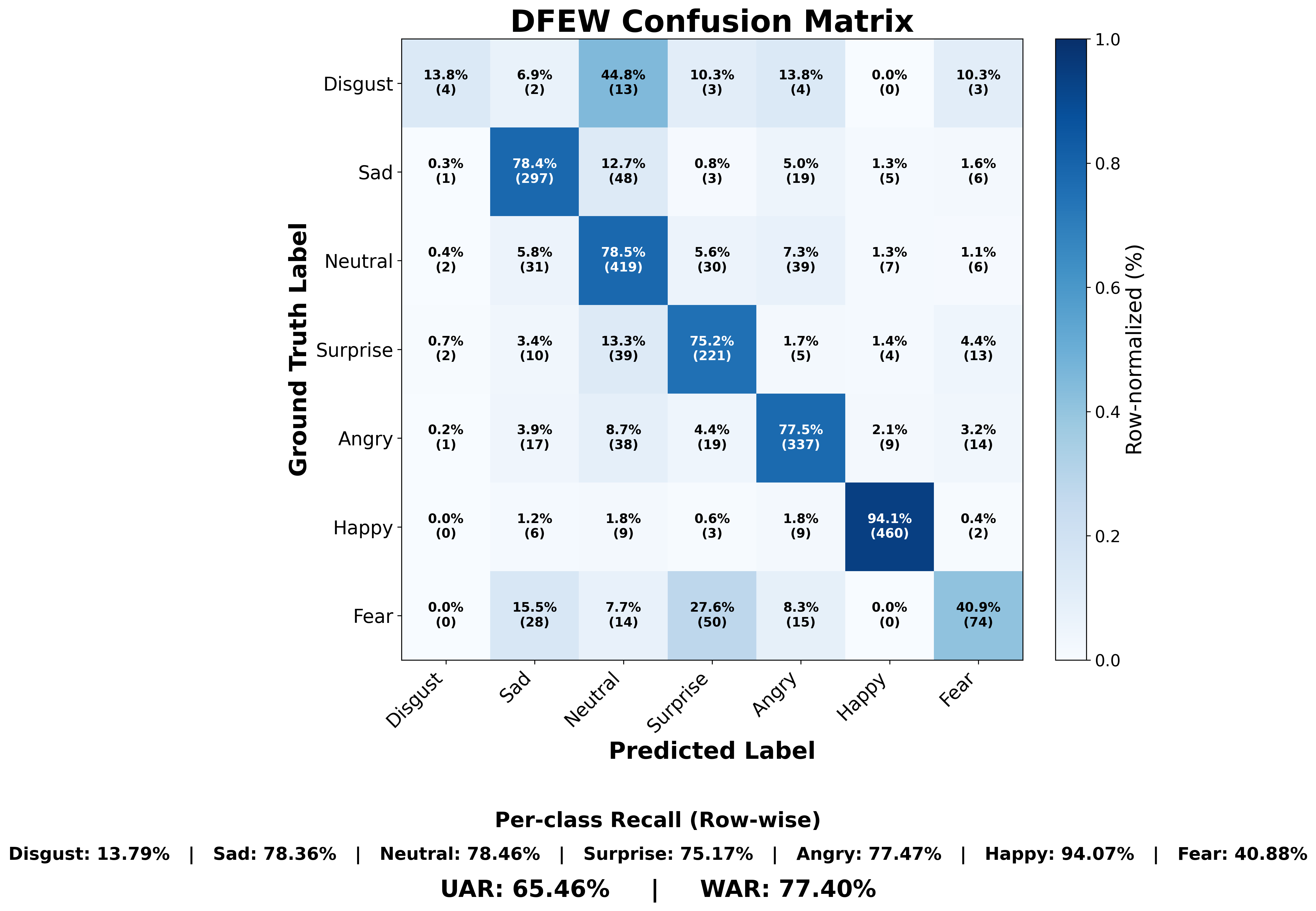}
&
\includegraphics[width=0.29\textwidth]{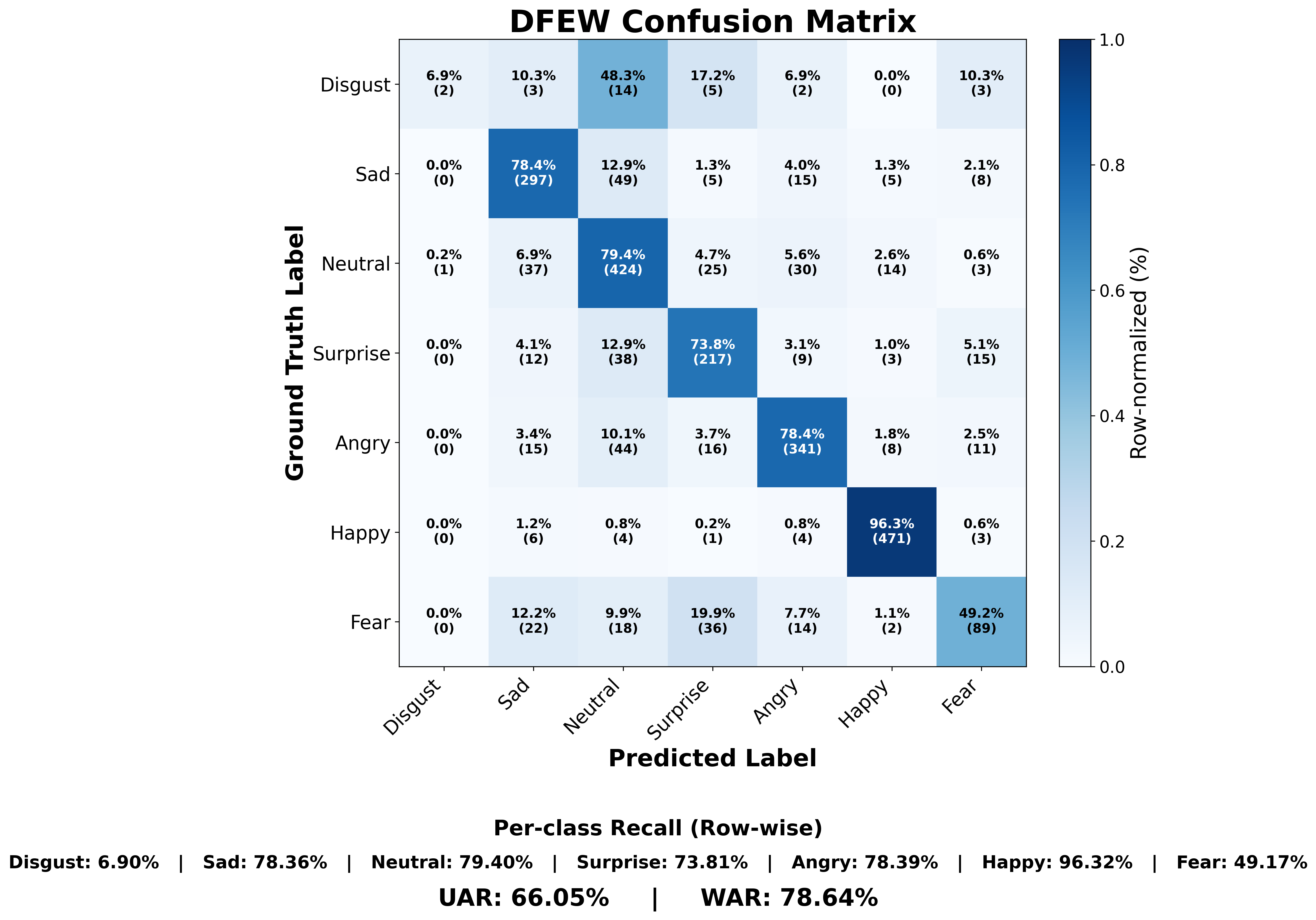}
\\[2mm]

\textbf{Fold 2}
&
\includegraphics[width=0.29\textwidth]{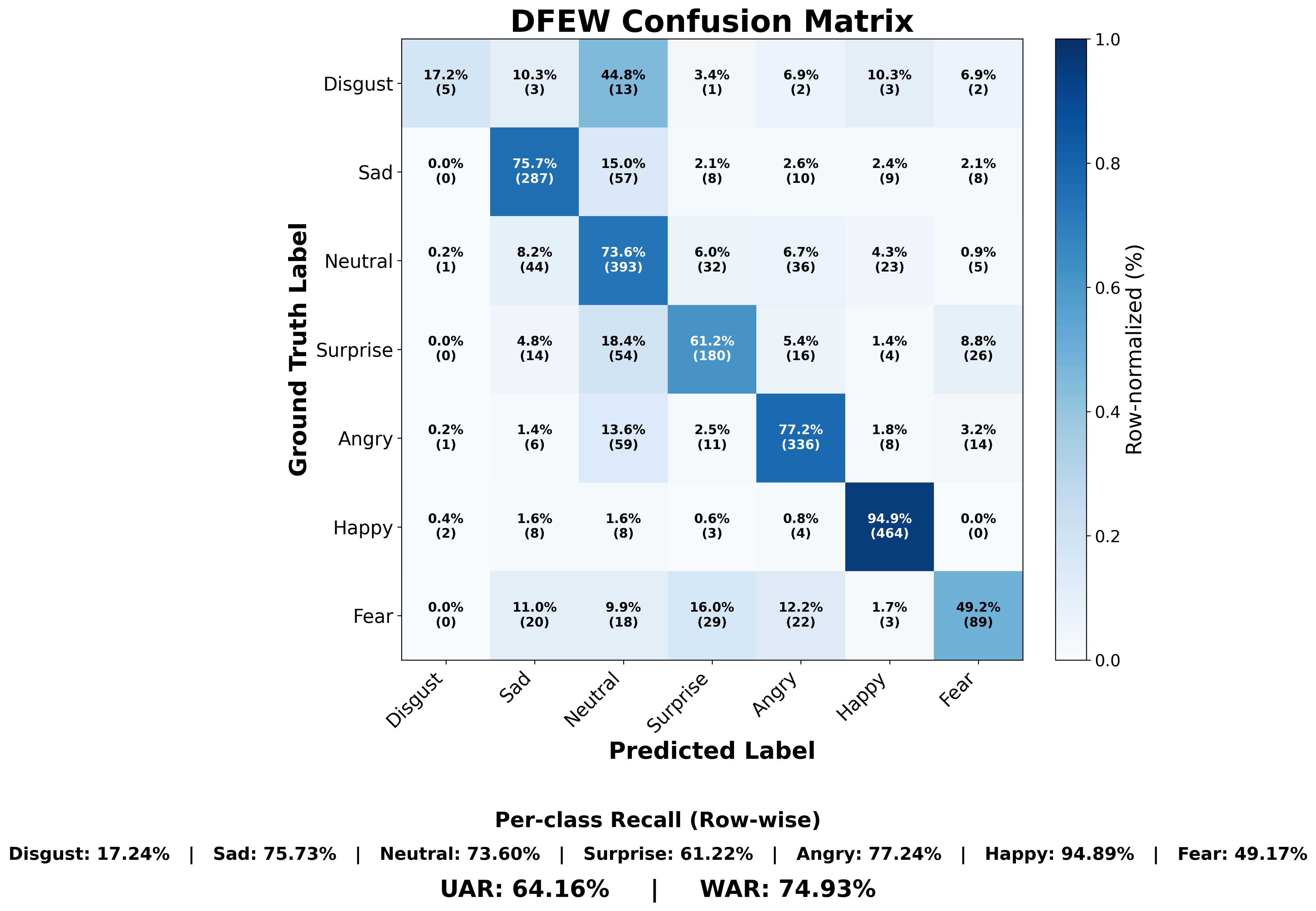}
&
\includegraphics[width=0.29\textwidth]{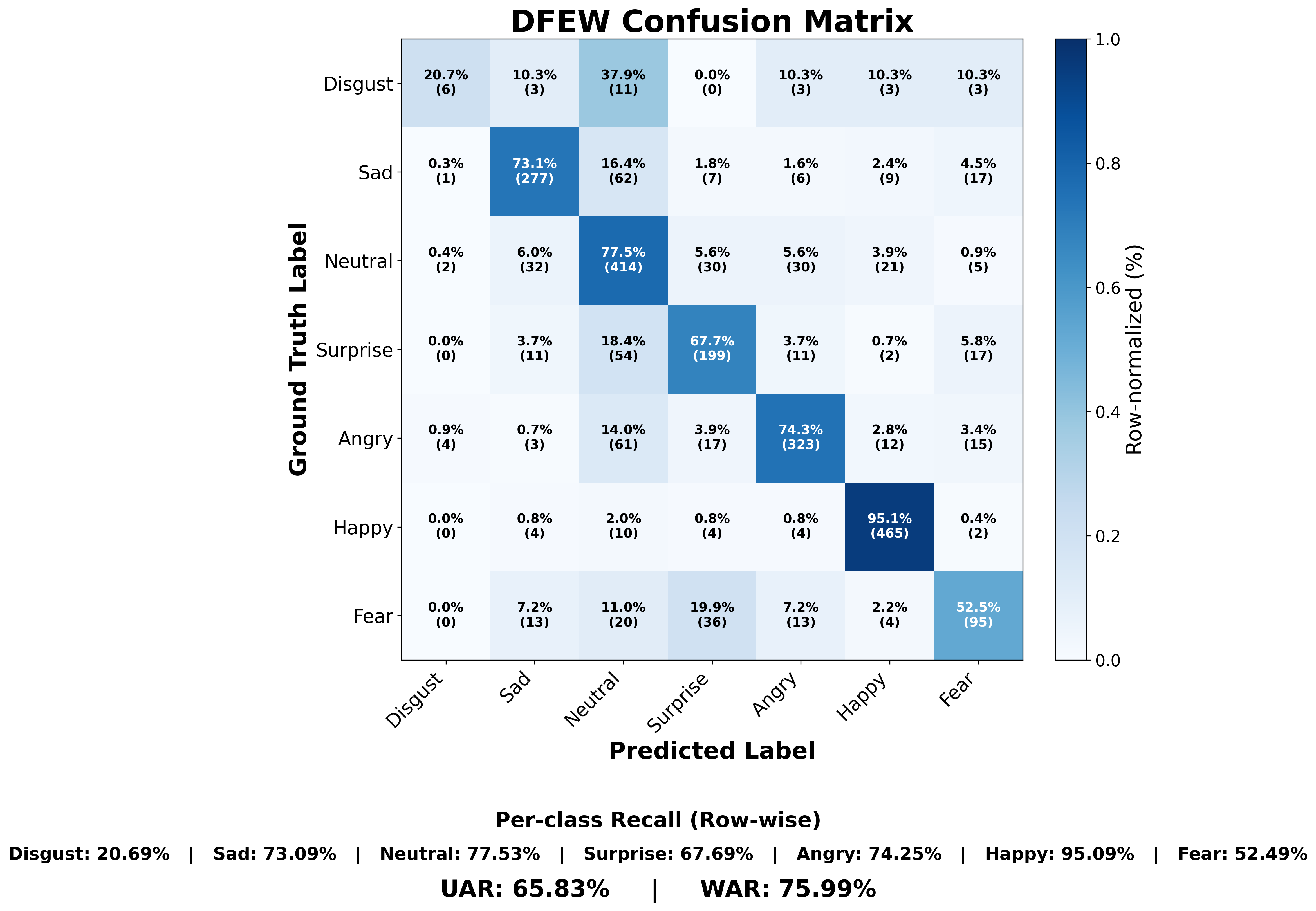}
&
\includegraphics[width=0.29\textwidth]{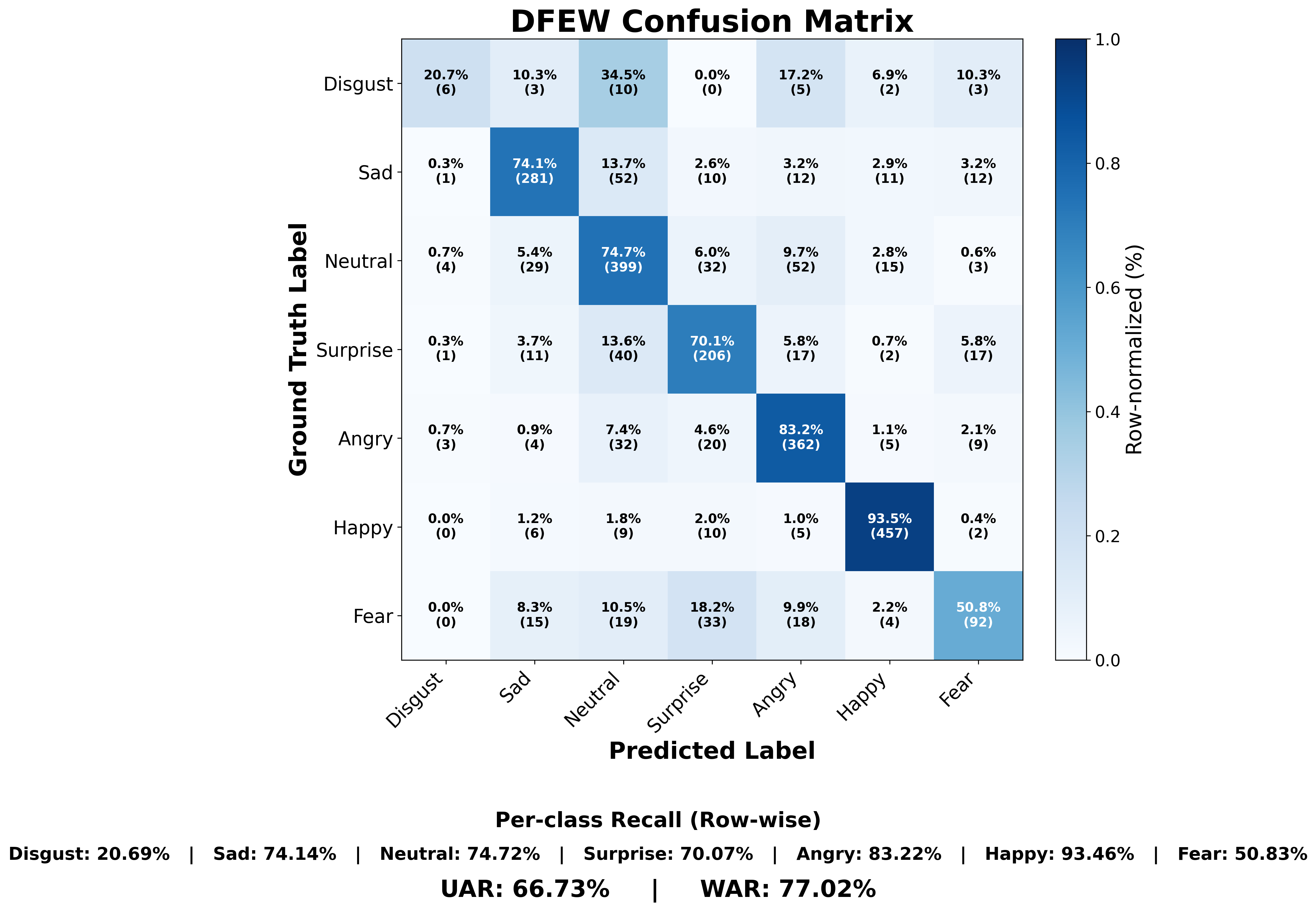}
\\[2mm]

\textbf{Fold 3}
&
\includegraphics[width=0.29\textwidth]{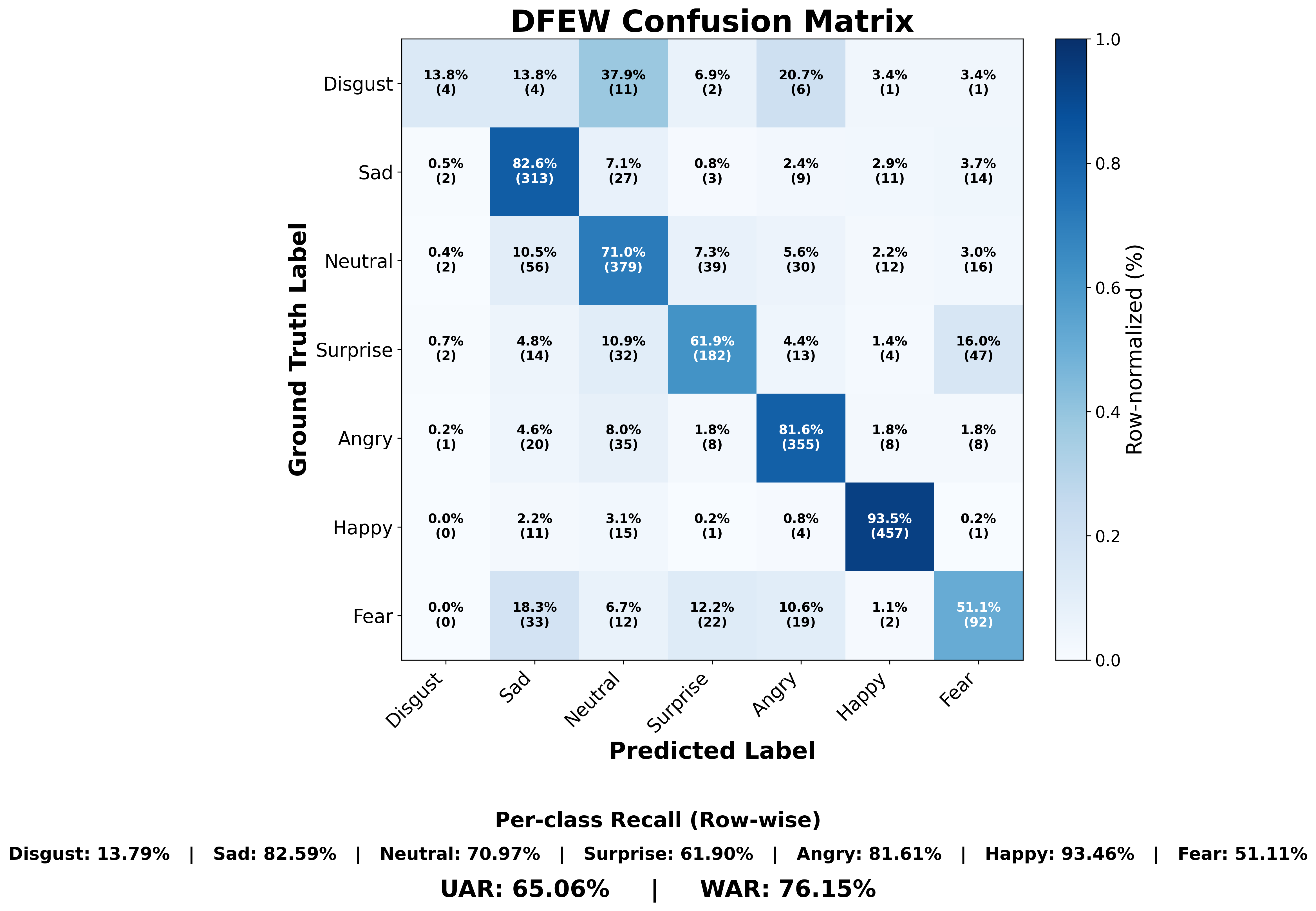}
&
\includegraphics[width=0.29\textwidth]{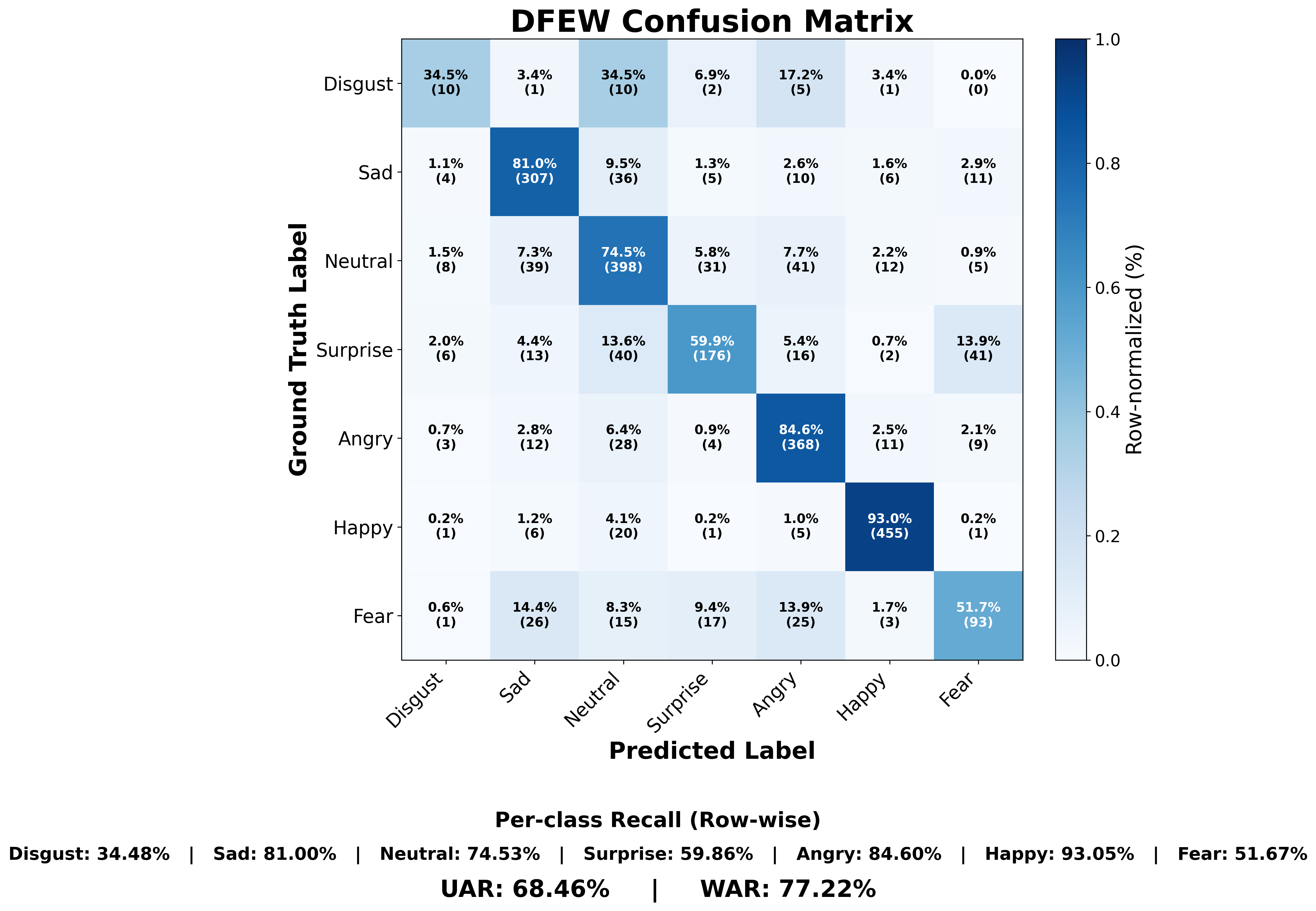}
&
\includegraphics[width=0.29\textwidth]{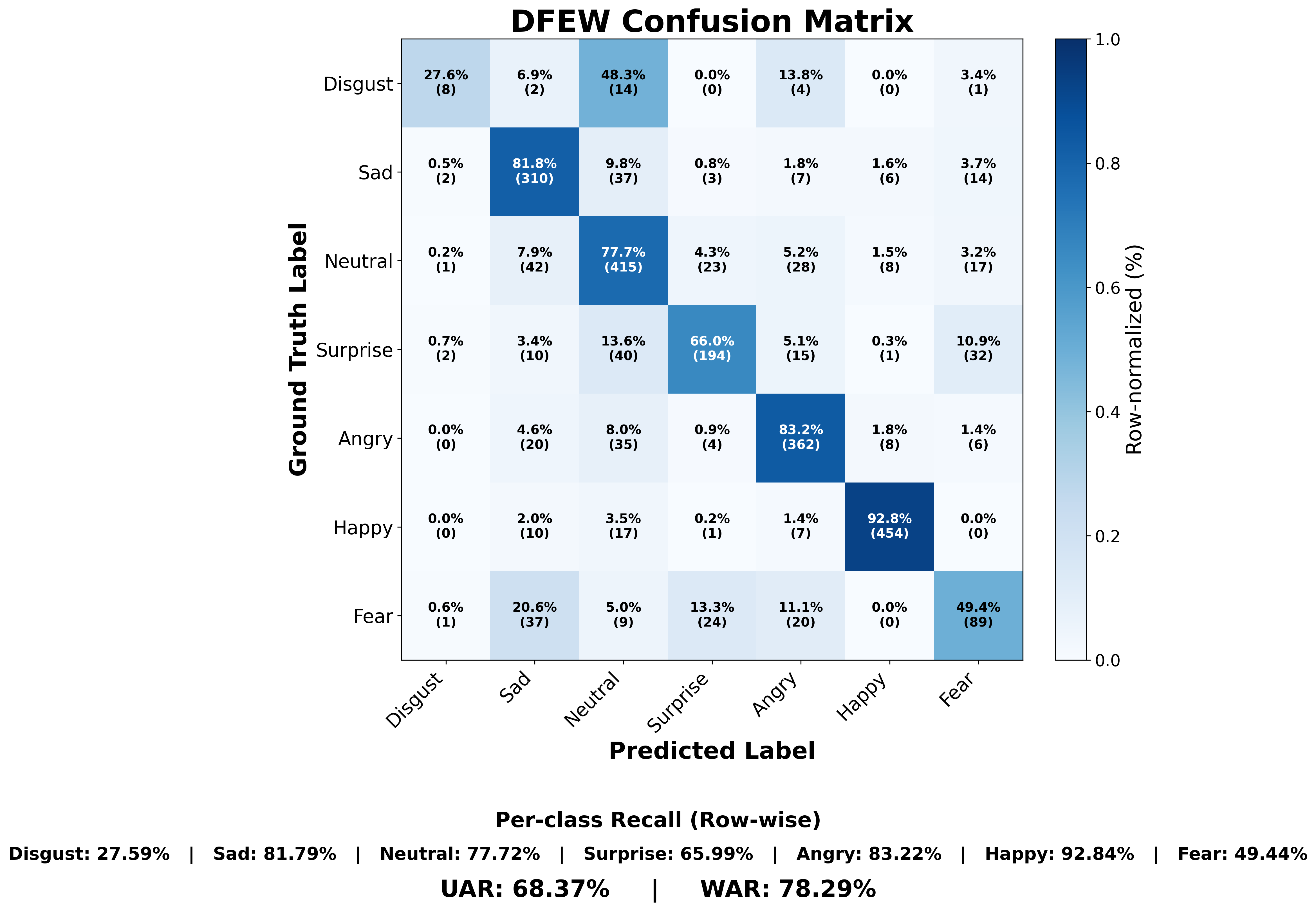}
\\[2mm]

\textbf{Fold 4}
&
\includegraphics[width=0.29\textwidth]{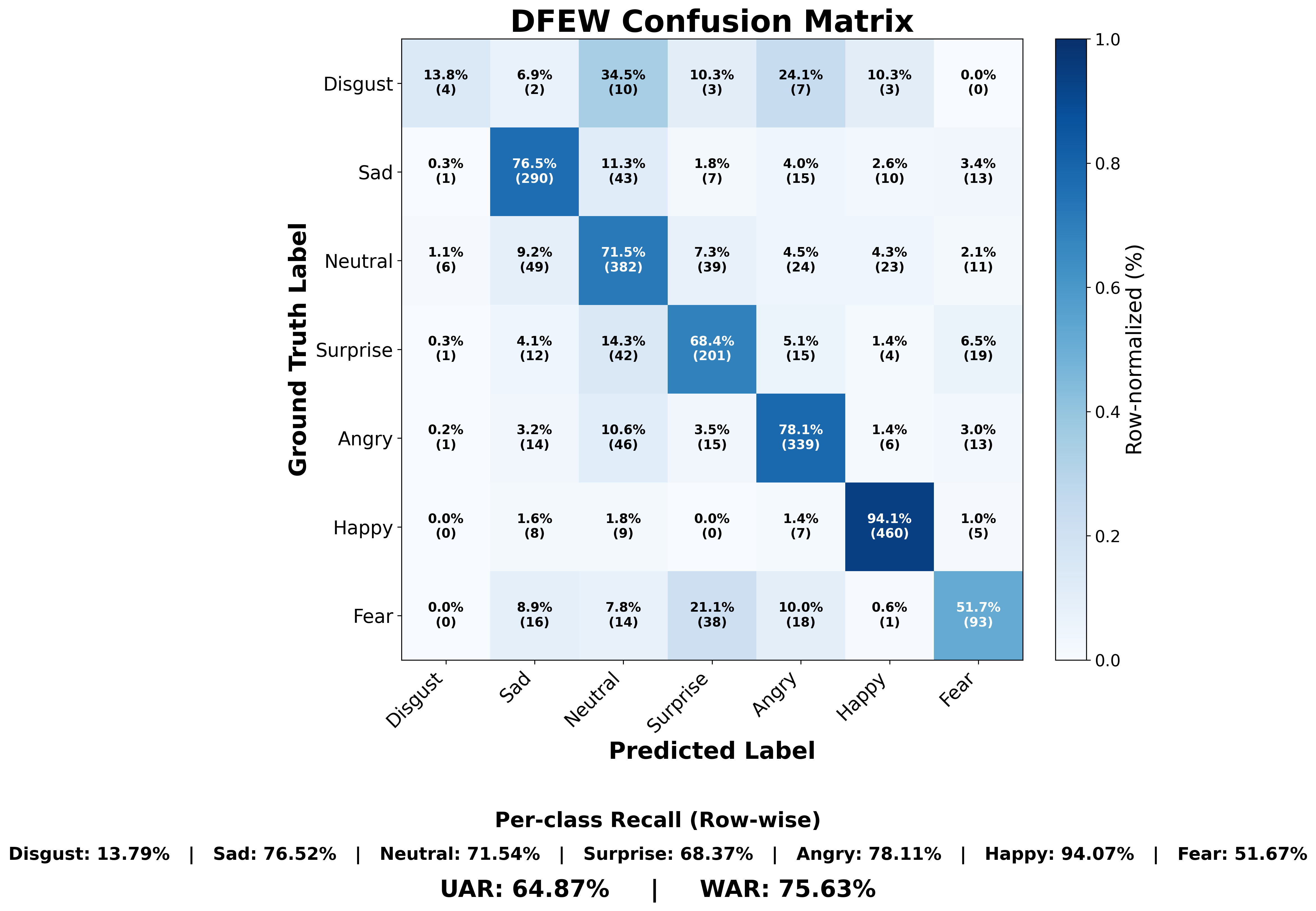}
&
\includegraphics[width=0.29\textwidth]{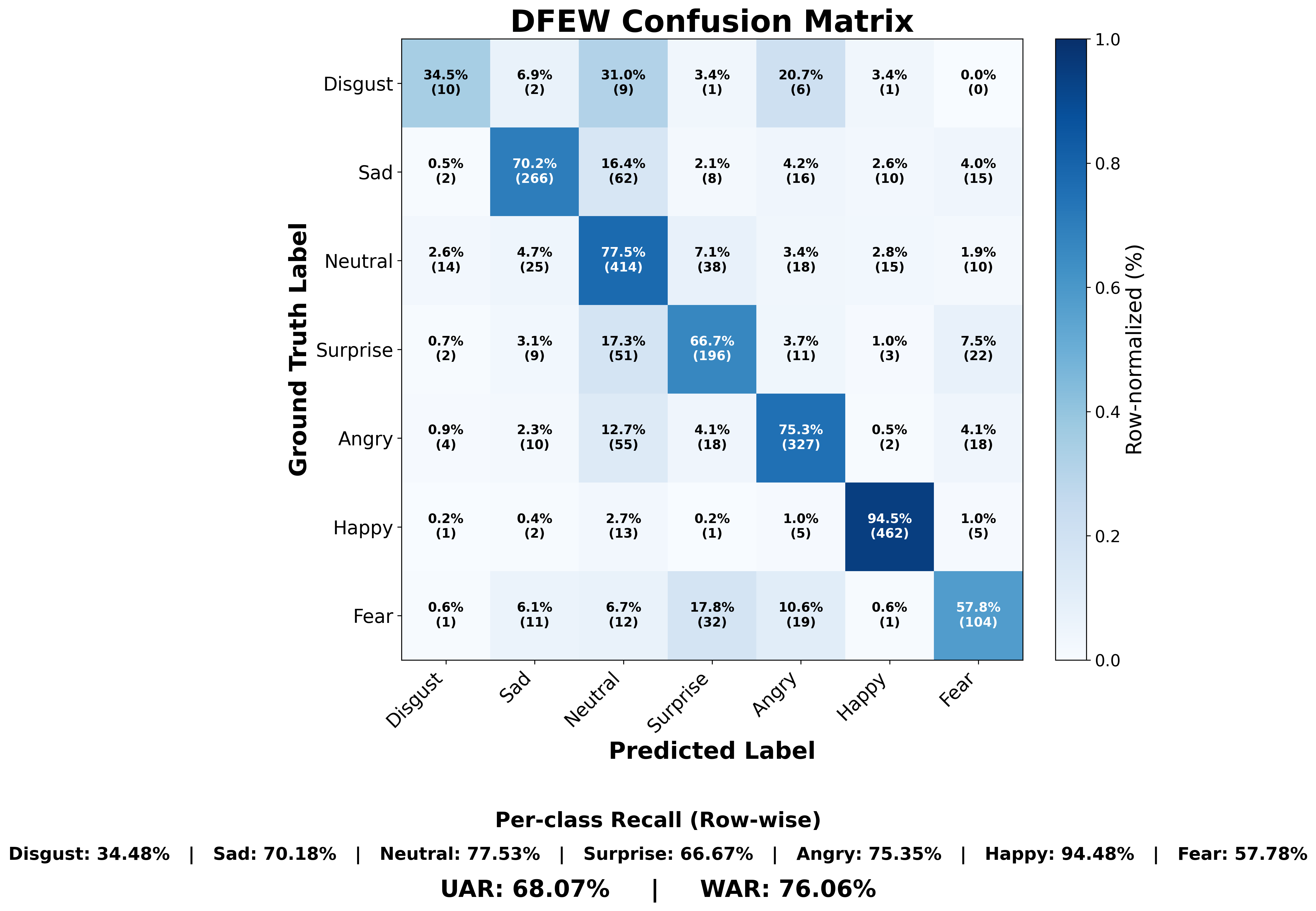}
&
\includegraphics[width=0.29\textwidth]{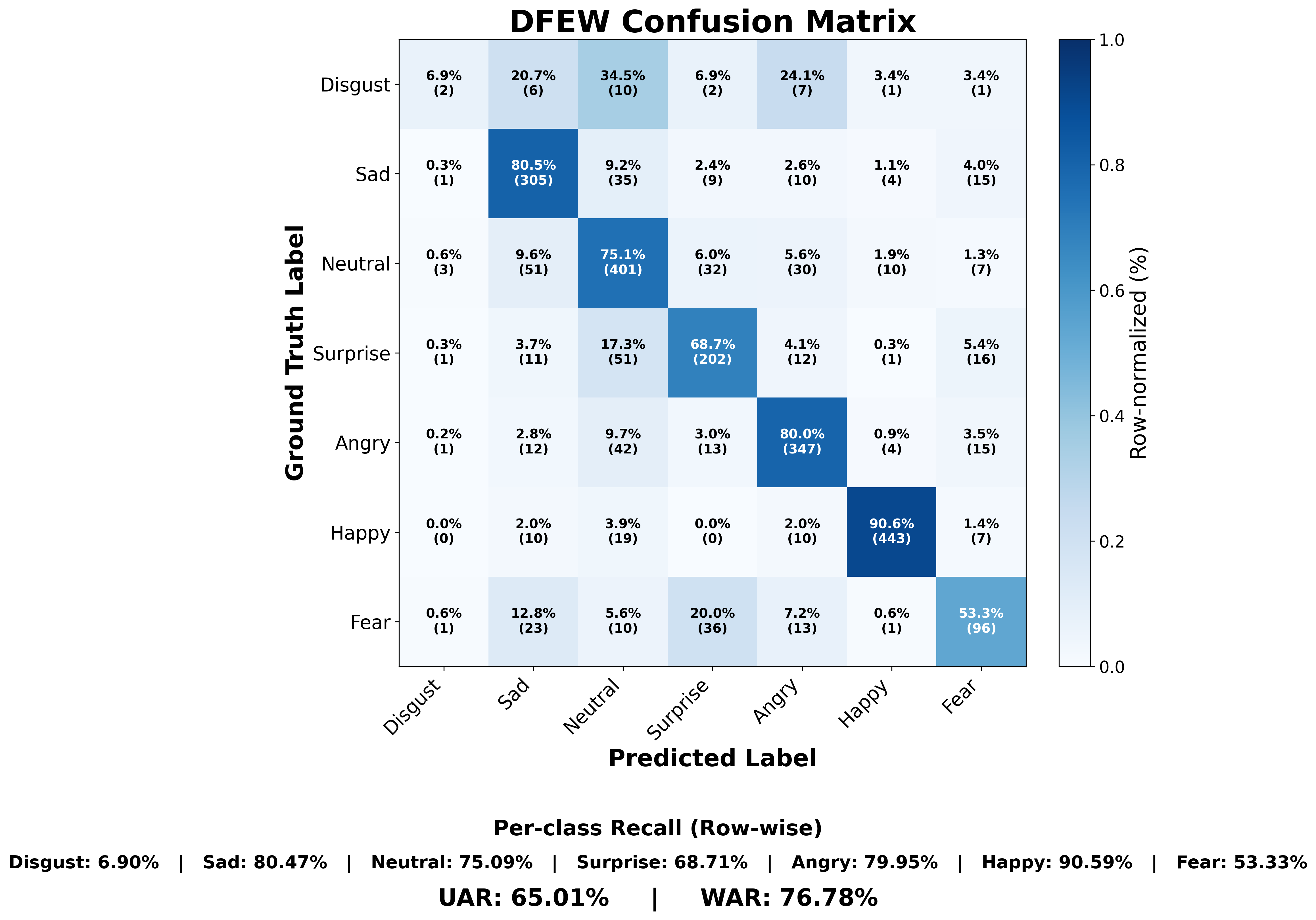}
\\[2mm]

\textbf{Fold 5}
&
\includegraphics[width=0.29\textwidth]{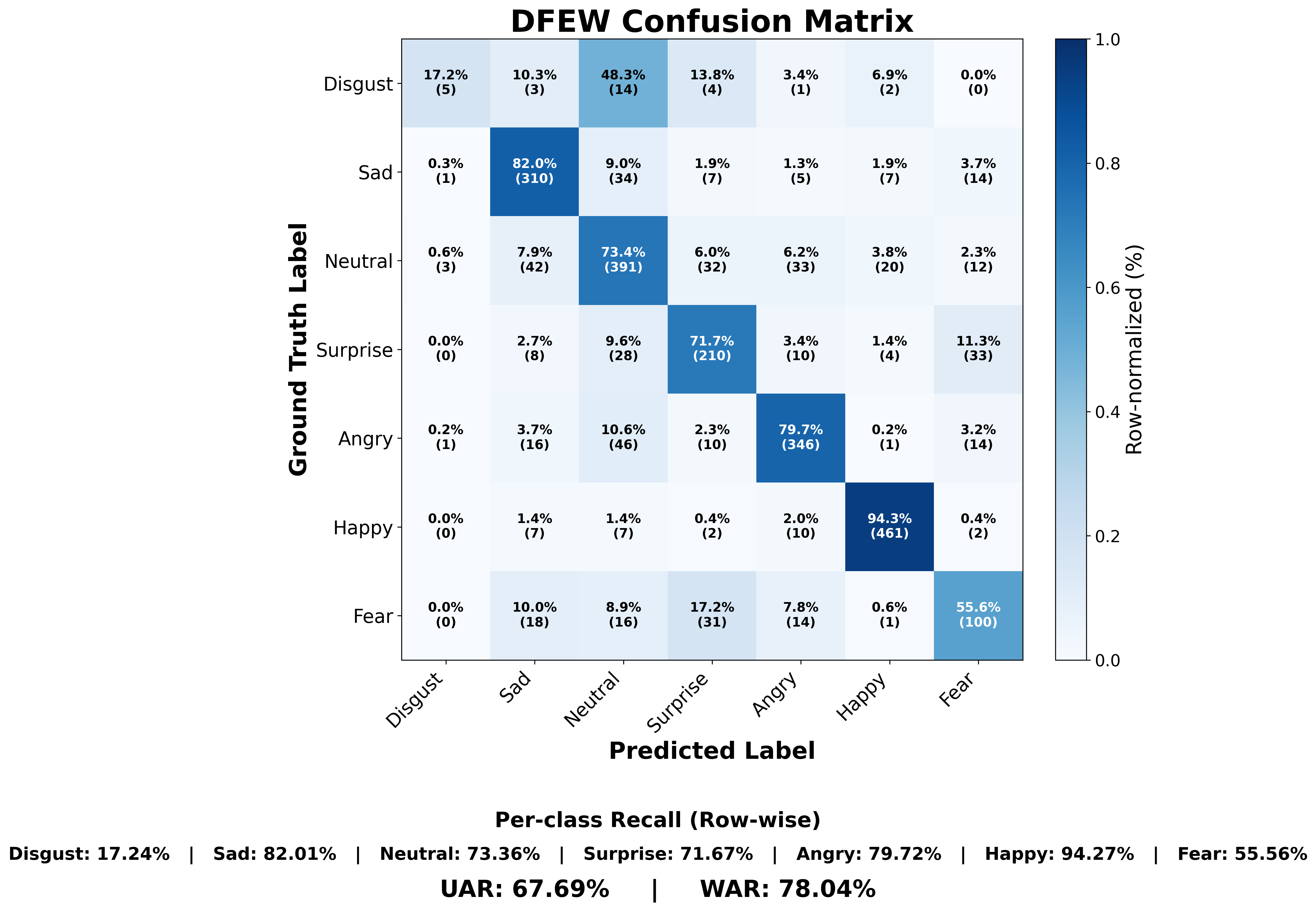}
&
\includegraphics[width=0.29\textwidth]{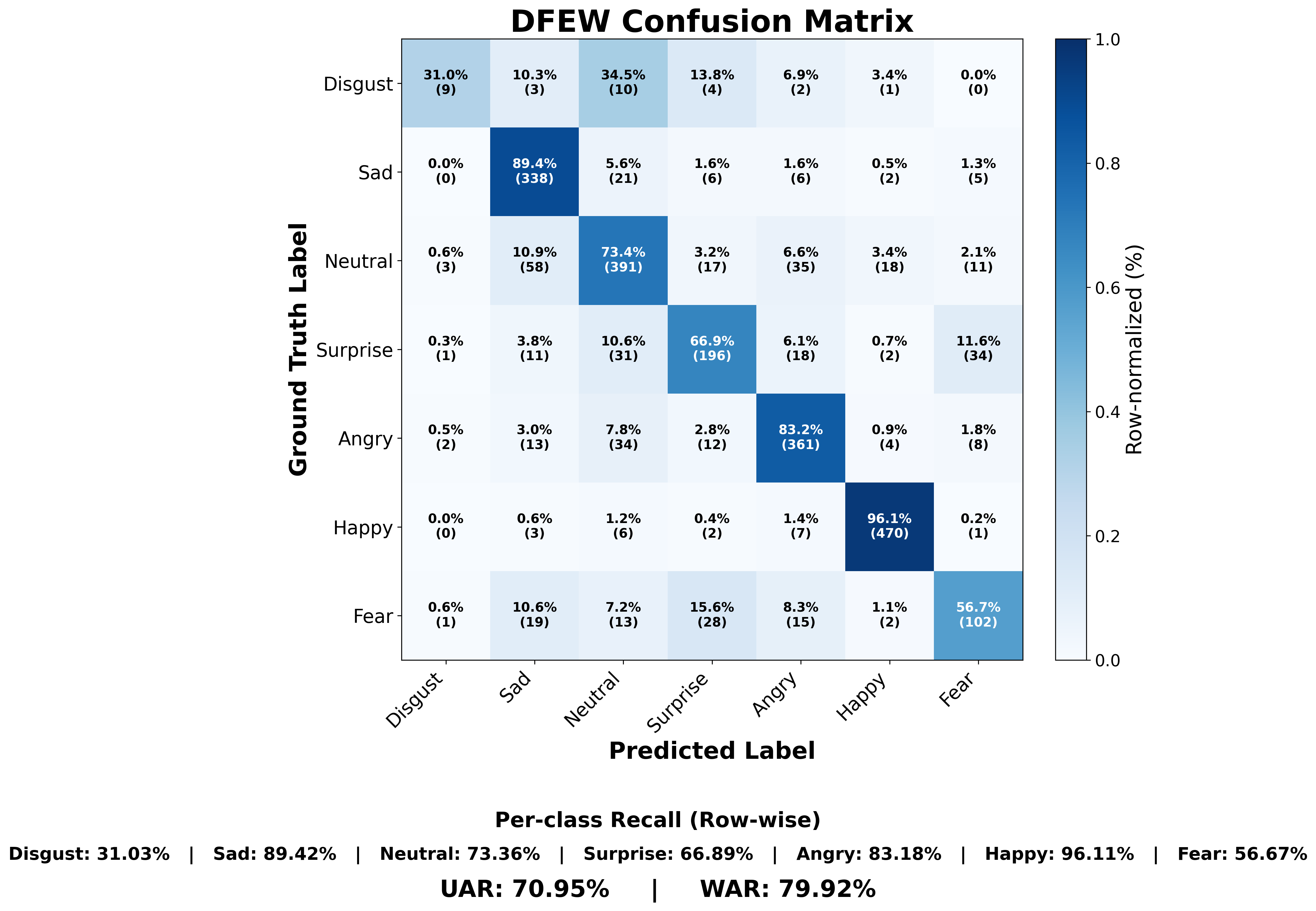}
&
\includegraphics[width=0.29\textwidth]{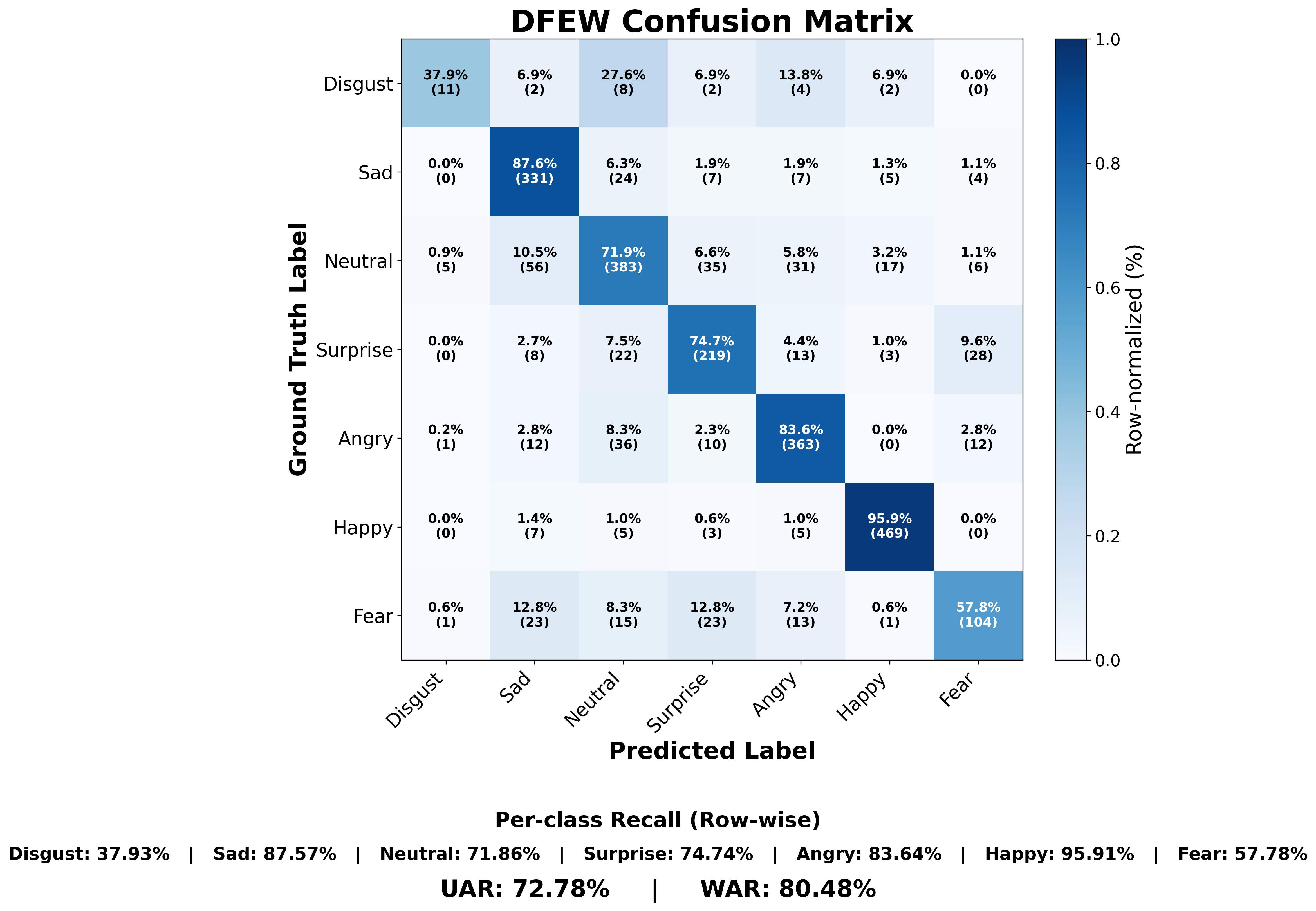}

\end{tabular}

\caption{
\textbf{Confusion matrices on DFEW across all five folds. Best viewed digitally in color with zoom.}
Rows correspond to the five cross-validation folds, and columns correspond to ViT-B, ViT-L, and ViT-H with FlashLite.
}
\label{fig:dfew_confusion_all}
\end{figure}
\begin{figure}[t]
\centering
\setlength{\tabcolsep}{2pt}
\begin{tabular}{cccc}

&
\textbf{ViT-B}
&
\textbf{ViT-L}
&
\textbf{ViT-H}
\\[2mm]

\raisebox{0.035\textheight}{\textbf{Fold 1}}
&
\includegraphics[width=0.29\textwidth]{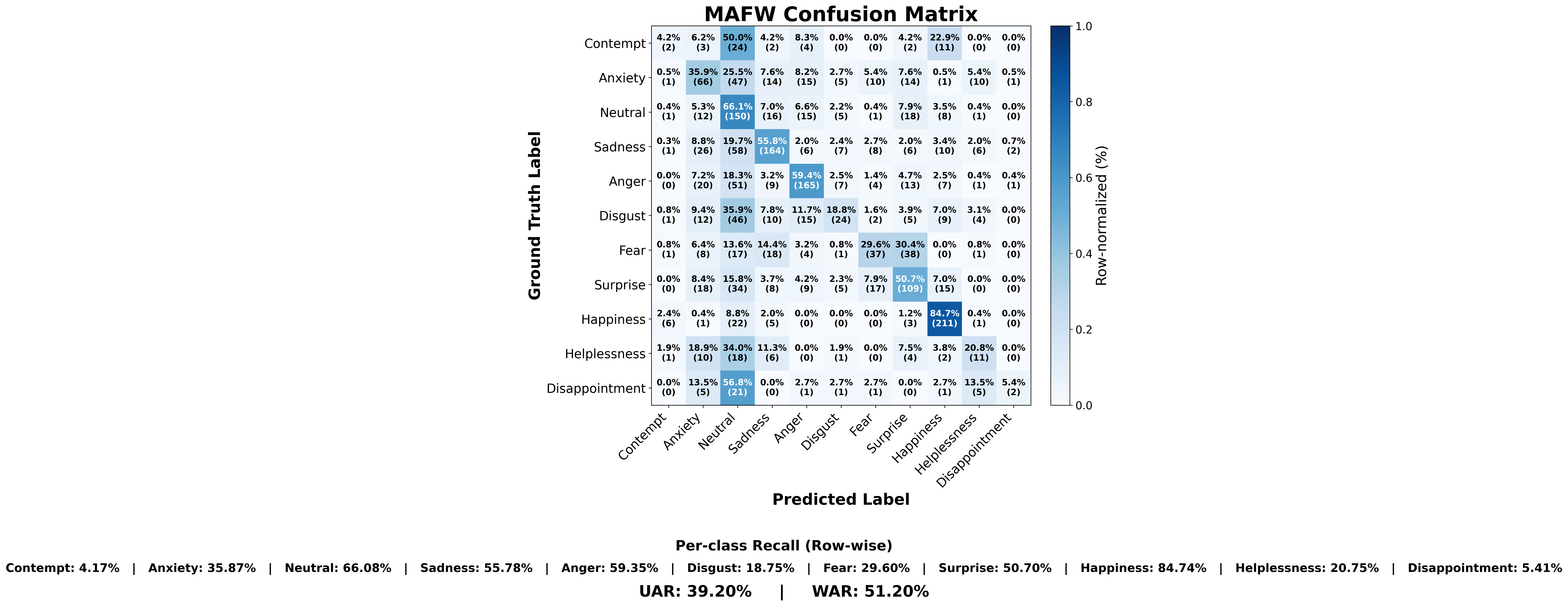}
&
\includegraphics[width=0.29\textwidth]{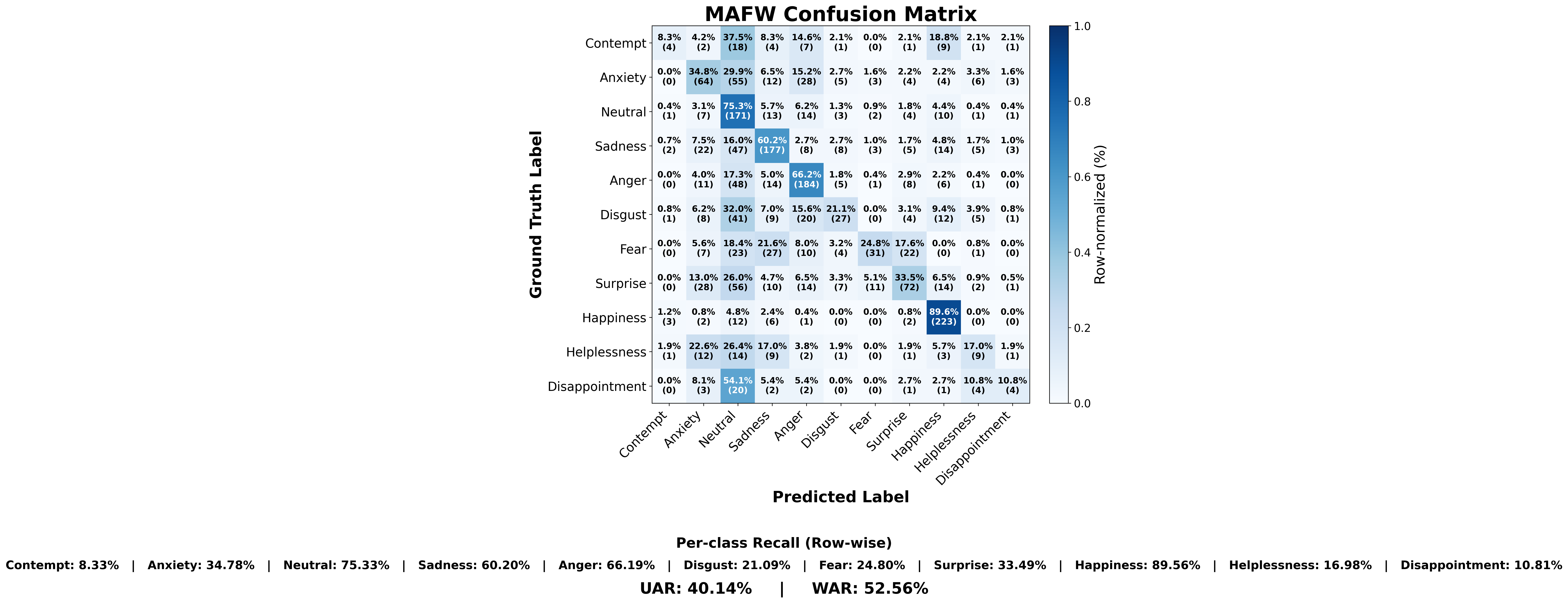}
&
\includegraphics[width=0.29\textwidth]{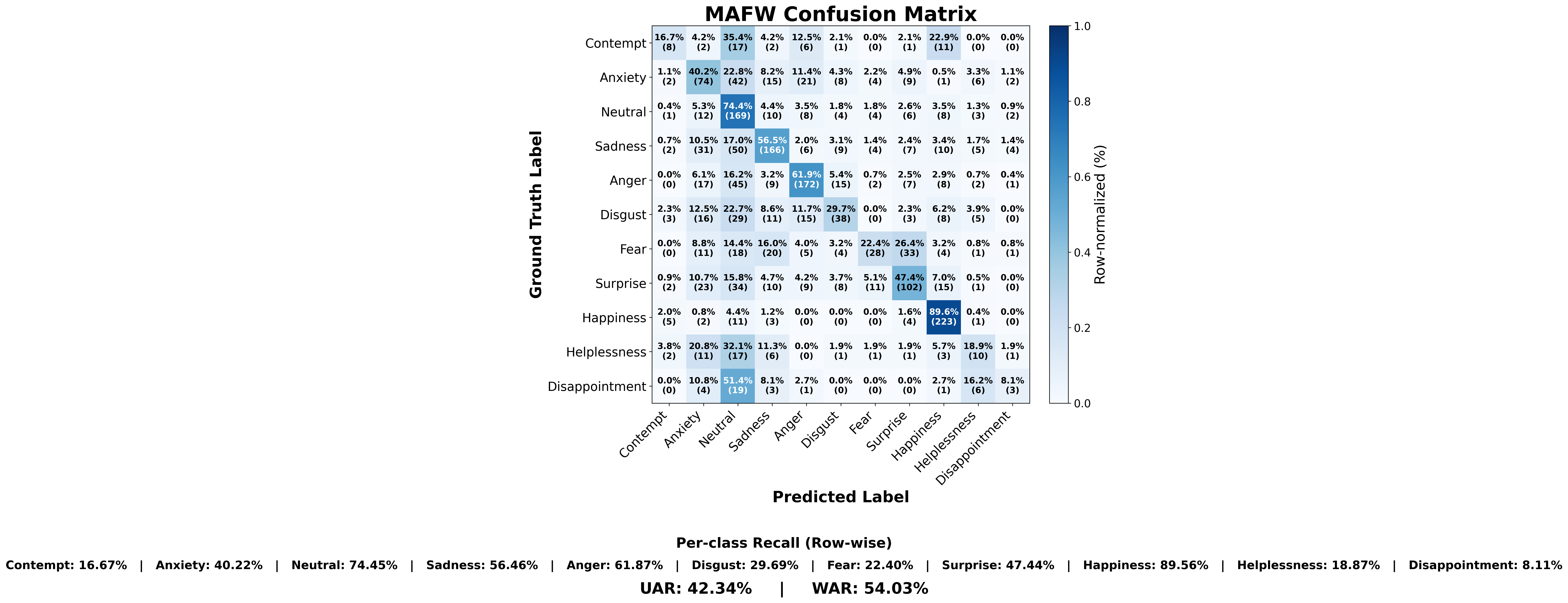}
\\[2mm]

\raisebox{0.035\textheight}{\textbf{Fold 2}}
&
\includegraphics[width=0.29\textwidth]{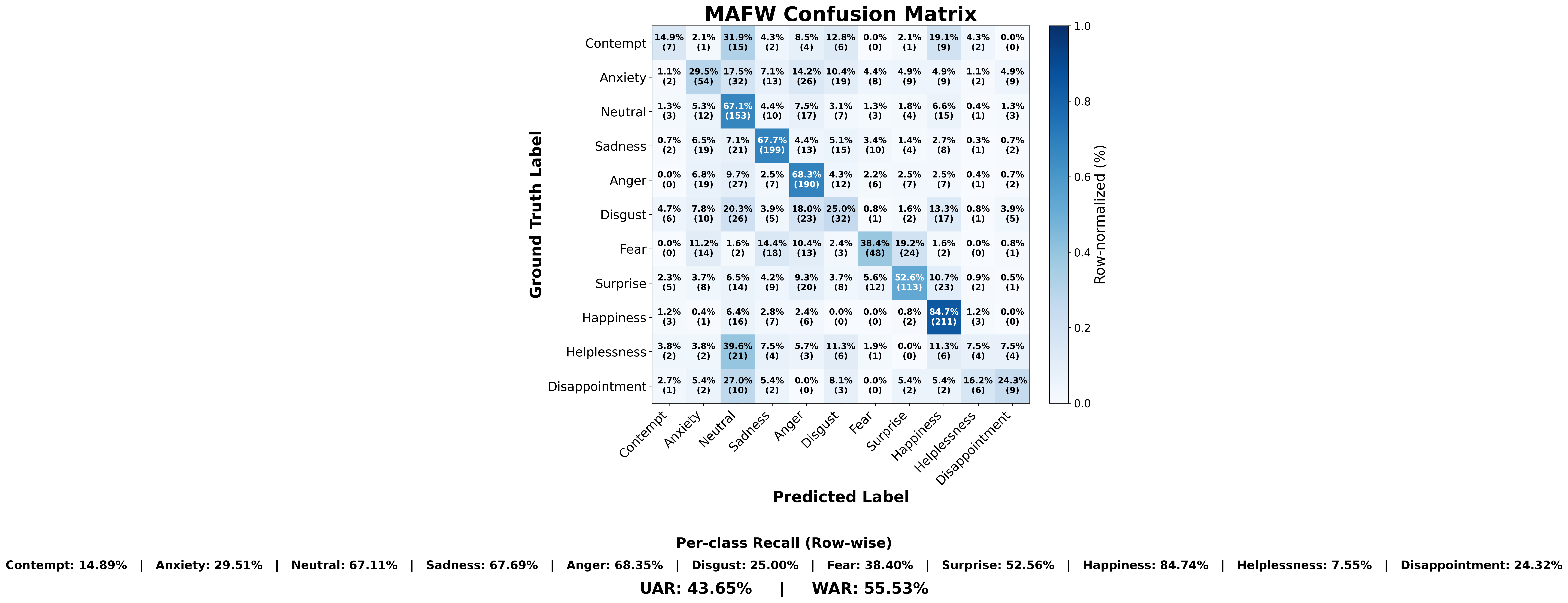}
&
\includegraphics[width=0.29\textwidth]{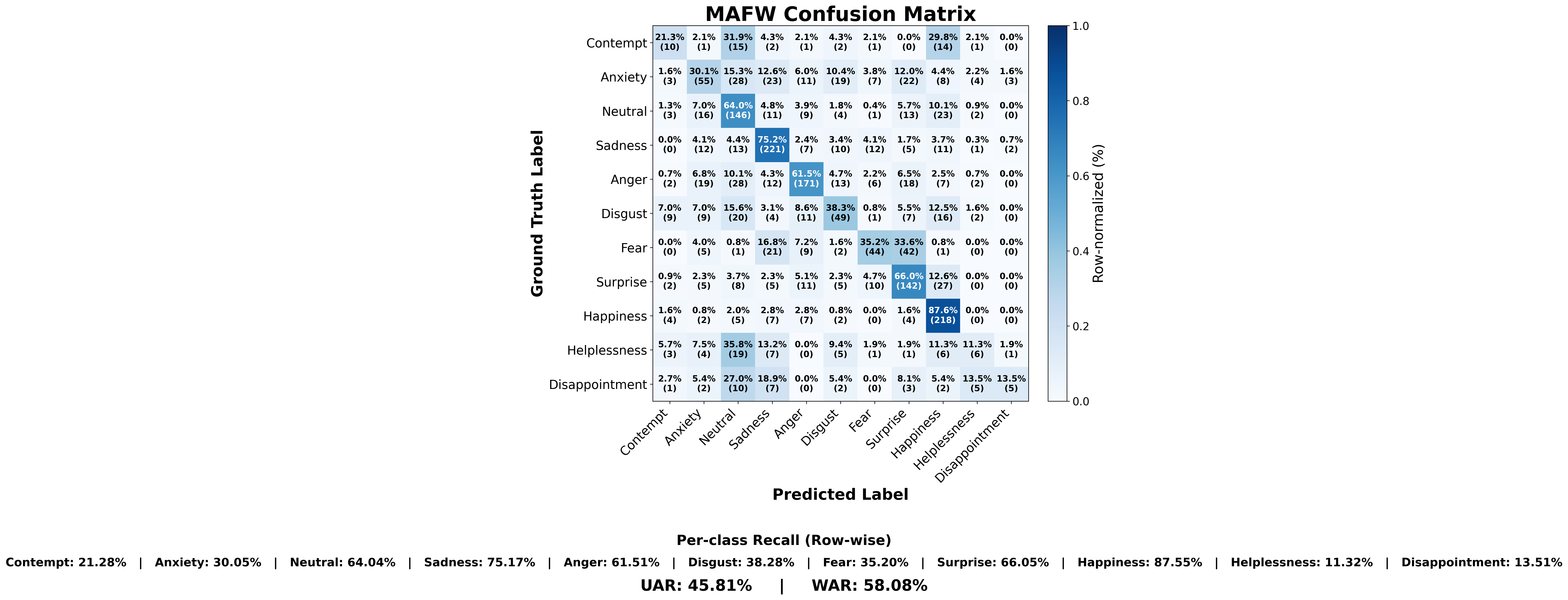}
&
\includegraphics[width=0.29\textwidth]{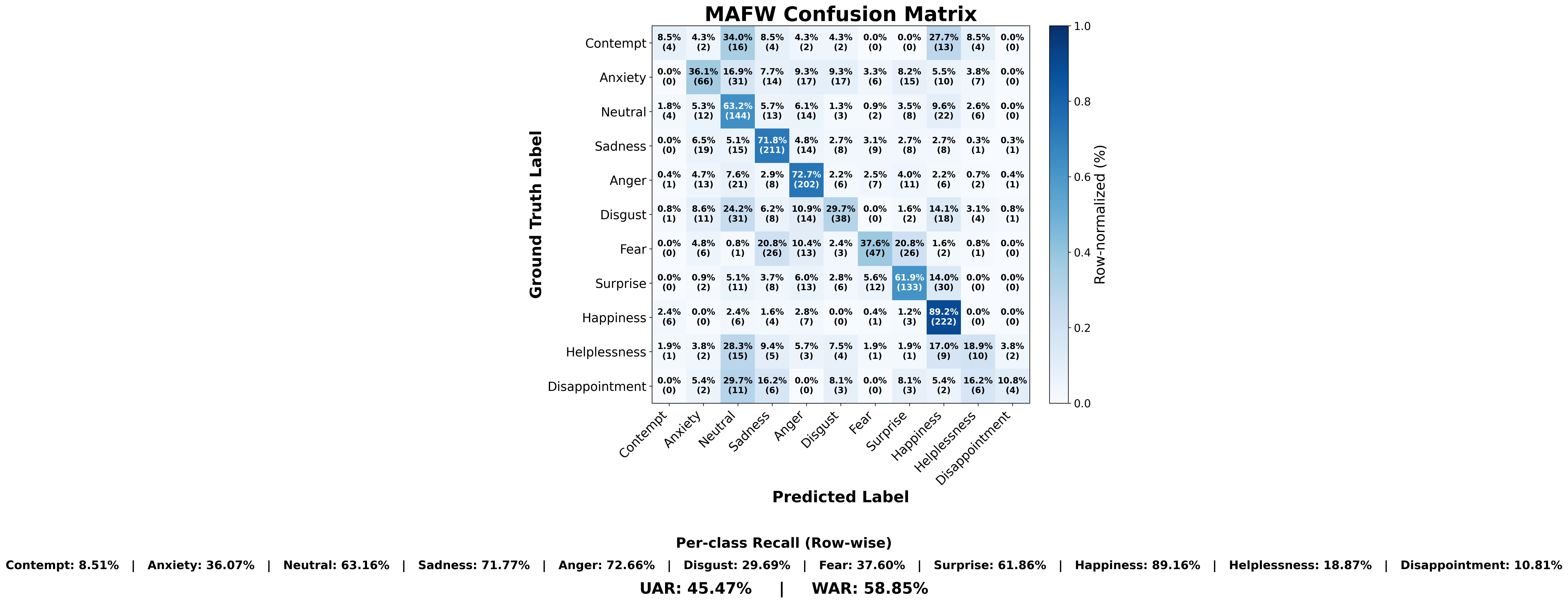}
\\[2mm]

\raisebox{0.035\textheight}{\textbf{Fold 3}}
&
\includegraphics[width=0.29\textwidth]{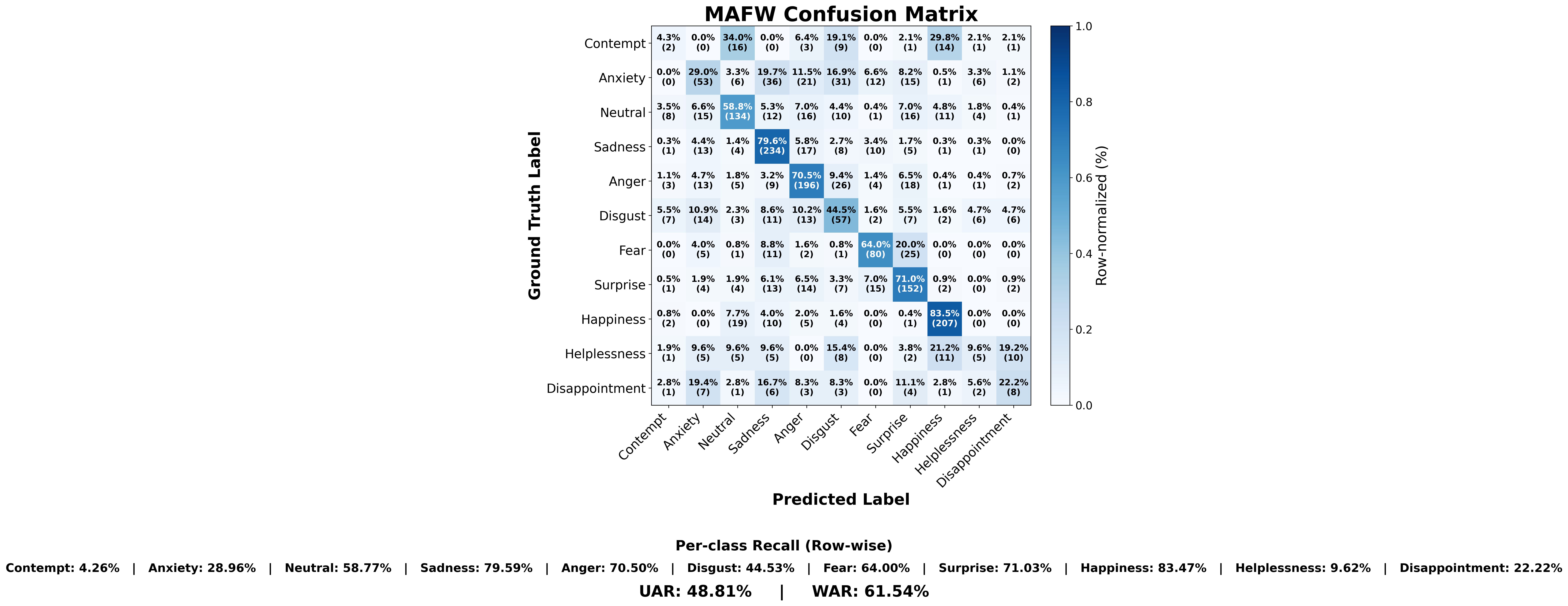}
&
\includegraphics[width=0.29\textwidth]{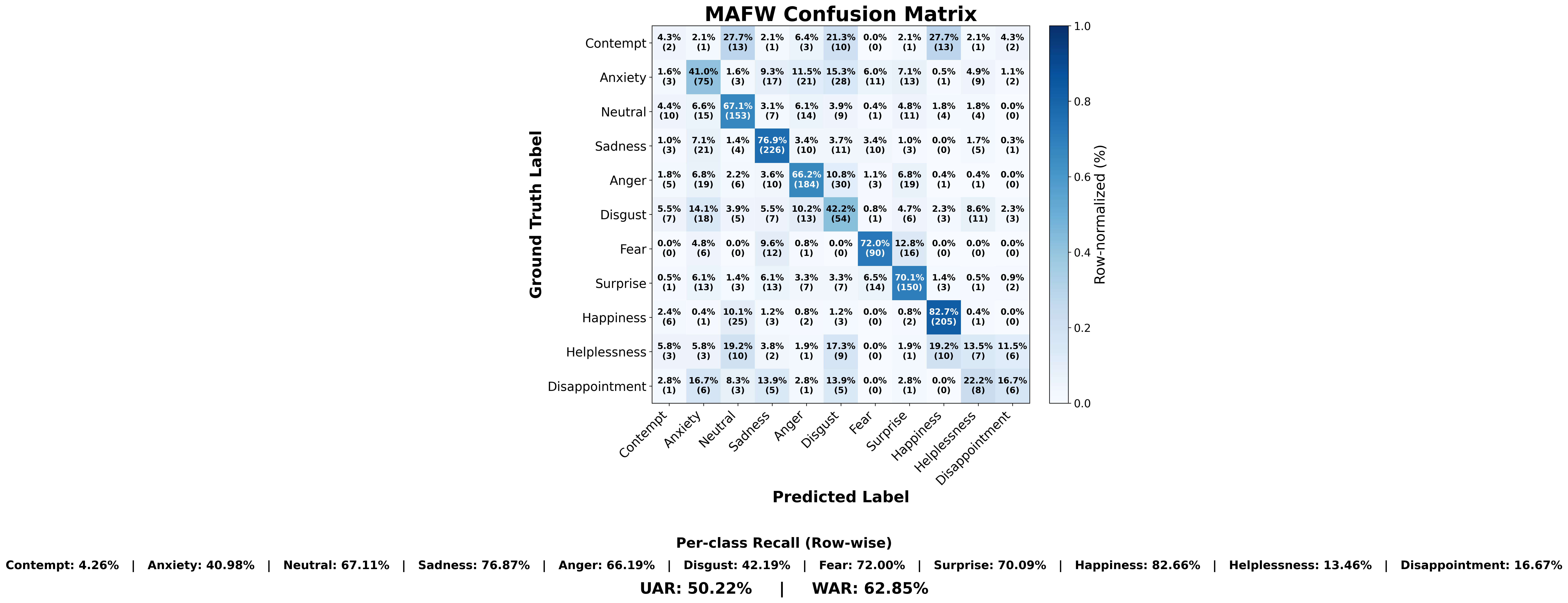}
&
\includegraphics[width=0.29\textwidth]{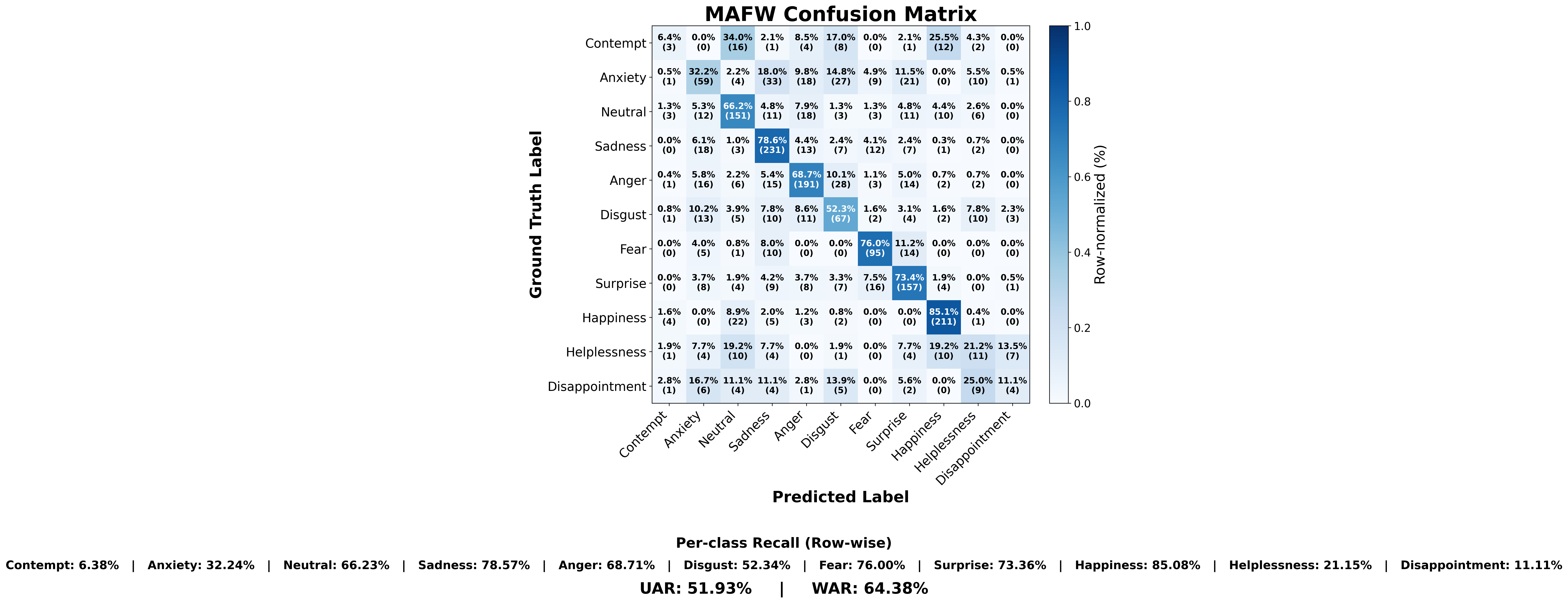}
\\[2mm]

\raisebox{0.035\textheight}{\textbf{Fold 4}}
&
\includegraphics[width=0.29\textwidth]{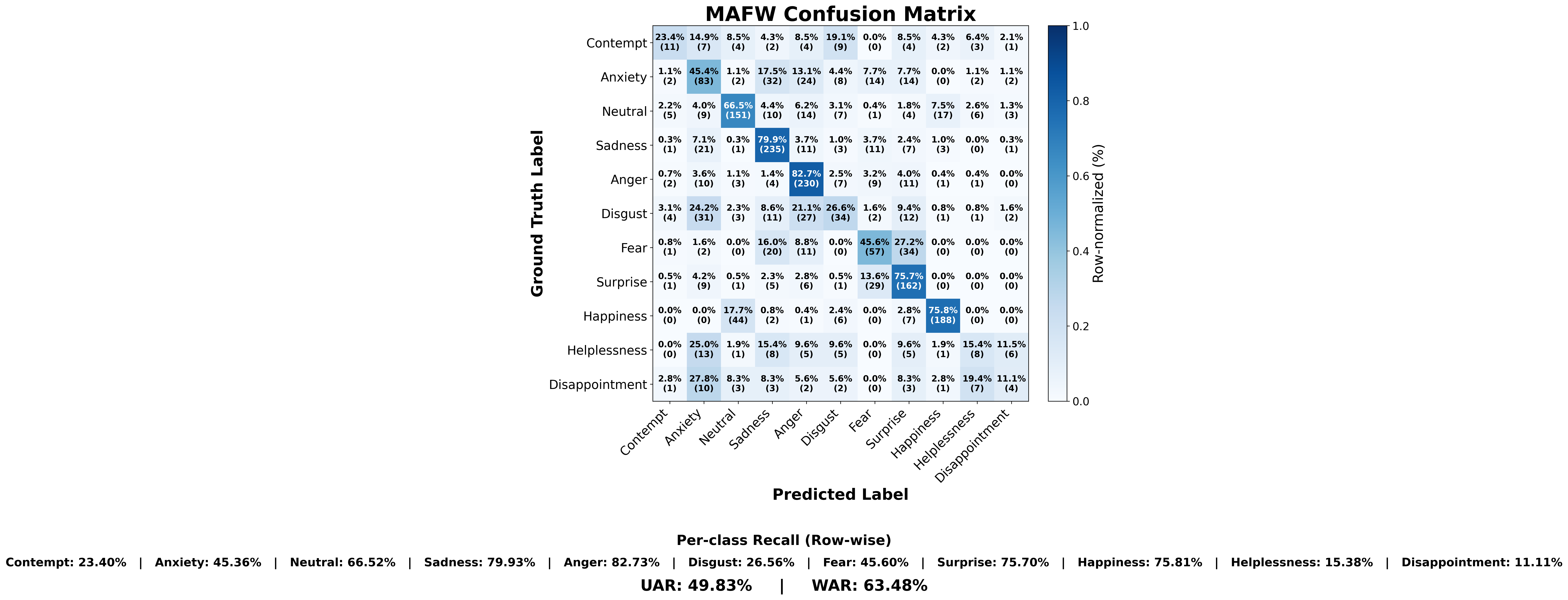}
&
\includegraphics[width=0.29\textwidth]{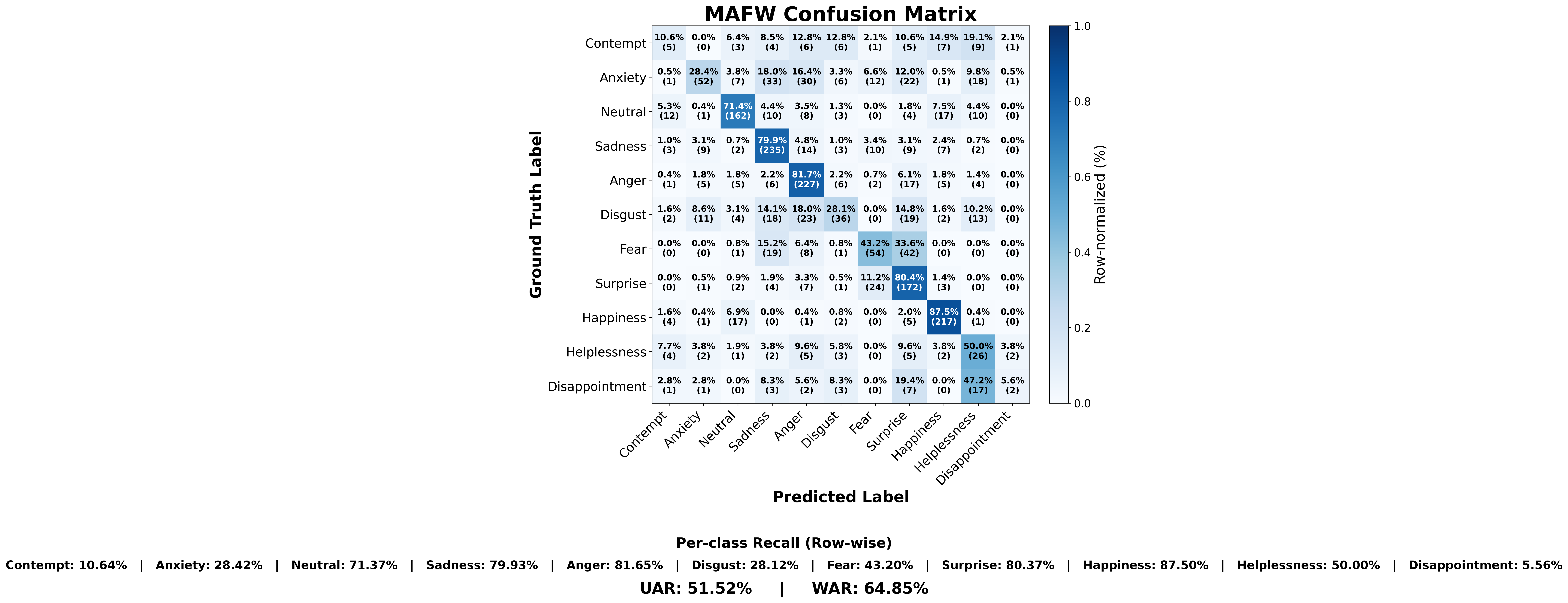}
&
\includegraphics[width=0.29\textwidth]{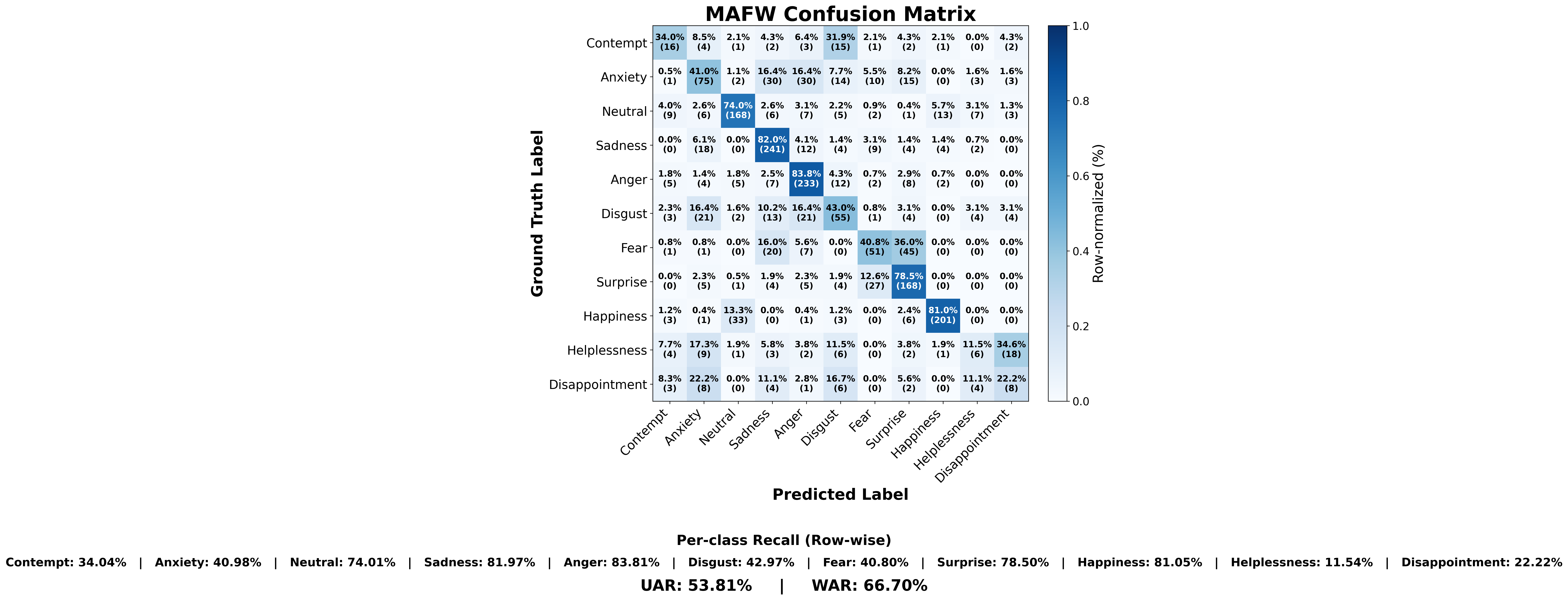}
\\[2mm]

\raisebox{0.035\textheight}{\textbf{Fold 5}}
&
\includegraphics[width=0.29\textwidth]{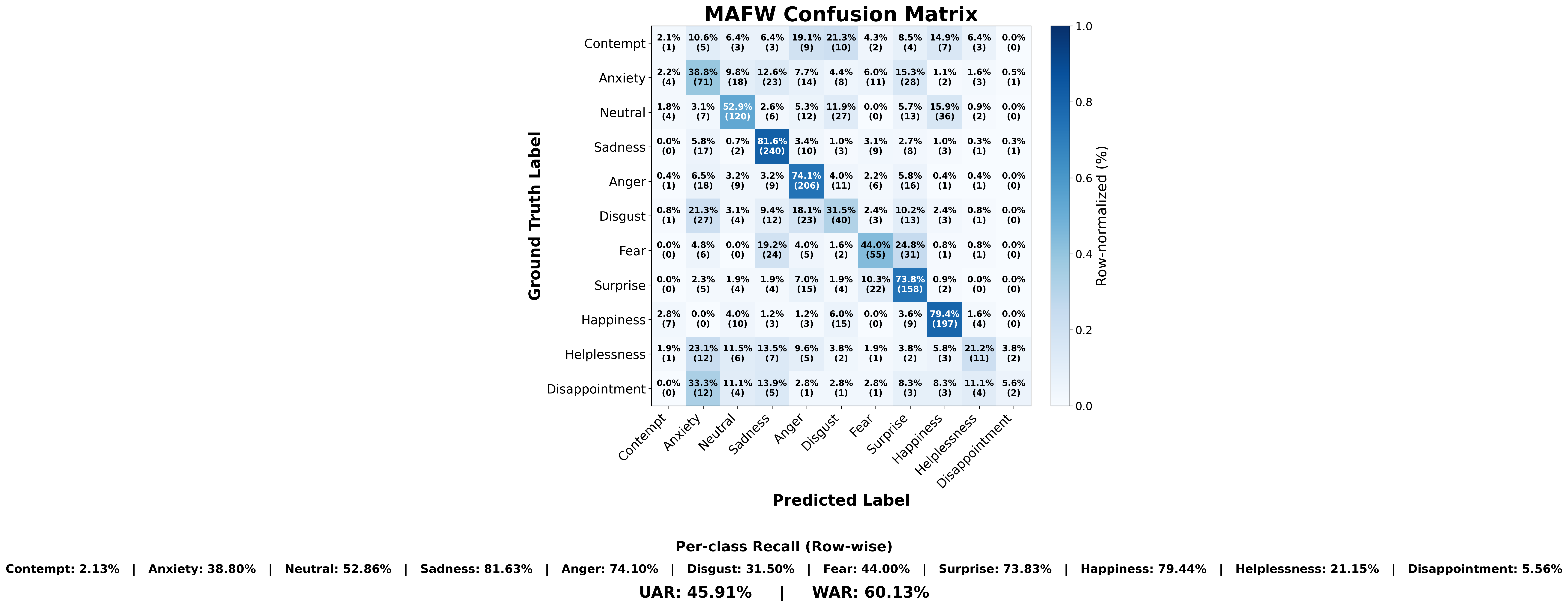}
&
\includegraphics[width=0.29\textwidth]{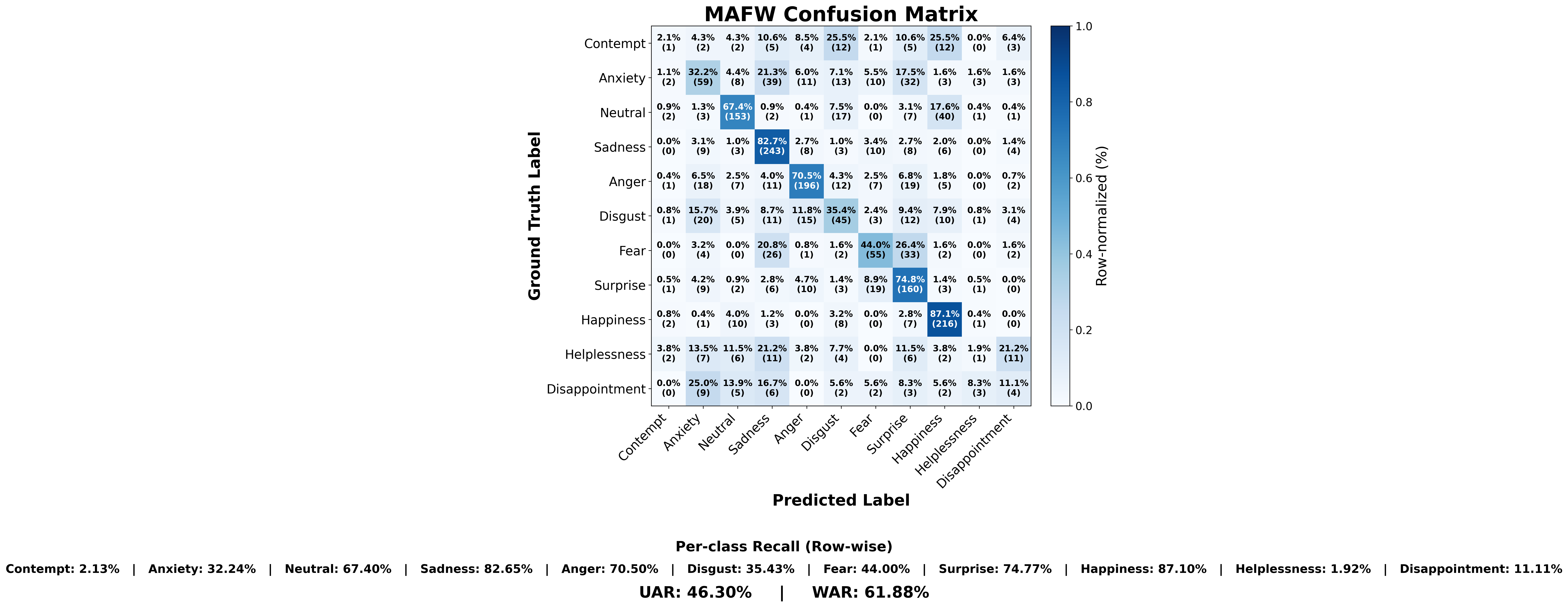}
&
\includegraphics[width=0.29\textwidth]{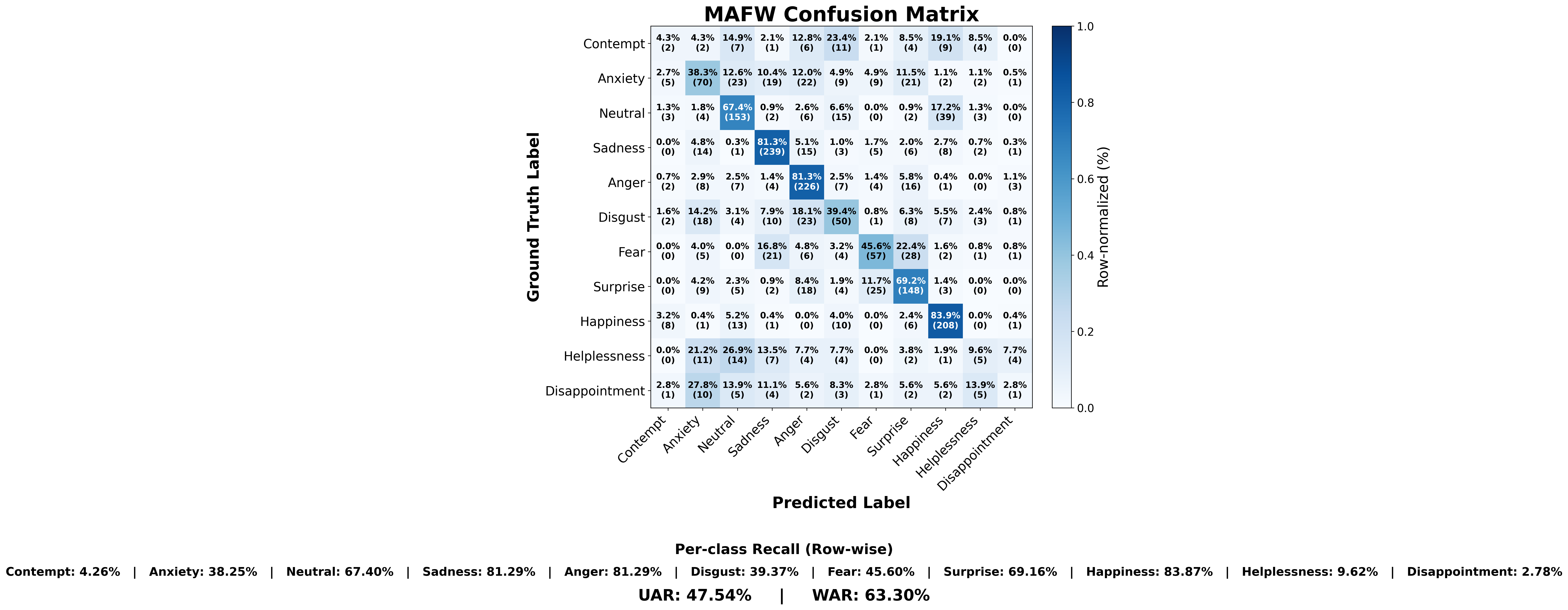}

\end{tabular}

\caption{
\textbf{Confusion matrices on MAFW across all five folds. Best viewed digitally in color with zoom.}
Rows correspond to the five cross-validation folds, and columns correspond to ViT-B, ViT-L, and ViT-H with FlashLite.
}
\label{fig:mafw_confusion_all}
\end{figure}
\begin{figure}[t]
\centering
\setlength{\tabcolsep}{2pt}
\begin{tabular}{ccc}

\textbf{ViT-B}
&
\textbf{ViT-L}
&
\textbf{ViT-H}
\\[2mm]

\includegraphics[width=0.31\textwidth]{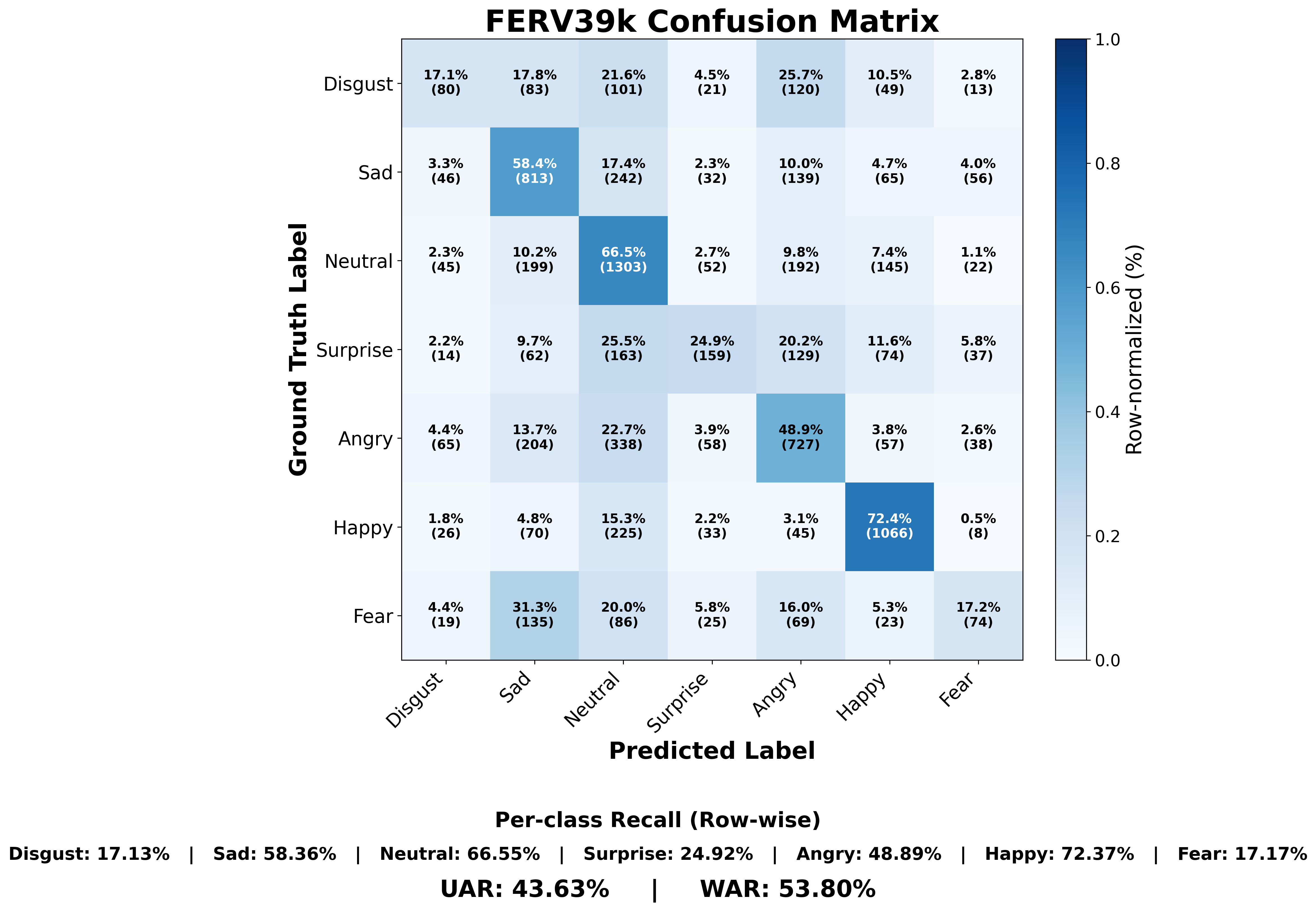}
&
\includegraphics[width=0.31\textwidth]{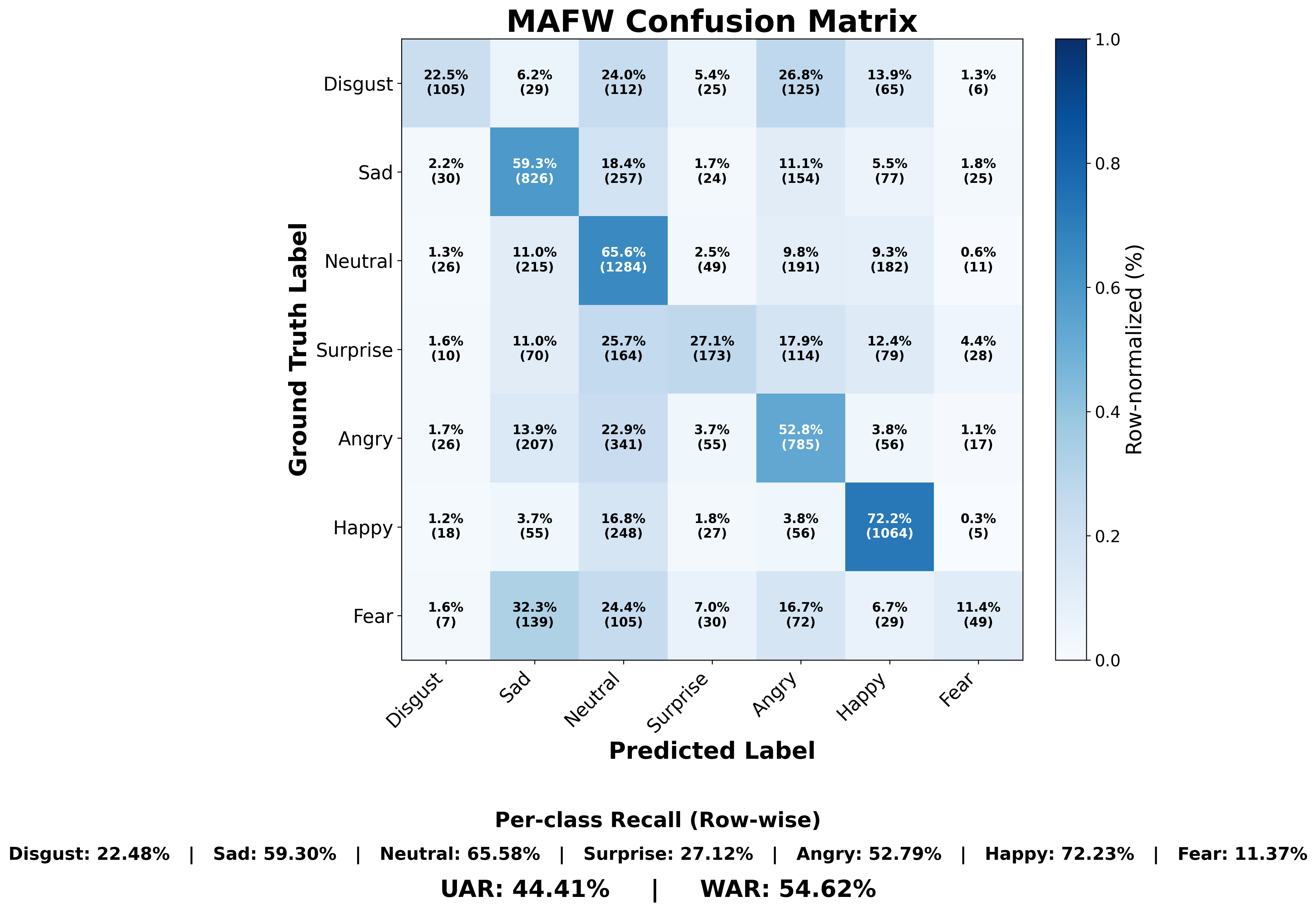}
&
\includegraphics[width=0.31\textwidth]{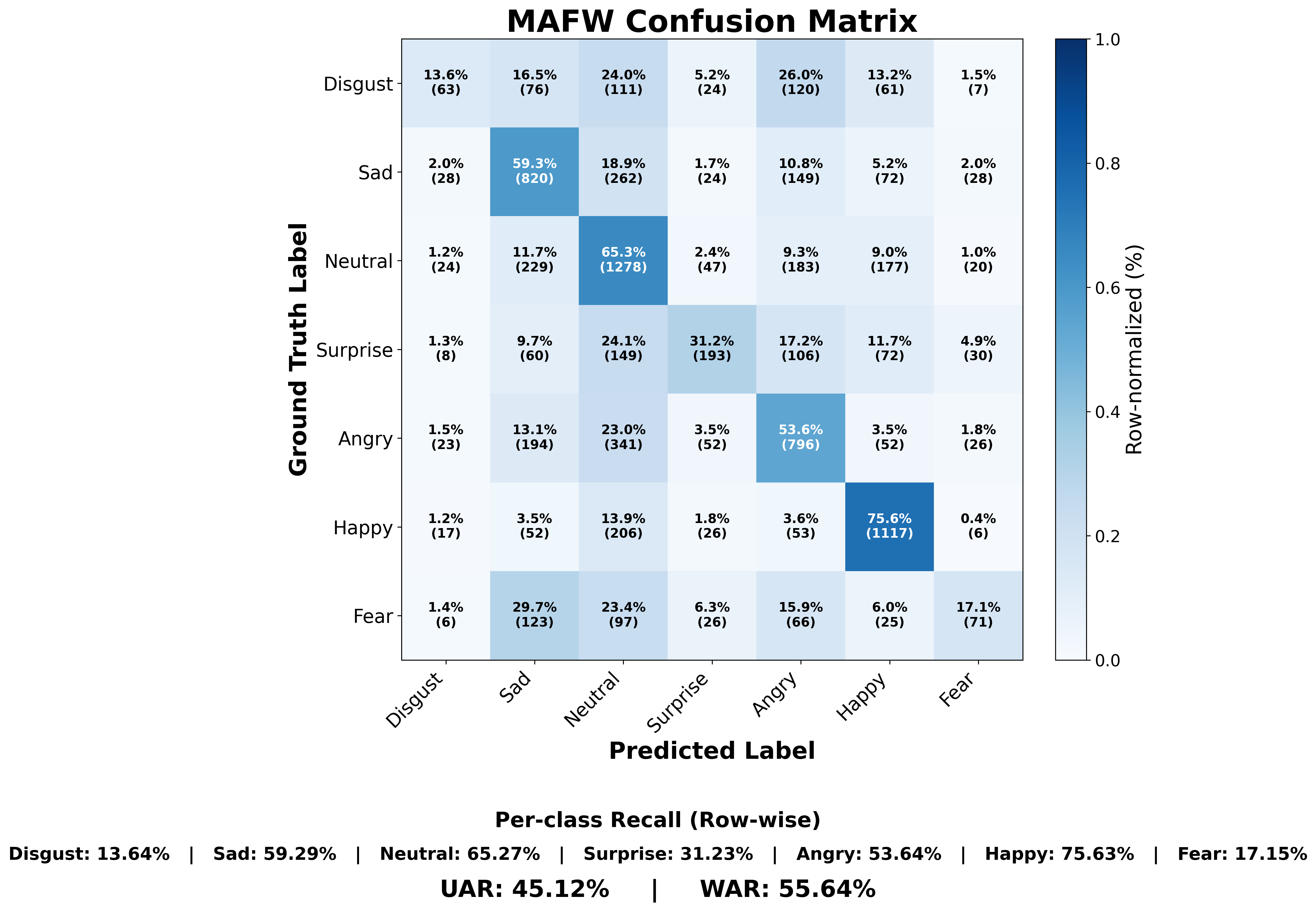}

\end{tabular}

\caption{
\textbf{Confusion matrices on FERV39k. Best viewed digitally in color with zoom.}
Columns correspond to ViT-B, ViT-L, and ViT-H with FlashLite.
}
\label{fig:ferv39k_confusion_all}
\end{figure}

\fi


\end{document}